\pdfoutput=1
\documentclass{article} %
\usepackage{iclr2027_conference,times}

\usepackage{amsmath,amsfonts,bm}

\def\eqref#1{equation~\ref{#1}}

\def\1{\bm{1}}

\DeclareMathAlphabet{\mathsfit}{\encodingdefault}{\sfdefault}{m}{sl}
\SetMathAlphabet{\mathsfit}{bold}{\encodingdefault}{\sfdefault}{bx}{n}

\usepackage{hyperref}
\usepackage{url}
\usepackage[utf8]{inputenc}
\usepackage[T1]{fontenc}
\usepackage{hyperref}
\usepackage{url}
\usepackage{booktabs}
\usepackage{amsfonts}
\usepackage{nicefrac}
\usepackage{microtype}
\usepackage{xcolor}
\usepackage{colortbl}
\usepackage{etoc}

\usepackage{amsmath}
\usepackage{amssymb}
\usepackage{amsthm}
\usepackage{mathtools}

\theoremstyle{plain}
\newtheorem{theorem}{Theorem}[section]

\newtheorem{proposition}[theorem]{Proposition}
\newtheorem{corollary}[theorem]{Corollary}

\usepackage{amsmath, amssymb, amsthm}
\theoremstyle{definition}
\newtheorem{definition}[theorem]{Definition}
\newtheorem{assumption}[theorem]{Assumption}

\newtheoremstyle{customremark}
  {}{}
  {\normalfont}
  {}
  {\bfseries}
  {.}
  { }
  {\thmname{#1}\thmnumber{ #2}\thmnote{ (#3)}}

\theoremstyle{customremark}
\newtheorem{remark}[theorem]{Remark}

\usepackage{graphicx}
\usepackage{float}     
\usepackage{subcaption}
\usepackage{algorithm}
\usepackage{algpseudocode}
\algtext*{EndIf}
\usepackage{tabularx}
\usepackage{longtable}
\usepackage{multirow}
\usepackage{makecell}
\usepackage{wrapfig}
\usepackage{placeins}
\usepackage{needspace}
\usepackage{amssymb}
\usepackage{pifont}
\usepackage{tcolorbox}
\tcbuselibrary{skins, breakable}
\tcbset{
  manuscriptbox/.style={
    colback=black!2!white,
    colframe=black!70!white,
    colbacktitle=black!10!white,
    coltitle=black,
    coltext=black,
    fonttitle=\bfseries,
    boxrule=0.5pt,
    titlerule=0.3pt,
    arc=0pt,
    left=10pt, right=10pt, top=4pt, bottom=4pt
  }
}
\usepackage{enumitem}

\newcommand{\xmark}{\ding{55}}

\title{Fisher-Guided Progressive Parameter Selection for Adaptive Fine-Tuning}

\author{
Ghodsiyeh Rostami$^{1}$, Po-Han Chen$^{1}$, Mahdi S. Hosseini$^{1,2}$ \\
\normalfont
$^{1}$ Concordia University, Montr\'{e}al, Canada \\
$^{2}$ Mila - Qu\'{e}bec AI Institute, Montr\'{e}al, Canada \\
\texttt{rose.rostami@mail.concordia.ca} \\
\texttt{\{pohan.chen, mahdi.hosseini\}@concordia.ca}
}

\iclrfinalcopy %
\begin{document}
\etocdepthtag.toc{main}

\maketitle

\begin{abstract}
Parameter-efficient fine-tuning often selects trainable parameters before adaptation using architectural heuristics, without accounting for their varying importance during training. We introduce \textbf{FisherAdapTune}, which progressively selects parameter groups based on temporal drift in their Fisher information. Under a local Gaussian approximation, we bound the divergence between the fine-tuned posterior and pretrained prior by accumulated Fisher-weighted update costs, motivating curvature-aware selection. FisherAdapTune measures Jensen-Shannon distance between successive Fisher-value distributions and uses an adaptive threshold to freeze stabilized groups. Across VTAB-1k classification tasks, it achieves the highest macro Top-1 accuracy among the compared methods with a smaller average trainable set than full fine-tuning. Across four segmentation backbones, it maintains competitive in-distribution performance and improves zero-shot transfer in several settings. Its selections reveal architecture-dependent patterns where input, output, and normalization parameters can remain trainable as attention and MLP groups freeze at different rates. The results support Fisher structural drift as a task-dependent signal for allocating updates during adaptation. We release our \href{https://github.com/AtlasAnalyticsLab/FisherAdapTune}{code} publicly to enable further application of our proposed approach.
\end{abstract}

\vspace{-5pt}
\begin{figure}[H]
    \centering
    \includegraphics[width=0.85\linewidth]{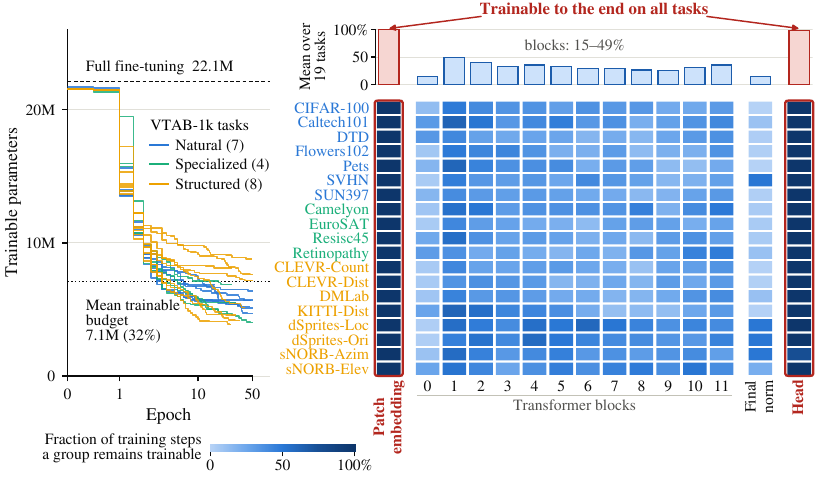}
\caption{\textbf{FisherAdapTune uses variations in Fisher structure to progressively select parameters for each task and architecture.} 
Across VTAB-1k tasks with DINOv2 ViT-S/14, input and output layers remain trainable the longest, consistent with adaptation to input distribution shifts and task-specific outputs. See Sections \ref{sec:methodology} and Appendix \ref{appendix:param_selection} for further discussion.}
    \label{fig:placeholder}
\end{figure}
\vspace{-15pt}

\section{Introduction}
\vspace{-0.2cm}
Adapting pretrained deep neural networks to downstream tasks is a central problem in modern machine learning. While large-scale pretraining enables models to learn rich and transferable representations, leveraging these representations under limited data, distribution shifts, and computational constraints remains challenging \citep{zhuangComprehensiveSurveyTransfer2021,kumarFineTuningCanDistort2021}. In practice, downstream environments are dynamic where data distributions evolve, tasks change, and models must be updated frequently. This makes efficient and robust adaptation essential for scalable deployment, especially when retraining from scratch is infeasible.
Full fine-tuning, where all parameters are updated for the target task, is computationally expensive and memory-intensive, especially for large models, and can lead to overfitting or degradation in generalization under low-data regimes \citep{zhangParameterEfficientFineTuningFoundation2025}. 
Moreover, recent studies suggest that classical scaling laws do not directly transfer to fine-tuning settings, with performance depending on complex interactions between model size, pretraining, and data regime \citep{tayScaleEfficientlyInsights2021,linSelectingLargeLanguage2024,parkHowVisionTransformers2021}. These observations highlight the need for principled adaptation strategies beyond naïve full fine-tuning.

\begin{wraptable}[10]{r}{0.60\textwidth}
\centering
\vspace{-7pt}
\captionsetup{font=footnotesize,skip=3pt}
\caption{Comparison of PEFT categories.
$^{1}$Parameter Efficiency,
$^{2}$Task-Aware Configuration,
$^{3}$Architectural Modifications,
$^{4}$Theory-Grounded,
$^{5}$Generalization,
$^{6}$Inference Overhead.}
\label{tab:peft_comparison}
\scriptsize
\setlength{\tabcolsep}{2pt}
\renewcommand{\arraystretch}{0.9}
\resizebox{\linewidth}{!}{%
\begin{tabular}{@{}lcccccc@{}}
\toprule
\textbf{Method} &
\makecell{\textbf{Param.}\\\textbf{Eff.$^{1}$}} &
\makecell{\textbf{Task-}\\\textbf{Aware$^{2}$}} &
\makecell{\textbf{Arch.}\\\textbf{Mod.$^{3}$}} &
\textbf{Theory.$^{4}$} &
\textbf{Gen.$^{5}$} &
\makecell{\textbf{Inf.}\\\textbf{OH$^{6}$}} \\
\midrule
Full Fine-Tuning               & \xmark     & \xmark & \xmark & \xmark & \xmark & \xmark \\
Reparameterization \citep{huLoRALowRankAdaptation2021}             & \checkmark & \xmark & \xmark & \checkmark & \xmark & \xmark \\
Selective \citep{zakenBitFitSimpleParameterefficient2022}  & \checkmark & \xmark & \xmark & \xmark & \xmark & \xmark \\
Prompt Tuning \citep{bahngExploringVisualPrompts2022}               & \checkmark & \xmark & \checkmark & \xmark & \xmark & \checkmark \\
Adapter-based \citep{chenAdaptFormerAdaptingVision2022a}    & \checkmark & \xmark & \checkmark & \xmark & \xmark & \checkmark \\
\midrule
\textbf{Ours (FisherAdapTune)} & \checkmark & \checkmark & \xmark & \checkmark & \checkmark & \xmark \\
\bottomrule
\end{tabular}%
}
\end{wraptable}

To address these challenges, parameter-efficient fine-tuning reduces the trainable parameter set while seeking to preserve performance. Existing approaches include low-rank reparameterization, such as LoRA \citep{wangLoRAProAreLowRank2024,heParameterEfficientModelAdaptation2023a,zhang2023adalora}, learnable prompts \citep{bahngExploringVisualPrompts2022,huangDiversityAwareMetaVisual2023,jiaVisualPromptTuning2022}, lightweight adapters \citep{houlsby2019parameter,chenAdaptFormerAdaptingVision2022a,chenVisionTransformerAdapter2022,jieConvolutionalBypassesAre2024}, and selective tuning of predefined subsets, such as biases or normalization layers \citep{zakenBitFitSimpleParameterefficient2022,zhaoTuningLayerNormAttention2023a}. As summarized in Table~\ref{tab:peft_comparison}, these methods can achieve competitive performance with fewer trainable parameters. Yet many rely on fixed architectural choices that cannot account for how parameter relevance evolves during training. Although data-driven selection has been explored \citep{zhaoMaskingEfficientAlternative2020}, a principled framework for allocating trainable capacity to task-relevant parameters throughout fine-tuning remains an open challenge.

\begin{figure}[H]
    \centering
    \begin{minipage}[t]{0.60\linewidth}
        \vspace{0pt}
        \centering
        \includegraphics[width=\linewidth]{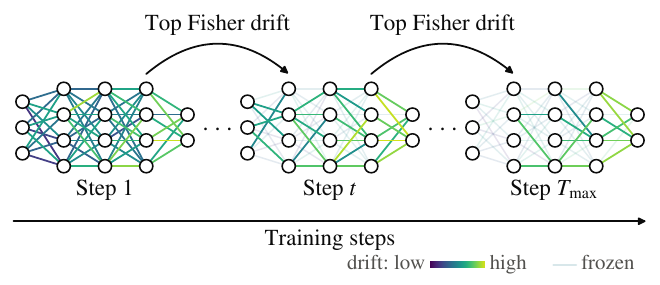}
        \caption{\textbf{FisherAdapTune} periodically measures Fisher information reorientation via Jensen-Shannon distance between consecutive Fisher distributions to identify stabilized parameter groups. Parameter groups with high Fisher drift remain tunable, while stabilized groups are frozen.}
        \label{fig:fisher_adap_tune}
    \end{minipage}\hfill
    \begin{minipage}[t]{0.37\linewidth}
        \vspace{0pt}
        \centering
        \includegraphics[width=0.9\linewidth]{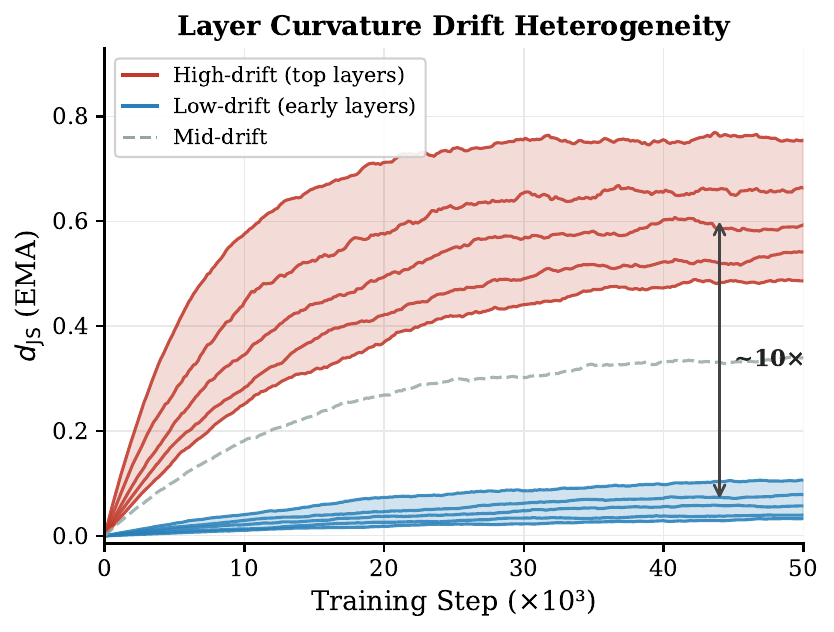}
        \caption{Layer-wise Fisher drift during full fine-tuning is uneven, motivating FisherAdapTune's progressive freezing of stabilized layers.}
        \label{fig:intro_layers_drift}
    \end{minipage}
\end{figure}
\vspace{-11pt}

In this work, we propose \textbf{FisherAdapTune}, a Fisher-guided Adaptive Fine-Tuning framework for dynamic parameter selection during fine-tuning (Figure \ref{fig:fisher_adap_tune}).
A PAC-Bayesian analysis relates divergence from the pretrained prior to accumulated Fisher-weighted updates, motivating curvature-aware parameter selection during fine-tuning.
Building upon this, our approach is grounded in the temporal evolution of the Fisher Information Matrix, which captures local curvature and parameter sensitivity. Instead of relying on static heuristics, we measure the \emph{Jensen-Shannon (JS) distance} between successive Fisher distributions to track how parameter groups evolve during training (Figure \ref{fig:intro_layers_drift}). 
This provides a data-driven signal for tracking how parameter groups evolve: groups with persistent Fisher drift remain candidates for continued updating, whereas groups with low drift become candidates for freezing.
 This yields a dynamic selective fine-tuning strategy that continuously adjusts the trainable parameter set throughout optimization. By leveraging Fisher \emph{structural drift}, FisherAdapTune identifies task-relevant parameters in a principled manner and avoids unnecessary updates.
We validate FisherAdapTune on VTAB-1k classification tasks and a domain-specific dense prediction task, demonstrating that it achieves competitive or superior performance compared to existing PEFT baselines. Furthermore, our approach improves robustness under distribution shifts, highlighting the importance of adaptive, task-aware parameter selection for generalization.

Our \textbf{main contributions} are:
\begin{itemize}[
    leftmargin=1.0em,
    labelsep=0.5em,
    itemsep=4pt,
    parsep=0pt,
    topsep=-2pt
]
\item We derive a PAC-Bayesian bound that relates divergence from the pretrained prior to accumulated Fisher-weighted updates, providing a theoretical motivation for curvature-aware selection.\par
\item
We develop a task-dependent selection criterion that measures temporal changes in group-wise Fisher-value distributions using Jensen-Shannon distance.\par
\item
We introduce \textbf{FisherAdapTune}, an adaptive, model-agnostic fine-tuning framework that progressively freezes groups with stabilized Fisher drift, discovering architecture-dependent trainable sets without a prescribed layer-freezing order.\par
\item
Evaluate FisherAdapTune on VTAB-1k classification tasks and a downstream segmentation task across four backbones, achieving the highest macro classification accuracy among the evaluated methods with a reduced trainable budget, alongside competitive segmentation and improved zero-shot transfer in several settings.\par
\end{itemize}

\section{Related Work}
\vspace{-0.2cm}
\textbf{Parameter-Efficient Fine-Tuning} aims to adapt pretrained models with reduced computational and memory cost while maintaining performance \citep{zhangParameterEfficientFineTuningFoundation2025,hanParameterEfficientFineTuningLarge2024}. Selective fine-tuning methods, inspired in part by the Lottery Ticket Hypothesis \citep{frankleLotteryTicketHypothesis2019a}, update only a subset of parameters to mitigate overfitting and improve efficiency \citep{aghajanyanIntrinsicDimensionalityExplains2020}. Existing approaches identify important parameters heuristically using magnitude of updates (e.g., LT-SFT \citep{ansellComposableSparseFineTuning2023}), architectural choices (e.g., LayerNorm tuning \citep{zhaoTuningLayerNormAttention2023a}), or probabilistic criteria (e.g., BayesTune \citep{kimBayesTuneBayesianSparse2023}, while others rely on learnable masks (Masking \citep{zhaoMaskingEfficientAlternative2020}, Diff-Pruning \citep{guoParameterEfficientTransferLearning2021} or gradient dynamics (AutoFreeze \citep{liuAutoFreezeAutomaticallyFreezing2021}) to guide selection. Despite promising empirical results, these methods largely depend on heuristic signals and lack a principled, theoretically grounded mechanism for identifying task-relevant parameters. 
On the other hand, 
\textbf{Fisher-guided methods} use the Fisher Information Matrix to estimate parameter sensitivity \citep{kirkpatrick2017overcoming,miMakeSharpnessAwareMinimization2022}. Child-Tuning \citep{xuRaiseChildLarge2021} and FISH-Mask \citep{sungTrainingNeuralNetworks2021} use diagonal Fisher estimates to select sparse parameter subsets prior to fine-tuning. More recently, FisherTune \citep{zhaoFisherTuneFisherGuidedRobust2025} uses a domain-related Fisher formulation to identify parameters sensitive to task and domain shifts, with the goal of improving cross-domain generalization in semantic segmentation. Despite these advances, principled allocation of trainable capacity throughout fine-tuning remains an open challenge.

\section{Methodology}\label{sec:methodology}
Effective selective fine-tuning requires identifying both \emph{where adaptation remains necessary} and \emph{when parameter groups can be frozen}. We introduce \textbf{FisherAdapTune}, a selective fine-tuning framework, which uses the temporal evolution of Fisher information to guide these decisions. Section~\ref{sec:fisher_drift} establishes the theoretical foundation by connecting Fisher-weighted updates to a PAC-Bayesian generalization bound.  Section~\ref{sec:curvature_shift} decomposes this cost across layers and motivates Fisher drift as an adaptation signal. Finally, Section~\ref{sec:fisher_adaptune_algorithm} introduces a Jensen-Shannon distance criterion for measuring this drift, presents empirical evidence of its heterogeneity across layers and parameter groups, and develops the progressive freezing algorithm,  with an analysis of the resulting parameter selections.

\subsection{Role of Fisher Information in Generalization and Adaptation}
\label{sec:fisher_drift}

We revisit Probably Approximately Correct (PAC)-Bayesian learning theory \citep{mcallesterPACBayesianModelAveraging1999, mcallesterPACBayesianTheorems1999, dziugaite2017computing, zhou2018non, lotfi2022pac} to derive a principled basis for fine-tuning.  Let $P$ denote a prior distribution over model parameters, instantiated by the model’s pretrained weights, and let $Q_T$ denote the posterior after $T$ steps of fine-tuning. With probability at least $1-\delta$ over the training sample, the theorem bounds the generalization error as follows:
\vspace{-1mm}
\begin{equation}
\mathbb{E}_{Q_T}\!\left[\mathcal{L}_{\mathrm{out}}\right]
\;\leq\;
\mathbb{E}_{Q_T}\!\left[\mathcal{L}_{\mathrm{in}}\right]
\;+\;
\sqrt{
  \frac{D_{\mathrm{KL}}(Q_T \| P) + \log(N/\delta)}{2N}
},
\label{eq:pac_bayes}
\end{equation}
where $\mathcal L_{\mathrm{in}}$ is the empirical risk on $N$ training samples drawn i.i.d. from a distribution $\mathcal D$, and $\mathcal L_{\mathrm{out}}$ is the population risk under the same $\mathcal D$. The bound is governed by $D_{\mathrm{KL}}(Q_T \| P)$ where excessive deviation from the pretrained prior loosens the guarantee \citep{lotfi2022pac}.

To understand how fine-tuning updates affect this divergence, we consider how 
a small change in parameters shifts the model's output distribution.
Let $p_{\boldsymbol{\theta}}(\mathbf{y}\mid\mathbf{x})$ denote the
predictive distribution over $\mathbf{y}$ produced by a model with
parameter vector $\boldsymbol{\theta} \in \mathbb{R}^{|\boldsymbol{\theta}|}$
given input $\mathbf{x}$. A natural measure of the effect of a
perturbation $\boldsymbol{\delta} \in \mathbb{R}^{|\boldsymbol{\theta}|}$
on the model's predictions is
$D_{\mathrm{KL}}(p_{\boldsymbol{\theta}}(\mathbf{y}\mid\mathbf{x})
\| p_{\boldsymbol{\theta}+\boldsymbol{\delta}}(\mathbf{y}\mid\mathbf{x}))$.
It can be shown \citep{dangelKroneckerfactoredApproximateCurvature2025,
amariInformationGeometryIts2016, soenTradeOffsDiagonalFisher2024} that
as $\|\boldsymbol{\delta}\| \to 0$, a second-order expansion averaged
over the input distribution yields:
\begin{equation}
\mathbb{E}_{\mathbf{x}}
\!\left[
D_{\mathrm{KL}}\!\left(
p_{\boldsymbol{\theta}+\boldsymbol{\delta}}
\;\|\;
p_{\boldsymbol{\theta}}
\right)
\right]
=
\tfrac{1}{2}\,
\boldsymbol{\delta}^{\top}
\mathbf{F}_{\boldsymbol{\theta}}
\boldsymbol{\delta}
+
\mathcal{O}(\|\boldsymbol{\delta}\|^3),
\label{eq:fisher_kl_relation}
\end{equation}
where $\mathbf{F}_{\boldsymbol{\theta}}$ is the Fisher Information
Matrix (FIM) \citep{amariNaturalGradientWorks1998}, capturing the
local curvature of the predictive distribution $\mathbf{F}_{\boldsymbol{\theta}}
=
\mathbb{E}_{\mathbf{x}}
\left[
\mathbb{E}_{\mathbf{y}\sim p_{\boldsymbol{\theta}}(\mathbf{y}\mid\mathbf{x})}
\left[
\nabla_{\boldsymbol{\theta}}
\log p_{\boldsymbol{\theta}}(\mathbf{y}\mid\mathbf{x})
\,
\nabla_{\boldsymbol{\theta}}
\log p_{\boldsymbol{\theta}}(\mathbf{y}\mid\mathbf{x})^{\top}
\right]
\right]$.
To obtain a tractable approximation of $D_{\mathrm{KL}}(Q_T | P)$ in terms of parameter updates that occur during iterative fine-tuning, we adopt a standard Laplace approximation around the pretrained solution \citep{mackay1992practical, ritter2018scalable, kirkpatrick2017overcoming}. Under this approximation, the weight-space distributions are locally modeled using the pretrained Fisher information as a common precision matrix:
$Q_t = \mathcal{N}(\boldsymbol{\theta}_t,\mathbf{F}_{\boldsymbol{\theta}_0}^{-1})$
for $t = 0,\ldots,T$, with $Q_0 = P$.
Writing $\boldsymbol{\delta}_t=\boldsymbol{\theta}_{t+1}-\boldsymbol{\theta}_t$ for the parameter update at iteration $t$, the shared-precision model gives
$D_{\mathrm{KL}}(Q_T \| P)=\tfrac12\bigl\|\sum_{t=0}^{T-1}\boldsymbol{\delta}_t\bigr\|^2_{\mathbf{F}_{\boldsymbol{\theta}_0}}$.
Applying the triangle and Cauchy-Schwarz inequalities in the Fisher norm bounds the divergence by the fine-tuning path's curvature-weighted energy:
\begin{tcolorbox}[
  manuscriptbox,
  title=\textbf{Generalization divergence governed by Fisher information}
]
\vspace{-2mm}
\begin{equation}
D_{\mathrm{KL}}(Q_T \| P)
\;\le\;
\frac{\bar\gamma\,T}{2}
\sum_{t=0}^{T-1}
\boldsymbol{\delta}_t^{\top}
\mathbf{F}_{\boldsymbol{\theta}_t}
\boldsymbol{\delta}_t,
\label{eq:kl_accumulation}
\end{equation}
where $\bar\gamma$ is the smallest constant satisfying $\boldsymbol{\delta}_t^{\top}\mathbf{F}_{\boldsymbol{\theta}_0}\boldsymbol{\delta}_t\le\bar\gamma\,\boldsymbol{\delta}_t^{\top}\mathbf{F}_{\boldsymbol{\theta}_t}\boldsymbol{\delta}_t$ for all $t$, with $\bar\gamma=1$ when the curvature along the update directions is unchanged.
\end{tcolorbox}

Equation~\ref{eq:kl_accumulation} bounds the divergence from the pretrained prior by an accumulation of curvature-weighted updates along the fine-tuning trajectory. Rather than depending solely on parameter displacement, each update is weighted by the local Fisher geometry, emphasizing changes along directions to which the model is most sensitive. Consequently, the KL term governing the PAC-Bayes bound in Eq.~\ref{eq:pac_bayes} is linked to \emph{curvature-aligned adaptation}, providing a principled basis for using Fisher information to characterize how different parameter groups contribute to fine-tuning. This motivates tracking the evolution of their Fisher structure to determine where continued adaptation remains necessary and where parameters can be progressively frozen. The full derivation and empirical validation are provided in Appendices \ref{appendix:proof_pac_bayes}, ~\ref{appendix:proof_kl_decomp} and~\ref{appendix:assumption_diagnostics}.

\subsection{Curvature Shift as an Adaptation Signal} \label{sec:curvature_shift}

To determine where continued adaptation remains necessary, we next decompose the Fisher-governed KL cost in Eq.~\ref{eq:kl_accumulation} across layers. Following standard structured approximations to the Fisher Information Matrix (FIM) \citep{martensOptimizingNeuralNetworks2015}, we use a block-diagonal approximation $\mathbf{F}_{\boldsymbol{\theta}}
\approx
\mathrm{diag}\!\left(
\mathbf{F}_{\boldsymbol{\theta}_1},
\ldots,
\mathbf{F}_{\boldsymbol{\theta}_L}
\right)$
under which the Eq.~\ref{eq:kl_accumulation} decomposes across layers as:
\begin{equation}
D_{\mathrm{KL}}(Q_T \| P)
\;\le\;
\frac{\bar\gamma\,T}{2}
\sum_{t=0}^{T-1}
\boldsymbol{\delta}_t^{\top}
\mathbf{F}_{\boldsymbol{\theta}_t}
\boldsymbol{\delta}_t
\;\approx\;
\frac{\bar\gamma\,T}{2}
\sum_{t=0}^{T-1}
\sum_{\ell=1}^{L}
c_{t,\ell},
\qquad
c_{t,\ell}
:=
\boldsymbol{\delta}_{t,\ell}^{\top}
\mathbf{F}_{\boldsymbol{\theta}_{t,\ell}}
\boldsymbol{\delta}_{t,\ell},
\label{eq:layerwise_bound}
\end{equation}
where the approximation is due solely to the block-diagonal structure. Since the FIM is positive semi-definite, $c_{t,\ell}\geq0$.
Consequently, freezing layer $\ell$ at iteration $t^*$ removes its subsequent Fisher-weighted contributions,
$
\Delta\mathcal{C}_{\ell}=
\frac{\bar\gamma\,T}{2}
\sum_{t=t^*}^{T-1}
c_{t,\ell}
\geq 0,
$
thereby preventing further accumulation of that layer's contribution to the KL bound and, consequently, to the PAC-Bayes generalization bound. However, reducing this cost alone does not justify freezing, since premature removal of a layer may suppress task-relevant adaptation. The key question is therefore \textit{which layers have adapted sufficiently such that further updates provide diminishing benefit?}
We characterize this condition through the \emph{stabilization} of a layer's Fisher-weighted contribution. Importantly, stabilization concerns the temporal evolution of the contribution rather than its absolute magnitude, where a layer may retain a nonzero cost while its role in ongoing adaptation has effectively ceased to evolve. Accordingly, when
$c_{t,\ell}-c_{t-1,\ell}\approx0$,
the layer's contribution to the Fisher-weighted adaptation cost changes only marginally across successive iterations, indicating saturation of its ongoing adaptation dynamics. We therefore define the per-layer contribution drift as:
\begin{equation}
\Delta c_{t,\ell}
:=
c_{t,\ell}-c_{t-1,\ell},
\label{eq:layer_cost_change}
\end{equation}
which quantifies the temporal evolution of each layer's contribution to the Fisher-weighted adaptation cost.
Under locally stable update magnitudes, the update-dependent component varies slowly, leaving changes in the local Fisher geometry as the dominant source of variation in $\Delta c_{t,\ell}$. This yields:
\begin{tcolorbox}[
  manuscriptbox,
  title=\textbf{Adaptation governed by FIM drift}\;]
\begin{equation}
c_{t,\ell} - c_{t-1,\ell}
\;\approx\;
\boldsymbol{\delta}_{t,\ell}^{\top}
\Delta \mathbf{F}_{t,\ell}
\boldsymbol{\delta}_{t,\ell},
\quad
\Delta \mathbf{F}_{t,\ell}
:=
\mathbf{F}_{\boldsymbol{\theta}_{t,\ell}}
-
\mathbf{F}_{\boldsymbol{\theta}_{t-1,\ell}}.
\label{eq:cost_drift}
\end{equation}
\end{tcolorbox}
Equation~\ref{eq:cost_drift} links the evolution of a layer's adaptation cost directly to temporal changes in its Fisher geometry. When the Fisher structure stabilizes, $\Delta\mathbf{F}_{t,\ell}\approx0$, the layer's contribution changes only marginally, indicating diminishing ongoing adaptation. Such layers become candidates for freezing, whereas layers exhibiting persistent Fisher drift remain actively involved in reshaping task-specific geometry and should continue to adapt (see the empirical analysis in Figures~\ref{fig:sam2_layers_js_drift} and~\ref{fig:sam2_chunks_js_drift_main}). 
The complete decomposition and underlying local stability assumptions are provided in Appendix~\ref{appendix:theory_details}. These assumptions are empirically assessed across four VTAB-1k tasks in Appendix~\ref{appendix:assumption_diagnostics}. Fisher-scale changes remain small ($0.10$-$0.17\%$), while structural Fisher drift persists, supporting the distinction between magnitude stability and structural evolution.

\subsection{Fisher-Guided Adaptive Selective Fine-Tuning Algorithm (FisherAdapTune)} \label{sec:fisher_adaptune_algorithm}
\vspace{0.5em}
\paragraph{A Scale-Robust Measure of Fisher Structural Drift.}The preceding analysis motivates the temporal variation of the FIM as a principled signal for parameter selection. To design a criterion that captures structural changes in Fisher information while remaining robust to scale, 
for each parameter group, we transform its Fisher values as
$\tilde f_i = \log_{10}(|f_i|+\varepsilon)$, and construct probability distributions
from histograms with shared bin edges for consecutive Fisher
states. We apply additive smoothing and normalize the histogram
counts, as detailed in Appendix~\ref{appendix:js-distances}.
This transformation mitigates the effect of large-magnitude entries while preserving the ordering of the Fisher entries. 
To quantify the evolution of this distribution across training iterations, we measure the divergence between consecutive Fisher-induced distributions using the Jensen-Shannon (JS) divergence. For two probability mass functions $p$ and $q$, the JS divergence is defined as:
\vspace{-1.9mm}
\begin{equation}
\mathrm{JS}(p,q)
=
\frac{1}{2}\mathrm{KL}(p \parallel M)
+
\frac{1}{2}\mathrm{KL}(q \parallel M),
\quad
M = \frac{1}{2}(p+q).
\end{equation}
And the JS distance is:
\vspace{-1.9mm}
\begin{equation}
\label{eq:js_distance}
d_{\mathrm{JS}}(p,q) = \sqrt{\mathrm{JS}(p,q)}. 
\end{equation}
which is symmetric and bounded. Appendix~\ref{appendix:js-distances} provides the computational details and characterizes its sensitivity to changes in the Fisher-value distribution and its robustness to uniform Fisher scaling.

\begin{figure}[!htbp]
    \centering
    \begin{minipage}[t]{0.52\linewidth}
        \vspace{0pt}
        \centering
        \includegraphics[width=\linewidth]{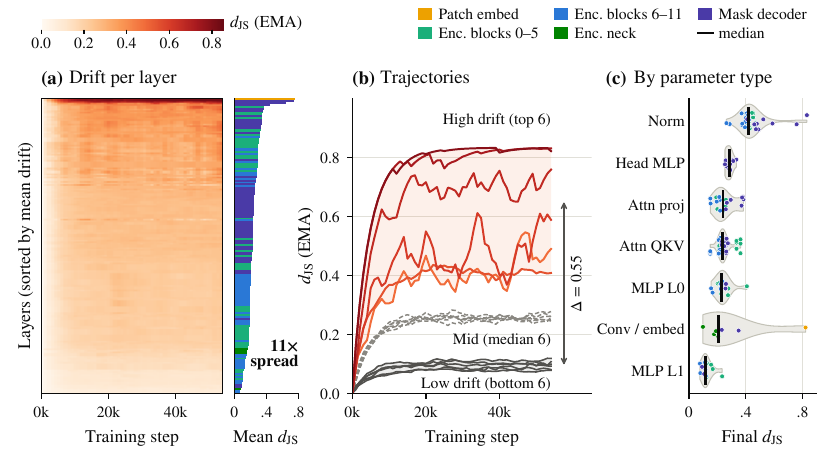}
        \caption{Layer-wise Jensen-Shannon distance between consecutive Fisher distributions in SAM2-Tiny during full fine-tuning on crack segmentation; $EMA[d_{\mathrm{JS}_{i}}^{t}(p(\tilde{\mathbf{F}}_{i}^{t-1}),p(\tilde{\mathbf{F}}_{i}^{t}))]$.}
        \label{fig:sam2_layers_js_drift}
\end{minipage}\hspace{0.015\linewidth}%
    \begin{minipage}[t]{0.45\linewidth}
        \vspace{0pt}
        \centering
\includegraphics[width=0.75\linewidth]{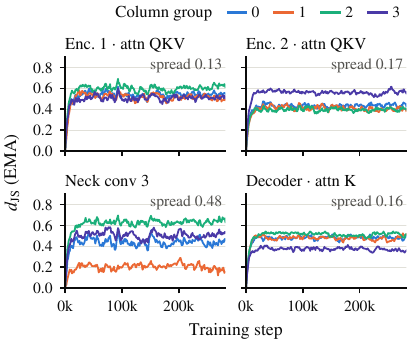}
        \caption{Intra-layer Fisher structural drift heterogeneity in SAM2-Tiny during full fine-tuning on crack segmentation.}
        \label{fig:sam2_chunks_js_drift_main}
    \end{minipage}
\end{figure}

\textbf{Empirical Analysis: Fisher structural drift is heterogeneous across layers and parameter groups.}
As shown in Appendix Figure~\ref{fig:assumption_check}, Fisher structure can continue to evolve despite locally stable Fisher scales and update magnitudes. We next examine whether this structural drift is distributed uniformly across layers and parameter groups. Using the JS distance defined above and the efficient diagonal Kronecker approximation introduced by AdaFisher \citep{gomesAdaFisherAdaptiveSecond2024}, we track Fisher structural drift during full fine-tuning. The approximation is given as:
\vspace{-5pt}
\begin{equation}
\tilde{\mathbf{F}}^{D}_\ell
=
\mathbf{H}_{D_{\ell-1}}
\otimes
\mathbf{S}_{D_\ell},
\label{eq:adafisher_fisher_form}
\end{equation}
where $\mathbf{H}_{D_{\ell-1}}$ and $\mathbf{S}_{D_\ell}$ denote diagonal activation- and gradient-correlation factors, respectively. This formulation reduces the computational and memory complexity of Fisher estimation to linear in the number of parameters. We measure $d_{\mathrm{JS}}(p_{t,\ell},p_{t-1,\ell})$ throughout full fine-tuning of SAM2 \citep{raviSAM2Segment2024} and SegFormer \citep{xieSegFormerSimpleEfficient2021} on crack segmentation.
Figure~\ref{fig:sam2_layers_js_drift} reveals evident layer-wise heterogeneity in SAM2-Tiny, with an $11\times$ difference in mean drift between high-drift and near-stationary layers. This separation emerges early ($\sim$5k steps), persists throughout training, and varies systematically across parameter types. A near-zero $d_{\mathrm{JS}}$ indicates a stable Fisher-value distribution, whereas persistent drift indicates continued structural change. 
These observations support Fisher structural drift as an operational signal of continued adaptation.
The parameter-group analysis in Figure~\ref{fig:sam2_chunks_js_drift_main}, shown in full in Appendix Figure~\ref{fig:sam2_chunks_js_drift}, further reveals substantial intra-layer heterogeneity: drift differences between groups exceed $0.5$ in some layers and $0.1$ across much of the network. Similar behavior is observed for SegFormer in Appendix Figure~\ref{fig:additional_segformer_js_drift_observation}. Collectively, these results show that \textit{Fisher structural drift is heterogeneous across layers and parameter groups}, motivating adaptive group-wise selection rather than a uniform freezing strategy.

\begin{tcolorbox}[
  manuscriptbox,
  colback=red!3!white,
  colframe=red!45!black,
  colbacktitle=red!10!white,
  coltitle=red!35!black,
  title=\textbf{Fisher Structural Drift Reveals Where Adaptation Persists}
]
Our theoretical analysis and empirical observations exhibit that Fisher structural drift is non-uniform across layers and parameter groups, indicating that only a subset of parameters remain dynamically relevant throughout fine-tuning. Therefore, stabilized groups can be frozen, while groups with persistent curvature drift remain trainable.
\end{tcolorbox}

\par\vspace{3pt}
\begin{wrapfloat}{algorithm}[24]{r}{0.58\linewidth}
\noindent\rule{\linewidth}{0.8pt}
{\captionsetup{skip=0pt,aboveskip=5pt,belowskip=0pt,labelfont=bf,textfont=bf,justification=raggedright,singlelinecheck=false}%
\caption{FisherAdapTune Algorithm}%
\label{alg:fisher_adap_tune_alg}}%
\small
\begin{algorithmic}[1]
\Require Pre-Trained Model $\theta$, dataset $\mathcal{D}$, hyperparameters: EMA factor $\beta$, threshold scale $\lambda$, block count $K$, fisher collection interval $m$, freezing interval $n$, total steps $T_{\max}$.

\For{$t = 1, \ldots, T_{\max}$}
    \State Compute $\nabla_\theta \mathcal{L}(\mathcal{D};\theta)$ via forward/backward pass

    \If{$t \bmod m = 0$}
        \State Estimate per-layer Fisher curvature $\tilde{\mathbf{F}}_{D_i}^t$ via Eq. \ref{eq:adafisher_fisher_form}
        \State Partition $\tilde{\mathbf{F}}_{D_i}^t$ into $K$ blocks $\bigl\{\tilde{\mathbf{F}}_{w_{ik}}^t\bigr\}_{k=1}^{K}$
        \State Build PMFs $p(\tilde{\mathbf{F}}_{w_{ik}}^t)$ via Appendix Eq. \ref{eq:fisher_probability}
        \State Measure curvature shift
        $d_{\mathrm{JS}_{ik}}^t(p_{t-1}, p_t)$ via Eq \ref{eq:js_distance}
        \State Smooth via EMA: $\hat{d}_{\mathrm{JS}_{ik}}^t \leftarrow \beta\,\hat{d}_{\mathrm{JS}_{ik}}^{t-1} + (1-\beta)\,d_{\mathrm{JS}_{ik}}^t$
        \State Compute mean curvature shift  $\bar{d}_{\mathrm{JS}_{ik}}$

    \EndIf

    \If{$t \bmod n = 0$}
        \State Compute global statistics of JS distances 
        \Statex \hspace{\algorithmicindent}$\mu \leftarrow \mathrm{mean}_{i,k}(\bar{d}_{\mathrm{JS}_{ik}})$;\enspace $\sigma \leftarrow \mathrm{std}_{i,k}(\bar{d}_{\mathrm{JS}_{ik}})$
        \State Set adaptive threshold $\tau \leftarrow \mu + \lambda\sigma$
        \State \textbf{Freeze} stabilized parameter groups ($\bar{d}_{\mathrm{JS}_{ik}} < \tau$)
    \EndIf
    \State \textbf{Update} only the remaining tunable parameter groups
    \Statex\hspace{\algorithmicindent} $\theta \leftarrow \theta - \eta\,\nabla_{\theta}\mathcal{L}(\mathcal{D};\theta)$

\EndFor
\end{algorithmic}
\vspace{1pt}\noindent\rule{\linewidth}{0.8pt}
\end{wrapfloat}

\noindent
\textbf{FisherAdapTune.}
Building upon the finding above, we propose \emph{FisherAdapTune}, a Fisher-guided Adaptive Fine-Tuning framework that updates only parameter groups that remain useful for adaptation. The motivation is the non-uniform \emph{Fisher structural drift} observed across layers and parameter blocks (Figure~\ref{fig:sam2_chunks_js_drift}) where some groups continue to reorient their local curvature, while others quickly stabilize. FisherAdapTune tracks this evolution with the Jensen-Shannon (JS) distance between consecutive Fisher distributions. Groups with low stable JS distance are treated as saturated and progressively frozen, where groups with persistent drift remain trainable.
All parameters are trainable at initialization, which creates an implicit warm-up phase before the first freezing decision. Fisher statistics are collected every $m$ steps and smoothed over time, while freezing is evaluated only every $n$ steps. 
This separates noisy curvature estimation from parameter selection and prevents premature freezing.
At each freezing point, FisherAdapTune applies an adaptive threshold over the smoothed JS scores, yielding a data-driven active set without manual layer schedules.
Algorithm~\ref{alg:fisher_adap_tune_alg} summarizes the procedure, and Appendix~\ref{appendix:algorithm_details} discusses additional implementation details and training overhead.

\par\noindent
\textbf{Which Parameters Does FisherAdapTune Select?}
Figure~\ref{fig:param_selection_dynamics_sam2} shows how the trainable
set evolves for SAM2-Tiny on the downstream crack segmentation task.
The budget falls from 31M to 271K parameters,  averaging 6.95M over
training. 

\par
\newsavebox{\parameterselectionfigurebox}
\sbox{\parameterselectionfigurebox}{%
    \begin{minipage}{0.39\textwidth}
        \centering
        \includegraphics[width=0.86\linewidth]{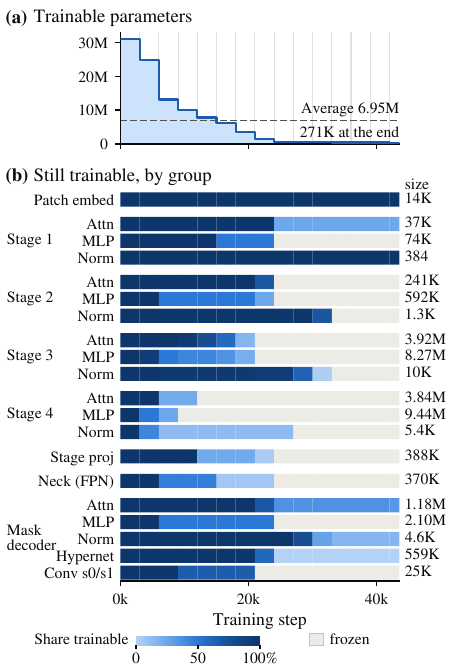}
        \captionof{figure}{Trainable parameter budget and group lifetimes for SAM2-Tiny in crack segmentation.}
        \label{fig:param_selection_dynamics_sam2}
    \end{minipage}%
}
\Needspace{\dimexpr\ht\parameterselectionfigurebox+\dp\parameterselectionfigurebox+2\intextsep+\baselineskip\relax}
\begin{wrapfigure}[20]{r}{0.39\textwidth}
    \usebox{\parameterselectionfigurebox}
\end{wrapfigure}
\noindent
 Large encoder attention and MLP groups freeze relatively
early, whereas patch embedding, several mask-decoder groups, and small
normalization groups remain trainable longer. This suggests that
trainable lifetime is not determined simply by parameter count.
Compact groups can continue adapting after much larger groups have
stabilized. The broader pattern depends on architecture and task.
DINOv2 consistently retains patch embedding and the classification
head across VTAB-1k tasks, while SegFormer-MiT-B3 sustains adaptation
in intermediate encoder groups (Appendix~\ref{appendix:param_selection}).
These patterns are consistent with the observation of
\citet{leeSurgicalFineTuningImproves2023} that useful adaptation
locations depend on the distribution shift. FisherAdapTune discovers
these differences during training without prescribing which layers
should remain trainable.

\section{Experiments \& Results}

\textbf{Experimental Setup.} We evaluate FisherAdapTune across all 19 VTAB-1k classification tasks, including the natural, specialized, and structured categories \citep{zhaiLargescaleStudyRepresentation2019}, using DINOv2 ViT-S/14. Our objective is to compare adaptation strategies under a common experimental protocol.
 We further consider crack segmentation, a challenging dense-prediction task involving thin, curvilinear structures and substantial distribution shift from pretraining data \citep{chenMindMarginalNoncrack2024a,chenDevilCrackOrientation2023}, using SAM2-Tiny, SAM2-Large, SegFormer-MiT-B0, and SegFormer-MiT-B3. Models are fine-tuned on OmniCrack30k \citep{benzOmniCrack30kBenchmarkCrack2024a} and evaluated both in-distribution and zero-shot on Concrete3k, Facade390, and Road420 \citep{liRealtimeHighresolutionNeural2023,wangAutomaticConcreteCrack2022,geFinetuningVisionFoundation2024}. We compare against full fine-tuning,
LoRA \citep{huLoRALowRankAdaptation2021}, BitFit \citep{zakenBitFitSimpleParameterefficient2022}, LayerNorm tuning \citep{zhaoTuningLayerNormAttention2023a}, decoder-only tuning, AutoFreeze \citep{liuAutoFreezeAutomaticallyFreezing2021}, Child-Tuning \citep{xuRaiseChildLarge2021}, FISH-Mask \citep{sungTrainingNeuralNetworks2021}, and random parameter selection. Full experimental details are provided in Appendix~\ref{appendix:experimental_setup}.

\textbf{Effective Number of Trainable Parameters.} Since FisherAdapTune progressively changes the active parameter set, we report the average number of parameters updated per iteration $\bar{N} = T^{-1}\sum_{t=1}^{T} N_t$,
where $N_t$ is the number of trainable parameters at iteration $t$ and $T$ is the number of training iterations. 
For static PEFT methods, the same parameters remain trainable throughout training, so $\bar{N}$ equals their fixed trainable count. For FisherAdapTune, the trainable set changes as parameter groups are progressively frozen, so $\bar{N}$ is the average number of trainable parameters across training iterations.

\par
\newsavebox{\comparativefigurebox}
\sbox{\comparativefigurebox}{%
    \begin{minipage}{0.55\linewidth}
        \centering
        \includegraphics[width=\linewidth]{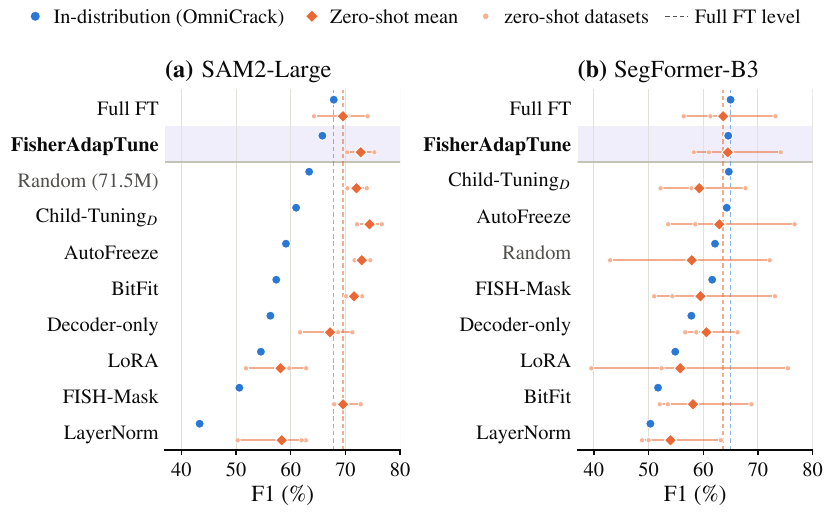}
        \captionof{figure}{Segmentation F1-scores for (a) SAM2-Large and (b) SegFormer-B3.}
        \label{fig:sam2_segformer_comparatives}
    \end{minipage}%
}
\Needspace{4\baselineskip}
\subsection{Comparative Results}
\noindent
\textbf{Image Classification.} The results in Table~\ref{tab:vtab19_peft_comparison} show that
FisherAdapTune achieves the highest macro Top-1 accuracy across 
VTAB-1k tasks ($61.62\%$), surpassing full fine-tuning ($59.97\%$)
while using a smaller effective trainable parameter budget. Unlike static PEFT methods and AutoFreeze, which follow a predefined
layer-freezing order, FisherAdapTune automatically identifies which
parameter groups to freeze based on the stabilization of their
Fisher drift.

\begin{table}[H]
\centering
\caption{%
  PEFT comparison across VTAB-1k tasks using DINOv2 ViT-S/14.
  ``Trainable'' denotes the effective number of trainable parameters. Bold and underlined values denote the best and second-best Top-1 accuracy per task, respectively.}
\label{tab:vtab19_peft_comparison}
\scriptsize
\setlength{\tabcolsep}{2pt}
\resizebox{\textwidth}{!}{%
\begin{tabular}{cc ccccccc cccc cccccccc c}
\toprule
& & \multicolumn{7}{c}{\textit{Natural}}
& \multicolumn{4}{c}{\textit{Specialized}}
& \multicolumn{8}{c}{\textit{Structured}} & \\
\cmidrule(lr){3-9} \cmidrule(lr){10-13} \cmidrule(lr){14-21}
Method & Metric
& \rotatebox{90}{CIFAR-100}
& \rotatebox{90}{Caltech-101}
& \rotatebox{90}{DTD}
& \rotatebox{90}{Flowers-102}
& \rotatebox{90}{Oxford Pets}
& \rotatebox{90}{SVHN}
& \rotatebox{90}{SUN397}
& \rotatebox{90}{EuroSAT}
& \rotatebox{90}{Patch-Cam}
& \rotatebox{90}{Resisc45}
& \rotatebox{90}{Retinopathy}
& \rotatebox{90}{Clevr-Count}
& \rotatebox{90}{Clevr-Dist}
& \rotatebox{90}{DMLab}
& \rotatebox{90}{KITTI-Dist}
& \rotatebox{90}{dSpr-Loc}
& \rotatebox{90}{dSpr-Ori}
& \rotatebox{90}{sNORB-Azim}
& \rotatebox{90}{sNORB-Elev}
& \rotatebox{90}{\shortstack[l]{Macro mean\\$\pm$ task SD}} \\
\midrule

\multirow{2}{*}{\textit{Full FT}}
& Top-1 (\%) & 56.76 & 85.15 & 62.66 & 95.89 & 86.45 & \textbf{88.88} & 38.49 & 89.08 & 84.30 & 85.14 & \textbf{76.91} & 42.86 & 46.71 & \textbf{53.59} & 47.40 & \textbf{23.47} & \textbf{31.66} & \underline{14.95} & 29.01 & $\underline{59.97_{\pm25.91}}$ \\
& Trainable (M) & 22.09 & 22.10 & 22.07 & 22.10 & 22.07 & 22.06 & 22.21 & 22.06 & 22.06 & 22.07 & 22.06 & 22.06 & 22.06 & 22.06 & 22.06 & 22.06 & 22.06 & 22.06 & 22.06 & $22.08_{\pm0.04}$ \\
\midrule

\multirow{2}{*}{\shortstack{LoRA}}
& Top-1 (\%) & \textbf{62.96} & \underline{85.88} & \textbf{75.80} & \textbf{97.54} & \textbf{92.34} & 53.77 & \textbf{46.71} & 93.91 & 82.97 & 82.37 & 75.47 & 52.73 & 50.39 & 46.44 & 53.45 & 7.25 & 26.30 & 12.66 & 27.37 & $59.28_{\pm27.49}$ \\
& Trainable (M) & 0.08 & 0.08 & 0.06 & 0.08 & 0.05 & 0.04 & 0.19 & 0.04 & 0.04 & 0.05 & 0.04 & 0.04 & 0.04 & 0.04 & 0.04 & 0.04 & 0.04 & 0.04 & 0.04 & $0.06_{\pm0.04}$ \\
\midrule

\multirow{2}{*}{\shortstack{BitFit}}
& Top-1 (\%) & 53.95 & 77.83 & 68.14 & 95.74 & 89.07 & 51.17 & 34.74 & 91.34 & 79.75 & 72.41 & 73.83 & 34.95 & 47.59 & 39.87 & \textbf{55.41} & 10.88 & 20.20 & 9.96 & 26.24 & $54.37_{\pm27.10}$ \\
& Trainable (M) & 0.09 & 0.09 & 0.07 & 0.09 & 0.07 & 0.06 & 0.20 & 0.06 & 0.05 & 0.07 & 0.05 & 0.05 & 0.05 & 0.05 & 0.05 & 0.06 & 0.06 & 0.06 & 0.05 & $0.07_{\pm0.03}$ \\
\midrule

\multirow{2}{*}{\shortstack{LayerNorm}}
& Top-1 (\%) & 54.02 & 78.96 & 68.14 & \underline{96.75} & 90.22 & 34.51 & \underline{40.71} & 87.08 & 79.00 & 69.06 & 73.58 & 28.15 & 25.13 & 34.90 & 46.69 & 6.77 & 15.81 & 6.80 & 20.77 & $50.37_{\pm29.38}$ \\
& Trainable (M) & 0.06 & 0.06 & 0.04 & 0.06 & 0.03 & 0.02 & 0.17 & 0.02 & 0.02 & 0.04 & 0.02 & 0.02 & 0.02 & 0.02 & 0.02 & 0.03 & 0.03 & 0.03 & 0.02 & $0.04_{\pm0.04}$ \\
\midrule

\multirow{2}{*}{\textit{Lin. Probe}}
& Top-1 (\%) & 45.02 & 74.60 & 65.59 & 95.46 & 88.66 & 29.63 & 38.79 & 84.55 & 78.36 & 61.78 & 73.58 & 25.27 & 24.35 & 32.81 & 38.40 & 6.52 & 15.12 & 6.52 & 16.59 & $47.45_{\pm29.21}$ \\
& Trainable (M) & 0.04 & 0.04 & 0.02 & 0.04 & 0.01 & $<$0.01 & 0.15 & $<$0.01 & $<$0.01 & 0.02 & $<$0.01 & $<$0.01 & $<$0.01 & $<$0.01 & $<$0.01 & 0.01 & 0.01 & 0.01 & $<$0.01 & $0.02_{\pm0.04}$ \\
\midrule

\multirow{2}{*}{\shortstack{AutoFreeze}}
& Top-1 (\%) & 55.28 & \textbf{88.41} & 68.94 & 95.09 & 85.94 & 84.47 & 36.91 & 94.71 & 84.32 & \textbf{85.95} & 75.32 & 39.92 & \textbf{54.66} & 45.18 & 51.48 & 16.77 & 9.94 & 13.10 & \textbf{32.21} & $58.87_{\pm28.57}$ \\
& Trainable (M) & 4.19 & 9.58 & 4.41 & 2.29 & 1.81 & 4.47 & 4.40 & 3.72 & 4.75 & 3.82 & 1.26 & 1.39 & 3.39 & 1.24 & 2.58 & 3.08 & 2.67 & 1.64 & 7.51 & $3.59_{\pm2.13}$ \\
\midrule

\multirow{2}{*}{\shortstack{Child-Tuning$_D$}}
& Top-1 (\%) & 48.86 & 84.51 & 67.50 & 95.61 & 88.09 & 76.44 & 38.61 & \underline{95.28} & \underline{85.40} & \underline{85.81} & 73.60 & 50.61 & 51.74 & 51.59 & 53.16 & 20.93 & 26.96 & 14.27 & 30.55 & $\underline{59.97_{\pm25.99}}$ \\
& Trainable (M) & 6.66 & 6.66 & 6.63 & 6.66 & 6.63 & 6.62 & 6.77 & 6.62 & 6.62 & 6.63 & 6.62 & 6.62 & 6.62 & 6.62 & 6.62 & 6.62 & 6.62 & 6.62 & 6.62 & $6.64_{\pm0.04}$ \\
\midrule

\multirow{2}{*}{\shortstack{FISH-Mask\\(2\%)}} 
  & Top-1 (\%)    & 39.37 & 75.85 & 71.01 & 95.28 & \underline{90.24} & 71.27 & 28.58 & 94.50 & 81.38 & 79.00 & 76.37 & \textbf{57.30} & \underline{53.03} & 47.25 & 53.31 & 19.71 & 18.92 & 8.51 & 24.32 & $57.12_{\pm27.58}$ \\*
  & Trainable (M) & 0.46 & 0.46 & 0.44 & 0.45 & 0.44 & 0.44 & 0.56 & 0.44 & 0.44 & 0.44 & 0.44 & 0.44 & 0.44 & 0.44 & 0.44 & 0.44 & 0.44 & 0.44 & 0.44 & $0.45_{\pm0.03}$ \\
\midrule

\rowcolor{blue!8}
& Top-1 (\%) & \underline{57.11} & 81.93 & \underline{71.65} & 96.03 & 85.20 & \underline{87.67} & 36.24 & \textbf{95.76} & \textbf{85.83} & 84.44 & \underline{76.86} & \underline{56.01} & 51.17 & \underline{52.50} & \underline{54.15} & \underline{21.28} & \underline{29.90} & \textbf{15.49} & \underline{31.59} & $\mathbf{61.62_{\pm25.87}}$ \\
\rowcolor{blue!8}
\multirow{-2}{*}{\shortstack{\textbf{FisherAdapTune}}}
& Trainable (M) & 6.25 & 9.30 & 5.62 & 6.61 & 6.63 & 7.56 & 6.02 & 6.65 & 7.22 & 6.62 & 6.57 & 5.78 & 5.75 & 6.09 & 7.08 & 9.57 & 8.85 & 7.98 & 8.93 & $7.11_{\pm1.25}$ \\

\bottomrule
\end{tabular}%
}
\end{table}

\Needspace{\dimexpr\ht\comparativefigurebox+\dp\comparativefigurebox+2\intextsep+\baselineskip\relax}
\begin{wrapfigure}[16]{r}{0.55\linewidth}
\vspace{-6pt}
    \usebox{\comparativefigurebox}
\end{wrapfigure}
\noindent
\textbf{Semantic Segmentation.} FisherAdapTune achieves a balance between in-distribution performance and zero-shot generalization. It remains competitive on the in-distribution test set while improving zero-shot performance in several settings (Figure~\ref{fig:sam2_segformer_comparatives}).
 Complete results are provided in Appendix Figure~\ref{fig:crack_segmentation_results} and Table~\ref{tab:sfinetuning_comparisons}). This suggests that its adaptive parameter selection enables effective task adaptation without sacrificing transferable representations. Its improvement over matched-budget random selection further shows that this balance depends on \emph{which} parameters are updated, rather than on parameter sparsity alone. Multi-seed results in Appendix Table~\ref{tab:crack_multiseed} further support the robustness of these findings.

\begin{wrapfigure}[12]{r}{0.53\linewidth}
\vspace{-11pt}
    \centering
    \includegraphics[width=\linewidth]{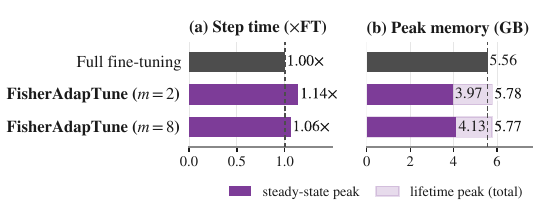}
    \caption{FisherAdapTune adds modest per-step overhead while reducing steady-state peak memory relative to full fine-tuning.}
    \label{fig:training_overhead_compact}
\end{wrapfigure}
\noindent
\textbf{Computational Cost.}
FisherAdapTune introduces a controllable runtime overhead
(Figure~\ref{fig:training_overhead_compact}). With frequent Fisher
collection ($m{=}2$), its per-step cost is $1.14\times$ that of full
fine-tuning. Increasing the collection interval to $m{=}8$ reduces this
overhead to $1.06\times$, while decreasing steady-state peak memory from
$5.56$\,GB to $4.13$\,GB ($26\%$). The lifetime peak remains comparable
to full fine-tuning since FisherAdapTune begins with an all-trainable
phase. Accuracy remains stable across the evaluated collection intervals (Section~\ref{par:hyperparam_sensitivity} and Appendix~\ref{appendix:hyperparameter_sensitivity}). Thus, collecting Fisher statistics less frequently reduces per-step cost with minimal change in accuracy across the evaluated intervals. Complete timing, memory, and baseline comparisons are provided in Appendix~\ref{appendix:overhead}.

\subsection{Ablation Studies}
\label{sec:ablations}

\par\noindent
\begin{minipage}[t]{0.45\textwidth}
\textbf{Parameter-Selection Signals.}
To further assess whether changes in Fisher structure provide a useful criterion for progressive freezing, we replace Fisher structural drift with Fisher magnitude, Fisher-magnitude drift, gradient RMS, update RMS, and random selection while keeping the remaining algorithm fixed.
Figure~\ref{fig:signal_heterogeneity_2x3} shows that Fisher structural drift remains heterogeneous across layers, distinguishing groups with ongoing adaptation from those approaching stabilization. On the other hand, gradient RMS is dominated by a few layers and declines broadly during training. As a first-order magnitude measure, it may diminish even while the approximate second-order structure continues to change.
Consistent with this distinction, Fisher structural drift achieves the strongest overall performance across eight VTAB-1k tasks (Table~\ref{tab:signal_ablation_matched} and Appendix Figure~\ref{fig:signal_ablation}). Full signal trajectories, comparisons with Fisher variants, and experimental details are provided in Appendix~\ref{appendix:signal_ablation}.
\end{minipage}\hfill
\begin{minipage}[t]{0.53\textwidth}
\vspace{-2pt}
  \centering
  \includegraphics[width=\linewidth]{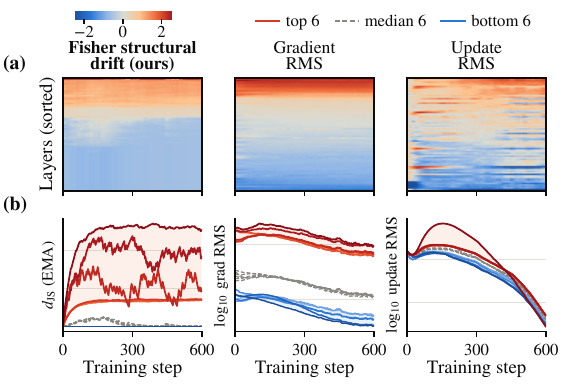}
  \captionof{figure}{(a) Layer-wise signal heatmaps and (b) representative trajectories, monitored during full fine-tuning of DINOv2 on CIFAR-100. Fisher structural drift remains heterogeneous as gradient and update RMS decay.}
\label{fig:signal_heterogeneity_2x3}
\end{minipage}

\begin{wrapfigure}[14]{r}{0.5\linewidth}
\vspace{-12pt}
    \centering
    \includegraphics[width=\linewidth]{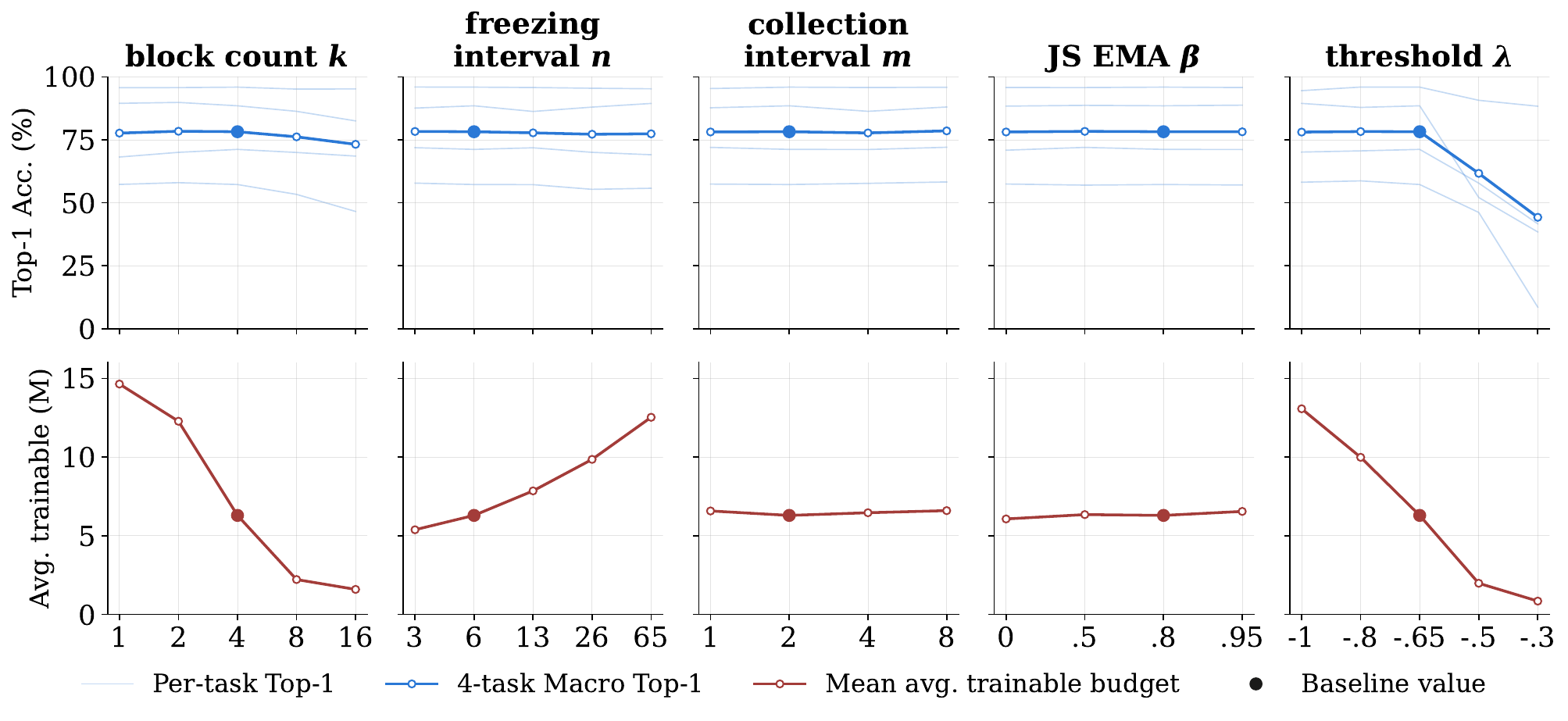}
    \caption{FisherAdapTune hyperparameter sensitivity. Top: macro Top-1 (light blue: individual tasks). Bottom: mean trainable budget. Accuracy is relatively insensitive to $n$, $m$, and $\beta$, while $\lambda$ and $K$ govern the accuracy-efficiency trade-off.}
    \label{fig:hpa_sensitivity}
\end{wrapfigure}
\begin{figure}[H]
    \centering
    \begin{subfigure}[t]{0.47\linewidth}
        \centering
        \includegraphics[width=\linewidth]{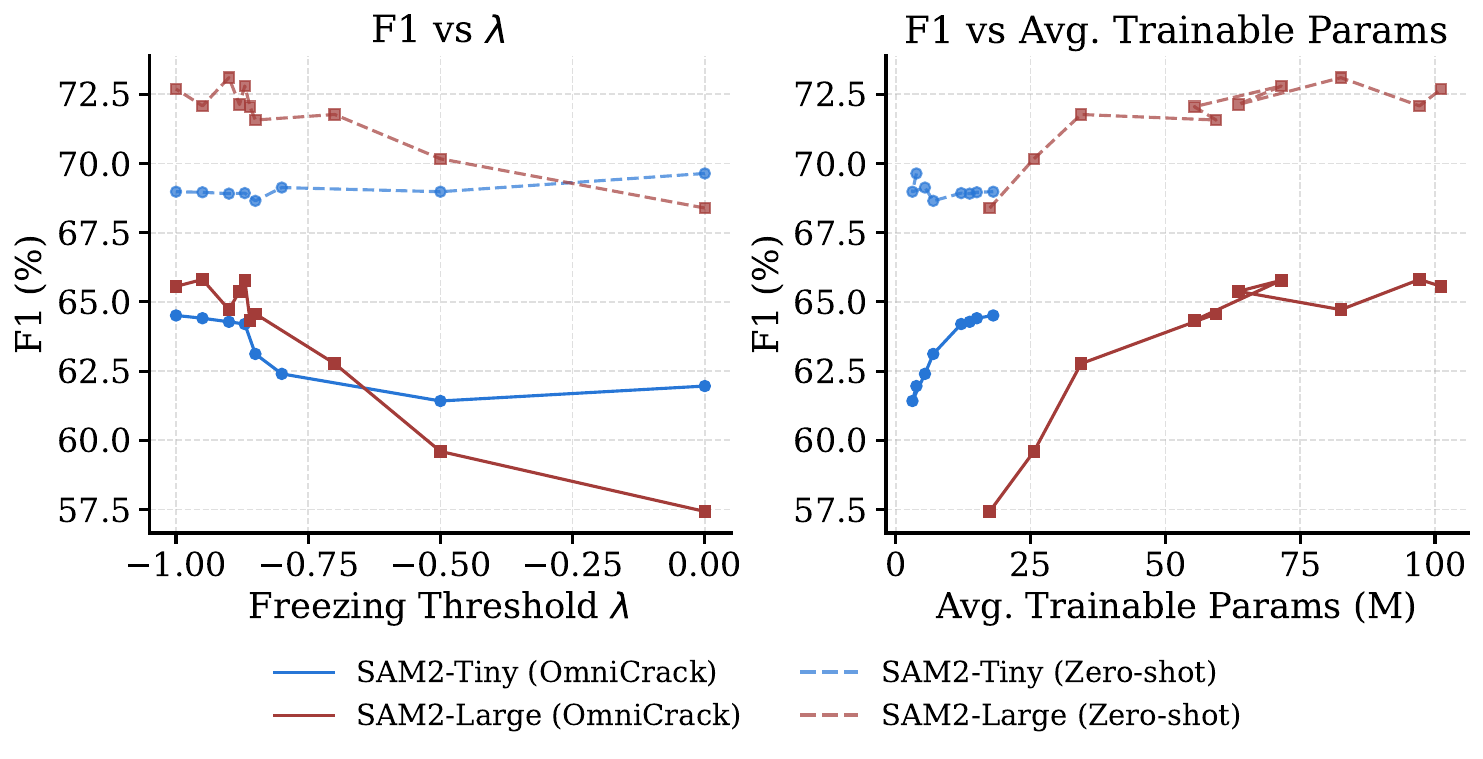}
        \caption{SAM2}
        \label{fig:sam2_ablations}
    \end{subfigure}\hspace{0.04\linewidth}%
    \begin{subfigure}[t]{0.47\linewidth}
        \centering
        \includegraphics[width=\linewidth]{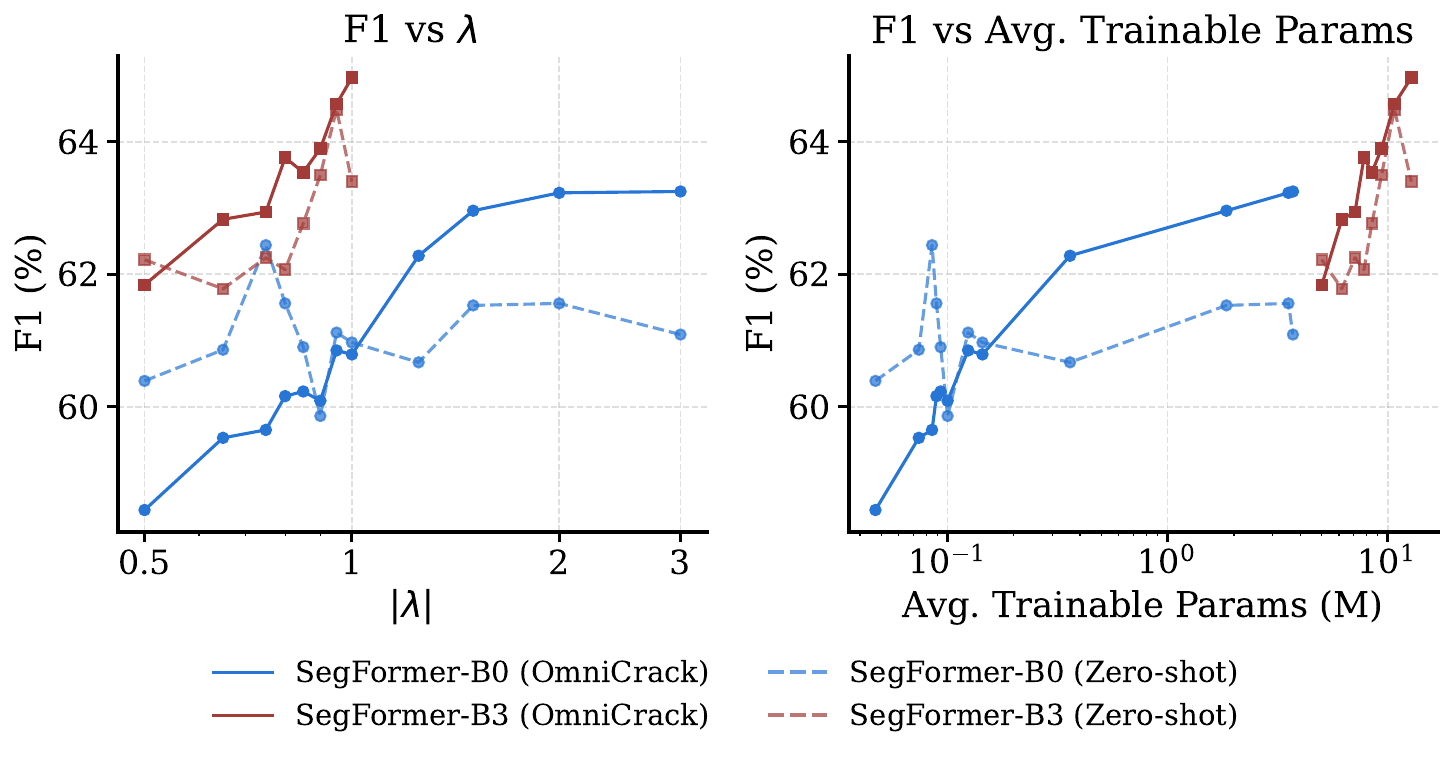}
        \caption{SegFormer}
        \label{fig:segformer_ablations}
    \end{subfigure}
    \caption{Ablation over the freezing threshold $\lambda$ in crack segmentation task. FisherAdapTune balances in-distribution adaptation and zero-shot generalization by avoiding both premature freezing and excessive source-task specialization.}
    \label{fig:lambda_ablations}
\end{figure}

\noindent
\textbf{Hyperparameter Sensitivity.} We analyze the sensitivity of FisherAdapTune to its main hyperparameters, including the parameter-group count $K$, Fisher interval $m$, freezing interval $n$, EMA factor $\beta$, and threshold scale $\lambda$. Appendix Table~\ref{tab:hparam_sensitivity} and Figure~\ref{fig:hpa_sensitivity} show stable performance across $m$, $n$, and $\beta$, with larger $m$ and $n$ reducing computational overhead as the frequency of Fisher collection and freezing decision reduces. In contrast, $K$ and $\lambda$ control the performance-efficiency trade-off, with $\lambda$ determining which adapting groups remain trainable. In additional segmentation experiments with SAM2 and SegFormer (Figure~\ref{fig:lambda_ablations}), smaller $\lambda$ values retain more groups and improve in-distribution performance, while zero-shot performance peaks at intermediate budgets. \label{par:hyperparam_sensitivity}

\FloatBarrier

\begin{table}[H]
\centering
\caption{Signal ablations on eight VTAB-1k tasks. $\lambda$ is tuned to match the trainable budget of Fisher structural drift. Trainable~(M) is the effective trainable parameter count. Bold and underlined values mark the best and second-best Top-1 accuracy per task.}
\label{tab:signal_ablation_matched}
\footnotesize
\setlength{\tabcolsep}{5pt}
\resizebox{\textwidth}{!}{%
\begin{tabular}{ccc cccccc cc c}
\toprule
& & & \multicolumn{6}{c}{\textit{Natural}} & \multicolumn{2}{c}{\textit{Specialized}} & \\
\cmidrule(lr){4-9} \cmidrule(lr){10-11}
Signal & $\lambda$ & Metric
  & CIFAR-100 & Caltech-101 & DTD & Flowers-102 & Oxford Pets & SVHN
  & EuroSAT & Patch-Cam & Macro mean $\pm$ task SD \\
\midrule
\multirow{2}{*}{Random} 
 & \multirow{2}{*}{$-1.25$}
  & Top-1 (\%)    & 23.12 & 56.22 & 59.04 & 70.11 & 68.74 & 67.24 & 90.48 & 75.80 & $63.84_{\pm19.52}$ \\
  & & Trainable (M) & 3.72 &  5.25 &  3.95 &  3.72 &  3.71 &  3.71 &  3.83 &  3.71 & $3.95_{\pm0.53}$ \\
\midrule
\multirow{2}{*}{\shortstack{Gradient\\RMS}} 
 & \multirow{2}{*}{$-0.85$}
  & Top-1 (\%)    & 49.36 & 80.07 & 65.43 & 77.22 & 21.72 & \textbf{88.96} & \underline{95.72} & \textbf{87.03} & $70.69_{\pm24.61}$ \\
  & & Trainable (M) & 7.96 &  6.88 &  6.14 &  3.58 &  2.42 & 13.46 & 10.74 & 13.29 & $8.06_{\pm4.15}$ \\
\midrule
\multirow{2}{*}{\shortstack{Update\\RMS}} 
 & \multirow{2}{*}{$-1.0$}
  & Top-1 (\%)    & \underline{52.12} & 67.96 & 66.97 & 95.35 & \underline{86.45} & 75.33 & 95.63 & 82.49 & $\underline{77.79_{\pm15.13}}$ \\
  & & Trainable (M) & 5.26 &  5.44 &  5.40 &  9.75 &  8.73 &  2.61 &  4.01 &  3.34 & $5.57_{\pm2.50}$ \\
\midrule
\multirow{2}{*}{\shortstack{Fisher\\magnitude}} 
 & \multirow{2}{*}{$-0.8$}
  & Top-1 (\%)    & 16.53 & \textbf{86.14} & \underline{67.98} & \underline{95.90} & \textbf{88.23} & 77.95 & 92.63 & 82.86 & $76.03_{\pm25.57}$ \\
  & & Trainable (M) & 1.68 & 21.56 & 21.54 & 21.56 & 21.53 &  1.80 &  1.40 &  1.44 & $11.56_{\pm10.67}$ \\
\midrule
\multirow{2}{*}{\shortstack{Fisher magnitude\\drift}} 
 & \multirow{2}{*}{$-0.2$}
  & Top-1 (\%)    & 10.03 & 28.46 & 26.17 & 16.04 & 38.13 & 26.56 & 74.56 & 77.26 & $37.15_{\pm25.35}$ \\
  & & Trainable (M) & 1.51 &  1.79 &  0.69 &  0.71 &  0.73 &  1.58 &  0.61 &  0.75 & $1.04_{\pm0.49}$ \\
\midrule
\rowcolor{blue!8}
  & & Top-1 (\%)    & \textbf{57.11} & \underline{81.93} & \textbf{71.65} & \textbf{96.03} & 85.20 & \underline{87.67} & \textbf{95.76} & \underline{85.83} & $\mathbf{82.65_{\pm12.91}}$ \\
\rowcolor{blue!8}
\multirow{-2}{*}{\shortstack{\textbf{Fisher structural}\\\textbf{drift (ours)}}} & \multirow{-2}{*}{$-0.65$}
  & Trainable (M) & 6.25 & 9.30 & 5.62 & 6.61 & 6.63 & 7.56 & 6.65 & 7.22 & $6.98_{\pm1.10}$ \\
\bottomrule
\end{tabular}
}
\end{table}

\section{Conclusion}
We introduced FisherAdapTune, a Fisher-guided adaptive fine-tuning framework that progressively freezes parameter groups as their Fisher structure stabilizes, selecting the trainable set without a prescribed layer schedule. Our PAC-Bayesian analysis relates Fisher-weighted updates to a generalization bound, motivating curvature-aware parameter selection during fine-tuning.
Across VTAB-1k classification tasks and crack segmentation, FisherAdapTune achieves the highest macro classification accuracy among the evaluated methods and competitive in-distribution and zero-shot segmentation performance while reducing the effective trainable-parameter budget. Interestingly, across different architectures and tasks, FisherAdapTune retains different combinations of input, output, and normalization groups trainable while freezing attention and MLP groups at architecture-dependent rates.
The results support Fisher drift as a task-aware signal for allocating updates. Future work will evaluate FisherAdapTune across broader architectures, datasets, and downstream tasks.

\subsection*{AI use statement}

In this work, generative AI tools were used to assist with language editing, including improving the clarity, grammar, organization, and conciseness of author-written text. They were also used to support the preparation and refinement of visualizations, including their layout, presentation, and captions. All research ideas, hypotheses, methodological choices, experimental designs, analyses, and interpretations originated from the authors. The AI tools did not independently formulate the research contributions or determine the conclusions of this work. All AI-assisted text and visual materials were carefully reviewed, revised, and verified by the authors. The authors take full responsibility for the accuracy, originality, and final content of the manuscript.

\subsection*{Reproducibility statement}

We support reproducibility by providing a complete description and pseudocode of FisherAdapTune in the main paper, together with the theoretical assumptions and derivations underlying the method. The experimental setup specifies the datasets, model architectures, adaptation baselines, evaluation metrics, and training protocol. The appendix further reports implementation details, hyperparameter settings and sensitivity analyses, Fisher-drift observations, ablation studies, and training-time and memory overhead. These materials are intended to provide the information necessary to reproduce the theoretical and empirical results reported in this work.

\bibliographystyle{plainnat}
\bibliography{references}

\clearpage
\appendix

\etocdepthtag.toc{appendix}
\etocsettagdepth{main}{none}
\etocsettagdepth{appendix}{subsection}

\etocsettocstyle{\section*{Appendix Contents}}{}
\tableofcontents

\clearpage
\section{Theoretical Details and Derivations}
\label{appendix:theoretical_details}

This appendix collects the assumptions, results, and derivations supporting the main-text theory. Appendix~\ref{appendix:proof_pac_bayes} presents the Fisher-based KL result together with its decomposition proof. Appendices~\ref{appendix:layerwise_decomposition} and~\ref{appendix:theory_details} develop the layer-wise decomposition, freezing argument, and contribution-drift derivation. Appendix~\ref{appendix:js-distances} describes the JS drift measure, and Appendix~\ref{appendix:assumption_diagnostics} collects the empirical assessments of the assumptions.

\subsection{PAC-Bayes Derivation and Fisher-Based KL Approximation}
\label{appendix:proof_pac_bayes}

\subsubsection{Local Fisher-Scale Stability}

\begin{assumption}[Local Fisher-scale stability]
\label{asm:slow_fisher}
The overall Fisher scale varies slowly during fine-tuning:
$\|\mathbf{F}_{\boldsymbol{\theta}_t}\|_F \approx
\|\mathbf{F}_{\boldsymbol{\theta}_0}\|_F$
uniformly over $t \in \{0,\ldots,T{-}1\}$.
This holds when the model is updated with a small learning rate from a
pretrained stationary point.
Crucially, this assumption constrains only the Fisher \emph{magnitude};
the normalized shape
$\mathbf{P}^t_\ell = \mathbf{F}_{\boldsymbol{\theta}_{t,\ell}}/
\|\mathbf{F}_{\boldsymbol{\theta}_{t,\ell}}\|_F$
is free to reorient across layers and iterations as the importance mass may redistribute across parameters while the layer's aggregate curvature stays constant.
\end{assumption}
\vspace{-0.7em}
The empirical assessment of Assumption~\ref{asm:slow_fisher} is provided in Appendix~\ref{appendix:assumption_diagnostics}. This condition is not required for the bound in Proposition~\ref{prop:kl_accumulation}, which handles the change from $\mathbf{F}_{\boldsymbol{\theta}_0}$ to $\mathbf{F}_{\boldsymbol{\theta}_t}$ through the constant $\bar\gamma$; its role is to establish that the overall Fisher magnitude is stable during fine-tuning, which motivates tracking changes in Fisher \emph{structure} rather than magnitude as the adaptation signal.

\subsubsection{KL Accumulation Result}

\begin{proposition}[KL accumulation along the fine-tuning trajectory]
\label{prop:kl_accumulation}
Let $Q_0 = P = \mathcal{N}(\boldsymbol{\theta}_0,\mathbf{F}_{\boldsymbol{\theta}_0}^{-1})$
denote the pretrained prior, let all posteriors share the pretrained
precision, $Q_t = \mathcal{N}(\boldsymbol{\theta}_t,\mathbf{F}_{\boldsymbol{\theta}_0}^{-1})$,
and let $\boldsymbol{\delta}_t=\boldsymbol{\theta}_{t+1}-\boldsymbol{\theta}_t$.
The following statements hold for any update rule generating
$\{\boldsymbol{\delta}_t\}_{t=0}^{T-1}$, irrespective of the optimizer and of the
form of regularization (proofs in Appendix~\ref{appendix:proof_kl_decomp}).
\vspace{-2mm}
\begin{enumerate}[itemsep=2pt,parsep=0pt]
\item \emph{Exact decomposition.}
\begin{equation}
D_{\mathrm{KL}}(Q_T \| P)
\;=\;
\sum_{t=0}^{T-1}
\!\left[
\frac{1}{2}\,
\boldsymbol{\delta}_t^{\top}
\mathbf{F}_{\boldsymbol{\theta}_0}
\boldsymbol{\delta}_t
\;+\;
\boldsymbol{\delta}_t^{\top}
\mathbf{F}_{\boldsymbol{\theta}_0}
(\boldsymbol{\theta}_t - \boldsymbol{\theta}_0)
\right],
\label{eq:kl_exact_decomp}
\end{equation}
where the cross terms collect all pairwise interactions between updates:
\begin{equation}
\sum_{t=0}^{T-1}
\boldsymbol{\delta}_t^{\top}\mathbf{F}_{\boldsymbol{\theta}_0}
(\boldsymbol{\theta}_t - \boldsymbol{\theta}_0)
\;=\;
\sum_{0\le s<t<T}
\boldsymbol{\delta}_t^{\top}\mathbf{F}_{\boldsymbol{\theta}_0}\boldsymbol{\delta}_s .
\label{eq:kl_cross_pairwise}
\end{equation}
\item \emph{Alignment factor.}
Equivalently,
$D_{\mathrm{KL}}(Q_T \| P)=\kappa_T\sum_{t}\tfrac12\,
\boldsymbol{\delta}_t^{\top}\mathbf{F}_{\boldsymbol{\theta}_0}\boldsymbol{\delta}_t$
with
\begin{equation}
\kappa_T
\;:=\;
\frac{\bigl\|\sum_{t=0}^{T-1}\boldsymbol{\delta}_t\bigr\|^2_{\mathbf{F}_{\boldsymbol{\theta}_0}}}
{\sum_{t=0}^{T-1}\|\boldsymbol{\delta}_t\|^2_{\mathbf{F}_{\boldsymbol{\theta}_0}}}
\;\in\;[0,T].
\label{eq:alignment_factor}
\end{equation}
$\kappa_T=1$ when the updates are mutually
$\mathbf{F}_{\boldsymbol{\theta}_0}$-orthogonal, $\kappa_T<1$ only when the
pairwise terms in Eq.~\ref{eq:kl_cross_pairwise} are negative on net, and
$\kappa_T=T$ when all updates coincide in the
$\mathbf{F}_{\boldsymbol{\theta}_0}$-geometry.
\item \emph{Path-energy bound.}
\begin{equation}
D_{\mathrm{KL}}(Q_T \| P)
\;\le\;
\frac12\Bigl(\sum_{t=0}^{T-1}
\|\boldsymbol{\delta}_t\|_{\mathbf{F}_{\boldsymbol{\theta}_0}}\Bigr)^{2}
\;\le\;
\frac{T}{2}\sum_{t=0}^{T-1}
\boldsymbol{\delta}_t^{\top}\mathbf{F}_{\boldsymbol{\theta}_0}\boldsymbol{\delta}_t .
\label{eq:kl_path_energy}
\end{equation}
\end{enumerate}

\begin{tcolorbox}[
  manuscriptbox,
  title=\textbf{Generalization bound governed by FIM}
]
Let $\bar\gamma$ be the smallest constant such that
$\boldsymbol{\delta}_t^{\top}\mathbf{F}_{\boldsymbol{\theta}_0}\boldsymbol{\delta}_t
\le\bar\gamma\,
\boldsymbol{\delta}_t^{\top}\mathbf{F}_{\boldsymbol{\theta}_t}\boldsymbol{\delta}_t$
for all $t$. Expressing the path energy in the local Fisher geometry gives:
\begin{equation}
D_{\mathrm{KL}}(Q_T \| P)
\;\le\;
\frac{\bar\gamma\,T}{2}
\sum_{t=0}^{T-1}
\boldsymbol{\delta}_t^{\top}
\mathbf{F}_{\boldsymbol{\theta}_t}
\boldsymbol{\delta}_t .
\label{eq:kl_accumulation_appendix}
\end{equation}
\vspace{-2mm}
\end{tcolorbox}
\end{proposition}

\begin{remark}[On the factor $T$ and the constant $\bar\gamma$]
\label{rem:kl_bound_constants}
By item~2, the unweighted quadratic sum
$\sum_t\tfrac12\boldsymbol{\delta}_t^{\top}\mathbf{F}_{\boldsymbol{\theta}_0}\boldsymbol{\delta}_t$
is itself an upper bound on $D_{\mathrm{KL}}(Q_T \| P)$ only when
$\kappa_T\le1$, i.e., when successive updates are orthogonal or anti-aligned on
net. Fine-tuning away from the pretrained initialization operates in the
opposite regime: consecutive updates are strongly aligned (mean cosine
similarity $0.87$-$0.89$, Appendix~\ref{appendix:assumption_diagnostics}), so
the pairwise terms in Eq.~\ref{eq:kl_cross_pairwise} are predominantly positive
and $\kappa_T>1$. Eq.~\ref{eq:kl_path_energy} replaces $\kappa_T$ by its largest
possible value. The factor $T$ is attained for aligned updates of equal
$\mathbf{F}_{\boldsymbol{\theta}_0}$-norm and therefore cannot be removed without
assumptions on the trajectory. Since $T$ and $\bar\gamma$ multiply every
per-step term equally, they do not affect how the bound is attributed across
layers or iterations.
The constant $\bar\gamma$ equals one when the curvature along the update
directions is unchanged from $\mathbf{F}_{\boldsymbol{\theta}_0}$.
Assumption~\ref{asm:slow_fisher} controls the overall scale of
$\mathbf{F}_{\boldsymbol{\theta}_t}$ relative to
$\mathbf{F}_{\boldsymbol{\theta}_0}$, whereas $\bar\gamma$ additionally absorbs
any reorientation of the Fisher along $\boldsymbol{\delta}_t$; it is finite
whenever $\boldsymbol{\delta}_t^{\top}\mathbf{F}_{\boldsymbol{\theta}_t}\boldsymbol{\delta}_t>0$
for every nonzero update, which holds for the damped diagonal Fisher estimates
used in practice.
The right-hand side of Eq.~\ref{eq:kl_accumulation_appendix} is therefore the
energy of the fine-tuning path measured under the evolving Fisher metric,
which upper-bounds the Mahalanobis displacement from the prior under the fixed
metric $\mathbf{F}_{\boldsymbol{\theta}_0}$ up to the factor $\bar\gamma T$; the
two quantities take the same form only when
$\mathbf{F}_{\boldsymbol{\theta}_t}=\mathbf{F}_{\boldsymbol{\theta}_0}$ along the
trajectory. All posteriors $Q_t$ retain the prior precision
$\mathbf{F}_{\boldsymbol{\theta}_0}$, so the Gaussian KL remains a pure quadratic
form, and $\mathbf{F}_{\boldsymbol{\theta}_t}$ enters only through this upper bound.
\end{remark}

\subsubsection{Proof of the KL Decomposition}
\label{appendix:proof_kl_decomp}

\begin{equation*}
D_{\mathrm{KL}}(Q_T \| P)
\;=\;
\sum_{t=0}^{T-1}
\!\left[
\frac{1}{2}\,
\boldsymbol{\delta}_t^{\top}
\mathbf{F}_{\boldsymbol{\theta}_0}
\boldsymbol{\delta}_t
\;+\;
\boldsymbol{\delta}_t^{\top}
\mathbf{F}_{\boldsymbol{\theta}_0}
(\boldsymbol{\theta}_t - \boldsymbol{\theta}_0)
\right].
\end{equation*}

\textbf{Step 1: KL between two Gaussians with the same covariance.}
Since $Q_T = \mathcal{N}(\boldsymbol{\theta}_T, \mathbf{F}_{\boldsymbol{\theta}_0}^{-1})$ and
$P = \mathcal{N}(\boldsymbol{\theta}_0, \mathbf{F}_{\boldsymbol{\theta}_0}^{-1})$ share the same covariance,
the standard KL formula for multivariate Gaussians reduces to a pure quadratic in the difference of means
(all log-determinant and trace terms cancel):
\begin{equation*}
D_{\mathrm{KL}}(Q_T \| P)
\;=\;
\frac{1}{2}
(\boldsymbol{\theta}_T - \boldsymbol{\theta}_0)^{\top}
\mathbf{F}_{\boldsymbol{\theta}_0}
(\boldsymbol{\theta}_T - \boldsymbol{\theta}_0)
\;=\;
\frac{1}{2}
\|\boldsymbol{\theta}_T - \boldsymbol{\theta}_0\|^{2}_{\mathbf{F}_{\boldsymbol{\theta}_0}}.
\end{equation*}

Where the $|\cdot|^2_{\mathbf{F}{\boldsymbol{\theta}_0}}$ is squared Mahalanobis norm weighted by the shared precision matrix. The distance measures how far the fine-tuned parameters $\boldsymbol{\theta}_T$ have drifted from the pretrained $\boldsymbol{\theta}_0$ in Fisher-geometry terms.

\textbf{Step 2: Expressing the total displacement as a telescoping sum.}
Define the per-step update $\boldsymbol{\delta}_t = \boldsymbol{\theta}_{t+1} - \boldsymbol{\theta}_t$.
Summing over all steps:
\begin{equation*}
\boldsymbol{\theta}_T - \boldsymbol{\theta}_0
\;=\;
\sum_{t=0}^{T-1} \boldsymbol{\delta}_t.
\end{equation*}
Substituting into Step~1:
\begin{equation*}
D_{\mathrm{KL}}(Q_T \| P)
\;=\;
\frac{1}{2}
\left\|\sum_{t=0}^{T-1}\boldsymbol{\delta}_t\right\|^{2}_{\mathbf{F}_{\boldsymbol{\theta}_0}}.
\end{equation*}

\textbf{Step 3: Expanding the squared norm via a telescoping identity.}
Let $\mathbf{F}_0 \equiv \mathbf{F}_{\boldsymbol{\theta}_0}$ for brevity.
Define the partial-sum vector $\mathbf{s}_t = \sum_{s=0}^{t-1}\boldsymbol{\delta}_s = \boldsymbol{\theta}_t - \boldsymbol{\theta}_0$,
with $\mathbf{s}_0 = \mathbf{0}$ and $\mathbf{s}_T = \boldsymbol{\theta}_T - \boldsymbol{\theta}_0$.
Then:
\begin{equation*}
\frac{1}{2}\|\mathbf{s}_T\|^{2}_{\mathbf{F}_0}
\;=\;
\sum_{t=0}^{T-1}
\left(
\frac{1}{2}\|\mathbf{s}_{t+1}\|^{2}_{\mathbf{F}_0}
-
\frac{1}{2}\|\mathbf{s}_{t}\|^{2}_{\mathbf{F}_0}
\right).
\end{equation*}
Since $\mathbf{s}_{t+1} = \mathbf{s}_t + \boldsymbol{\delta}_t$, each difference expands as:
\begin{align*}
\frac{1}{2}\|\mathbf{s}_t + \boldsymbol{\delta}_t\|^{2}_{\mathbf{F}_0}
-
\frac{1}{2}\|\mathbf{s}_t\|^{2}_{\mathbf{F}_0}
&\;=\;
\frac{1}{2}
\bigl(
\mathbf{s}_t^{\top}\mathbf{F}_0\mathbf{s}_t
+ 2\,\mathbf{s}_t^{\top}\mathbf{F}_0\boldsymbol{\delta}_t
+ \boldsymbol{\delta}_t^{\top}\mathbf{F}_0\boldsymbol{\delta}_t
\bigr)
-
\frac{1}{2}\mathbf{s}_t^{\top}\mathbf{F}_0\mathbf{s}_t \\
&\;=\;
\frac{1}{2}\,\boldsymbol{\delta}_t^{\top}\mathbf{F}_0\boldsymbol{\delta}_t
\;+\;
\boldsymbol{\delta}_t^{\top}\mathbf{F}_0\,\mathbf{s}_t.
\end{align*}
Substituting $\mathbf{s}_t = \boldsymbol{\theta}_t - \boldsymbol{\theta}_0$:
\begin{equation*}
\frac{1}{2}\|\mathbf{s}_{t+1}\|^{2}_{\mathbf{F}_0}
-
\frac{1}{2}\|\mathbf{s}_{t}\|^{2}_{\mathbf{F}_0}
\;=\;
\frac{1}{2}\,\boldsymbol{\delta}_t^{\top}\mathbf{F}_0\boldsymbol{\delta}_t
\;+\;
\boldsymbol{\delta}_t^{\top}\mathbf{F}_0(\boldsymbol{\theta}_t - \boldsymbol{\theta}_0).
\end{equation*}

\textbf{Step 4: Summing over all steps.}
Summing from $t=0$ to $T-1$ and using the result of Step~1:
\begin{equation*}
D_{\mathrm{KL}}(Q_T \| P)
\;=\;
\frac{1}{2}\|\mathbf{s}_T\|^{2}_{\mathbf{F}_0}
\;=\;
\sum_{t=0}^{T-1}
\!\left[
\frac{1}{2}\,\boldsymbol{\delta}_t^{\top}\mathbf{F}_0\boldsymbol{\delta}_t
\;+\;
\boldsymbol{\delta}_t^{\top}\mathbf{F}_0(\boldsymbol{\theta}_t - \boldsymbol{\theta}_0)
\right],
\end{equation*}
which gives Eq.~\ref{eq:kl_exact_decomp}. 

\textbf{Step 5: Pairwise form of the cross terms and the alignment factor.}
Substituting $\boldsymbol{\theta}_t-\boldsymbol{\theta}_0=\sum_{s=0}^{t-1}\boldsymbol{\delta}_s$
into each cross term gives
$\boldsymbol{\delta}_t^{\top}\mathbf{F}_0(\boldsymbol{\theta}_t-\boldsymbol{\theta}_0)
=\sum_{s<t}\boldsymbol{\delta}_t^{\top}\mathbf{F}_0\boldsymbol{\delta}_s$,
and summing over $t$ yields Eq.~\ref{eq:kl_cross_pairwise}. Dividing the result
of Step~2 by $\sum_t\tfrac12\|\boldsymbol{\delta}_t\|^2_{\mathbf{F}_0}$ gives
Eq.~\ref{eq:alignment_factor}; the range $\kappa_T\in[0,T]$ follows from
positive semi-definiteness and from Step~6.

\textbf{Step 6: Path-energy bound.}
Since $\mathbf{F}_0$ is positive semi-definite, $\|\cdot\|_{\mathbf{F}_0}$ is a
seminorm and satisfies the triangle inequality:
\begin{equation*}
\Bigl\|\sum_{t=0}^{T-1}\boldsymbol{\delta}_t\Bigr\|_{\mathbf{F}_0}
\;\le\;
\sum_{t=0}^{T-1}\|\boldsymbol{\delta}_t\|_{\mathbf{F}_0},
\end{equation*}
with equality when the vectors $\mathbf{F}_0^{1/2}\boldsymbol{\delta}_t$ are
non-negative multiples of a common direction. Squaring and applying the
Cauchy--Schwarz inequality,
$\bigl(\sum_{t=0}^{T-1}a_t\bigr)^2\le T\sum_{t=0}^{T-1}a_t^2$ with
$a_t=\|\boldsymbol{\delta}_t\|_{\mathbf{F}_0}$ (equality when all $a_t$ are
equal), gives Eq.~\ref{eq:kl_path_energy} after multiplying by $\tfrac12$ and
using Step~2.

\textbf{Step 7: Local Fisher geometry.}
By the definition of $\bar\gamma$,
$\boldsymbol{\delta}_t^{\top}\mathbf{F}_0\boldsymbol{\delta}_t
\le\bar\gamma\,\boldsymbol{\delta}_t^{\top}\mathbf{F}_{\boldsymbol{\theta}_t}\boldsymbol{\delta}_t$
for every $t$. Substituting into the right-hand side of
Eq.~\ref{eq:kl_path_energy} gives Eq.~\ref{eq:kl_accumulation_appendix}.
None of the steps uses the form of the update rule, so the bound holds for
adaptive optimizers and for coupled or decoupled weight decay alike.

\subsection{Layer-Wise KL Decomposition and Freezing}
\label{appendix:layerwise_decomposition}

 We decompose the Fisher-governed KL cost across layers and use the resulting curvature shifts as a proxy for adaptation relevance.
Following standard structured approximations to the Fisher Information
Matrix \citep{martensOptimizingNeuralNetworks2015}, we use a
block-diagonal decomposition across $L$ layers:
\vspace{-1.2em}
\begin{equation}
\mathbf{F}_{\boldsymbol{\theta}}
\approx
\mathrm{diag}\!\left(
\mathbf{F}_{\boldsymbol{\theta}_1},
\ldots,
\mathbf{F}_{\boldsymbol{\theta}_L}
\right).
\label{eq:block_diag_fisher_appendix}
\end{equation}
This approximation enables the Fisher-weighted contribution to be
decomposed layer-wise and is used only for tractable attribution of the
global curvature cost.

\begin{definition}[Per-layer KL cost]
\label{def:layer_cost}
The per-iteration per-layer contribution to the KL bound in
Eq.~\ref{eq:kl_accumulation_appendix} is
$c_{t,\ell} := \boldsymbol{\delta}_{t,\ell}^{\top}
\mathbf{F}_{\boldsymbol{\theta}_{t,\ell}}
\boldsymbol{\delta}_{t,\ell}$,
and its cumulative contribution is
$\mathcal{C}_\ell := \sum_{t=0}^{T-1} c_{t,\ell}$.
\end{definition}

\begin{proposition}[Layer-wise decomposition of the KL bound]
\label{prop:layerwise}
Under the layer-wise Fisher approximation in
Eq.~\ref{eq:block_diag_fisher_appendix}, the right-hand side of
Eq.~\ref{eq:kl_accumulation_appendix} decomposes as:
\vspace{-2mm}
\begin{equation}
D_{\mathrm{KL}}(Q_T \| P)
\;\le\;
\frac{\bar\gamma\,T}{2}
\sum_{t=0}^{T-1}
\boldsymbol{\delta}_t^{\top}
\mathbf{F}_{\boldsymbol{\theta}_t}
\boldsymbol{\delta}_t
\;\approx\;
\frac{\bar\gamma\,T}{2}
\sum_{\ell=1}^{L} \mathcal{C}_\ell,
\label{eq:layerwise_bound_appendix}
\end{equation}
where the approximation is due solely to the block-diagonal structure.
Since each $c_{t,\ell} = \boldsymbol{\delta}_{t,\ell}^{\top}
\mathbf{F}_{\boldsymbol{\theta}_{t,\ell}}\boldsymbol{\delta}_{t,\ell} \geq 0$
(a quadratic form under a positive semi-definite matrix), every term
contributes non-negatively to the bound.
\end{proposition}

\begin{corollary}[Freezing stabilized layers reduces the KL bound]
\label{cor:freezing}
If layer $\ell$ is frozen at iteration $t^*$, then
$\boldsymbol{\delta}_{t,\ell}=\mathbf{0}$ and hence $c_{t,\ell}=0$ for all
$t\ge t^*$. Holding the updates of the remaining layers fixed, the layer-wise
bound in Eq.~\ref{eq:layerwise_bound_appendix} is reduced by
$\Delta \mathcal{C}_\ell = \tfrac{\bar\gamma T}{2}\sum_{t=t^*}^{T-1} c_{t,\ell} \geq 0$,
since each $c_{t,\ell} \geq 0$.
\end{corollary}

\subsection{Curvature Shift as an Adaptation Signal}
\label{appendix:theory_details}

Corollary~\ref{cor:freezing} shows that freezing a layer removes a
non-negative contribution to the Fisher-weighted upper bound on KL
divergence, holding the remaining updates fixed. This provides a
curvature-based incentive for freezing, however, freezing may also suppress
task adaptation. Therefore, the key question is \emph{which layers can
be frozen while preserving essential adaptation dynamics}. We posit
that layers whose contributions have effectively \emph{stabilized} are
promising candidates. Importantly, stabilization concerns the
\emph{evolution} of the contribution rather than its absolute value.
If a layer's contribution changes only marginally across iterations,
$c_{t,\ell}-c_{t-1,\ell}\approx 0$, its curvature-weighted role is
becoming more stable, motivating a closer examination of whether
continued updates are needed. To characterize this behavior, we define
the per-layer contribution drift:
\vspace{-1mm}
\[
c_{t,\ell} - c_{t-1,\ell}
=
\boldsymbol{\delta}_{t,\ell}^{\top}
\mathbf{F}_{\boldsymbol{\theta}_{t,\ell}}
\boldsymbol{\delta}_{t,\ell}
-
\boldsymbol{\delta}_{t-1,\ell}^{\top}
\mathbf{F}_{\boldsymbol{\theta}_{t-1,\ell}}
\boldsymbol{\delta}_{t-1,\ell}.
\]

\subsubsection{Local Update-Magnitude Stability}

\begin{assumption}[Local update-magnitude stability]
\label{asm:smooth_updates}
Consecutive layer-wise updates have comparable magnitude in the local
Fisher metric, $\|\boldsymbol{\delta}_{t,\ell}\|_{\mathbf{F}_{\boldsymbol{\theta}_{t-1,\ell}}}
\approx
\|\boldsymbol{\delta}_{t-1,\ell}\|_{\mathbf{F}_{\boldsymbol{\theta}_{t-1,\ell}}},$
where $|\mathbf{v}|^2_{\mathbf{A}} := \mathbf{v}^{\top}\mathbf{A}\mathbf{v}$.
Such local magnitude stability can arise during fine-tuning with small
learning rates and adaptive or momentum-based optimizers, for which the
scale of parameter updates may vary slowly across consecutive iterations.
Importantly, comparable update magnitudes do not imply that adaptation has
converged, rather the update direction and the local geometry experienced by the
parameters may continue to evolve.
\end{assumption}

Distinguishing magnitude stability from convergence is important as a layer may continue receiving updates of similar magnitude while responding to an evolving task-specific geometry. Assumption~\ref{asm:smooth_updates} therefore constrains only the \emph{magnitude} of consecutive updates, and it does not require $\boldsymbol{\delta}_{t,\ell}$ and $\boldsymbol{\delta}_{t-1,\ell}$ to be close as vectors.
The empirical assessment of Assumption~\ref{asm:smooth_updates} is provided in Appendix~\ref{appendix:assumption_diagnostics}.

\subsubsection{Derivation of the Contribution Drift}

Under this local magnitude-stability condition, the temporal change in the per-layer contribution decomposes as:
\vspace{-0.7em}
\[
c_{t,\ell} - c_{t-1,\ell}
\;=\;
\underbrace{
\boldsymbol{\delta}_{t,\ell}^{\top}
\Delta \mathbf{F}_{t,\ell}
\boldsymbol{\delta}_{t,\ell}
}_{\text{curvature shift}}
\;+\;
\underbrace{
\Bigl(
\|\boldsymbol{\delta}_{t,\ell}\|^2_{\mathbf{F}_{\boldsymbol{\theta}_{t-1,\ell}}}
-
\|\boldsymbol{\delta}_{t-1,\ell}\|^2_{\mathbf{F}_{\boldsymbol{\theta}_{t-1,\ell}}}
\Bigr)
}_{\text{magnitude term}
}
\]
where $\Delta \mathbf{F}_{t,\ell}
:= \mathbf{F}_{\boldsymbol{\theta}_{t,\ell}}
- \mathbf{F}_{\boldsymbol{\theta}_{t-1,\ell}}$. Therefore, under Assumption ~\ref{asm:smooth_updates}:

\begin{tcolorbox}[
  manuscriptbox,
  title=\textbf{Adaptation governed by FIM drift}\;]
\begin{equation}
c_{t,\ell} - c_{t-1,\ell}
\;\approx\;
\boldsymbol{\delta}_{t,\ell}^{\top}
\Delta \mathbf{F}_{t,\ell}
\boldsymbol{\delta}_{t,\ell},
\quad
\Delta \mathbf{F}_{t,\ell}
:=
\mathbf{F}_{\boldsymbol{\theta}_{t,\ell}}
-
\mathbf{F}_{\boldsymbol{\theta}_{t-1,\ell}}.
\label{eq:cost_drift_appendix}
\end{equation}
\end{tcolorbox}

Equation~\ref{eq:cost_drift_appendix} shows that, under local
update-magnitude stability, the evolution of a layer's Fisher-weighted
contribution is approximately governed by the \textit{temporal variation
of its Fisher information}. When the Fisher stabilizes,
$\mathbf{F}_{\boldsymbol{\theta}_{t,\ell}}
\approx
\mathbf{F}_{\boldsymbol{\theta}_{t-1,\ell}}$,
the contribution drift $c_{t,\ell}-c_{t-1,\ell}$ becomes small, even
though the contribution itself may remain nonzero. This makes the layer a candidate for freezing: its curvature-weighted role changes only marginally, while continued updates still incur divergence cost.
Conversely, persistent Fisher drift suggests that a layer's role in
adaptation is still evolving and motivates keeping it trainable.

\subsection{Quantifying Fisher Structural Drift with Jensen-Shannon Distance}
\label{appendix:js-distances}

To compare Fisher structural drift across layers and training stages, for each layer $\ell$, we concatenate the Fisher values of its weights and biases into a single vector:
\begin{equation}
F_\ell=\{f_i\}_{i=1}^{N_\ell}.
\end{equation}
To stabilize heavy-tailed magnitudes, we first transform their magnitudes to log scale:
\begin{equation}
\label{eq:fisher_log}
\tilde{F}_\ell=\log_{10}(|F_\ell|+\varepsilon),
\qquad \varepsilon=10^{-12}.
\end{equation}

To compare two consecutive Fisher states, we construct histograms using a shared set of $B$ bins. The bin edges $\{b_k\}_{k=0}^{B}$ are determined from the joint 1st--99th percentile range of the two log-transformed vectors, ensuring aligned histogram support. The count in each bin is:
\begin{equation}
c_k
=
\#\left\{
i \;\middle|\;
b_k\leq\tilde f_i<b_{k+1}
\right\},
\qquad k=0,\ldots,B-1,
\end{equation}
where $\tilde f_i\in\tilde F_\ell$. To avoid zero-probability bins and ensure numerical stability in divergence computations, we apply additive smoothing and normalize:
\begin{equation}
p_k
=
\frac{c_k+\varepsilon}
{\sum_{j=0}^{B-1}(c_j+\varepsilon)},
\qquad
\sum_{k=0}^{B-1}p_k=1.
\end{equation}
The resulting probability mass function (PMF) for layer $\ell$ is:
\begin{equation}
\label{eq:fisher_probability}
p_\ell=(p_0,\ldots,p_{B-1})\in\mathbb{R}^B.
\end{equation}

Distributional differences are then measured using the Jensen-Shannon (JS) distance. The JS divergence between two PMFs $p$ and $q$ is defined as:
\begin{equation}
\mathrm{JS}(p,q)
=
\frac{1}{2}\mathrm{KL}(p\parallel m)
+
\frac{1}{2}\mathrm{KL}(q\parallel m),
\qquad
m=\frac{1}{2}(p+q),
\end{equation}
with the corresponding JS distance:
\begin{equation}
\label{eq:js_distance_appendix}
d_{\mathrm{JS}}(p,q)=\sqrt{\mathrm{JS}(p,q)}.
\end{equation}

This symmetric and bounded distance measures redistribution in the Fisher-value distribution during fine-tuning. Since $p_{t,\ell}$ is a binned, one-dimensional summary of the diagonal Fisher entries, $d_{\mathrm{JS}}$ is a proxy for change in the Fisher tensor rather than an exact test of tensor equality. The following proposition formalizes the information captured, and omitted, by this proxy. Let $w=(b_B-b_0)/B$ denote the bin width on the $\log_{10}$ scale, and for two PMFs $p,q$ let
\[
\Delta(p,q)=\sum_{k=0}^{B-1}\frac{(p_k-q_k)^2}{p_k+q_k}
\]
denote their triangular discrimination.

\begin{proposition}[JS distance as a proxy for Fisher structural drift]
\label{prop:js_structural}
Let $p_{t,\ell}$ and $p_{t-1,\ell}$ be the PMFs of Eq.~\ref{eq:fisher_probability} constructed from $\mathbf{F}^{t}_\ell$ and $\mathbf{F}^{t-1}_\ell$ on the shared bin grid $\{b_k\}_{k=0}^{B}$. The JS distance $d_{\mathrm{JS}}(p_{t,\ell},p_{t-1,\ell})$ satisfies:
\vspace{-2mm}
\begin{enumerate}[itemsep=2pt,parsep=0pt]
\item \emph{Distributional identity.}
  $d_{\mathrm{JS}}$ is a metric on PMFs, hence
  $d_{\mathrm{JS}}(p_{t,\ell},p_{t-1,\ell})=0$ if and only if
  $p_{t,\ell}=p_{t-1,\ell}$. In particular, $d_{\mathrm{JS}}=0$ whenever
  $\mathbf{F}^{t}_\ell=\mathbf{F}^{t-1}_\ell$. The converse does not hold:
  $d_{\mathrm{JS}}$ is unchanged by any permutation of the entries within the
  group and by any perturbation that moves no entry across a bin edge.
  Thus $d_{\mathrm{JS}}>0$ certifies that the distribution of
  log-sensitivities has changed, whereas $d_{\mathrm{JS}}\approx 0$ is a
  necessary, but not sufficient, condition for a stationary Fisher tensor.
\item \emph{Bounded sensitivity.}
  Let $m$ be the number of entries whose bin assignment differs between the
  two states (counting entries that enter or leave $[b_0,b_B]$), let
  $n_t,n_{t-1}$ be the in-range counts, and set
  $\rho=m/\max(n_t,n_{t-1})$. Then:
  \begin{equation}
  \label{eq:js_sensitivity_bound}
  d_{\mathrm{JS}}^2(p_{t,\ell},p_{t-1,\ell})
  \;\le\;
  \frac{\ln 2}{2}\,\Delta(p_{t,\ell},p_{t-1,\ell})
  \;\le\;
  \ln 2\cdot\min(1,\rho).
  \end{equation}
\item \emph{Approximate scale invariance.}
  Suppose the normalized shape tensor
  $\mathbf{P}^t_\ell=\mathbf{F}^t_\ell/\|\mathbf{F}^t_\ell\|_F$ is unchanged,
  i.e., $\mathbf{F}^t_\ell=\alpha\,\mathbf{F}^{t-1}_\ell$ for some $\alpha>0$,
  and let $\delta=|\log_{10}\alpha|$. Every log-transformed entry then shifts
  by at most $\delta$, so only entries lying within $\delta$ of one of the
  $B+1$ bin edges can change bins, and Eq.~\ref{eq:js_sensitivity_bound} holds
  with
  $m\le\#\{i:\min_k|\tilde f^{\,t-1}_i-b_k|\le\delta\}$.
  Hence $d_{\mathrm{JS}}$ is not exactly invariant to uniform rescaling, but
  the scale-induced component is controlled by the ratio $\delta/w$ of the
  log-scale shift to the bin width.
\end{enumerate}
Consequently, a near-zero JS distance indicates a stable Fisher-value distribution up to within-bin and permutation ambiguity, whereas persistent JS distance that exceeds the scale-induced component of item~3 indicates continued redistribution of curvature. Under the local stability conditions underlying Eq.~\ref{eq:cost_drift_appendix}, this distance provides an operational proxy for whether a layer or parameter group continues to contribute to adaptation.
\end{proposition}

\begin{proof}
\emph{Item 1.} The square root of the JS divergence is a metric on probability distributions \citep{endresNewMetricProbability2003}, which gives the equivalence with $p_{t,\ell}=p_{t-1,\ell}$. The counts $c_k$ depend on $\mathbf{F}_\ell$ only through the multiset $\{\tilde f_i\}$ and only through the bin containing each $\tilde f_i$, which gives the two invariances.

\par\smallskip
\noindent\emph{Item 2.} The first inequality is the upper bound of \citet{topsoeInequalitiesInformationDivergence2000}, $2\,\mathrm{JS}(p,q)\le\ln 2\cdot\Delta(p,q)$, with $\mathrm{JS}$ in nats. For the second, $|p_k-q_k|\le p_k+q_k$ gives $\Delta(p,q)\le\|p-q\|_1\le 2$. Write $p_k=(c_k+\varepsilon)/Z$ and $q_k=(c'_k+\varepsilon)/Z'$ with $Z=n+B\varepsilon$, $Z'=n'+B\varepsilon$, and label the two states so that $Z\ge Z'$. From
$p_k-q_k=(c_k-c'_k)/Z+(c'_k+\varepsilon)(1/Z-1/Z')$ we obtain
\begin{equation}
\|p-q\|_1\;\le\;\frac{\sum_k|c_k-c'_k|+|n-n'|}{Z}.
\end{equation}
An entry that moves between two in-range bins contributes $2$ to $\sum_k|c_k-c'_k|$ and $0$ to $|n-n'|$; an entry that enters or leaves $[b_0,b_B]$ contributes at most $1$ to each. The numerator is therefore at most $2m$, so $\|p-q\|_1\le 2m/Z\le 2\rho$, and $\tfrac{\ln 2}{2}\Delta\le\ln 2\cdot\min(1,\rho)$.

\par\smallskip
\noindent\emph{Item 3.} $\mathbf{P}^t_\ell=\mathbf{P}^{t-1}_\ell$ holds if and only if $\mathbf{F}^t_\ell=\alpha\mathbf{F}^{t-1}_\ell$ with $\alpha=\|\mathbf{F}^t_\ell\|_F/\|\mathbf{F}^{t-1}_\ell\|_F$. For $\alpha\ge 1$ and any $f_i$,
$0\le\log_{10}(\alpha|f_i|+\varepsilon)-\log_{10}(|f_i|+\varepsilon)\le\log_{10}\alpha$,
and symmetrically for $\alpha<1$, so each $\tilde f_i$ moves by at most $\delta$. An entry changes its bin assignment only if a bin edge lies between its old and new position, which requires the entry to be within $\delta$ of that edge. The result then follows directly from item~2.
\end{proof}

\subsection{Empirical Assessment of the Theoretical Assumptions}
\label{appendix:assumption_diagnostics}

\begin{figure}
    \centering
    \includegraphics[width=\linewidth]{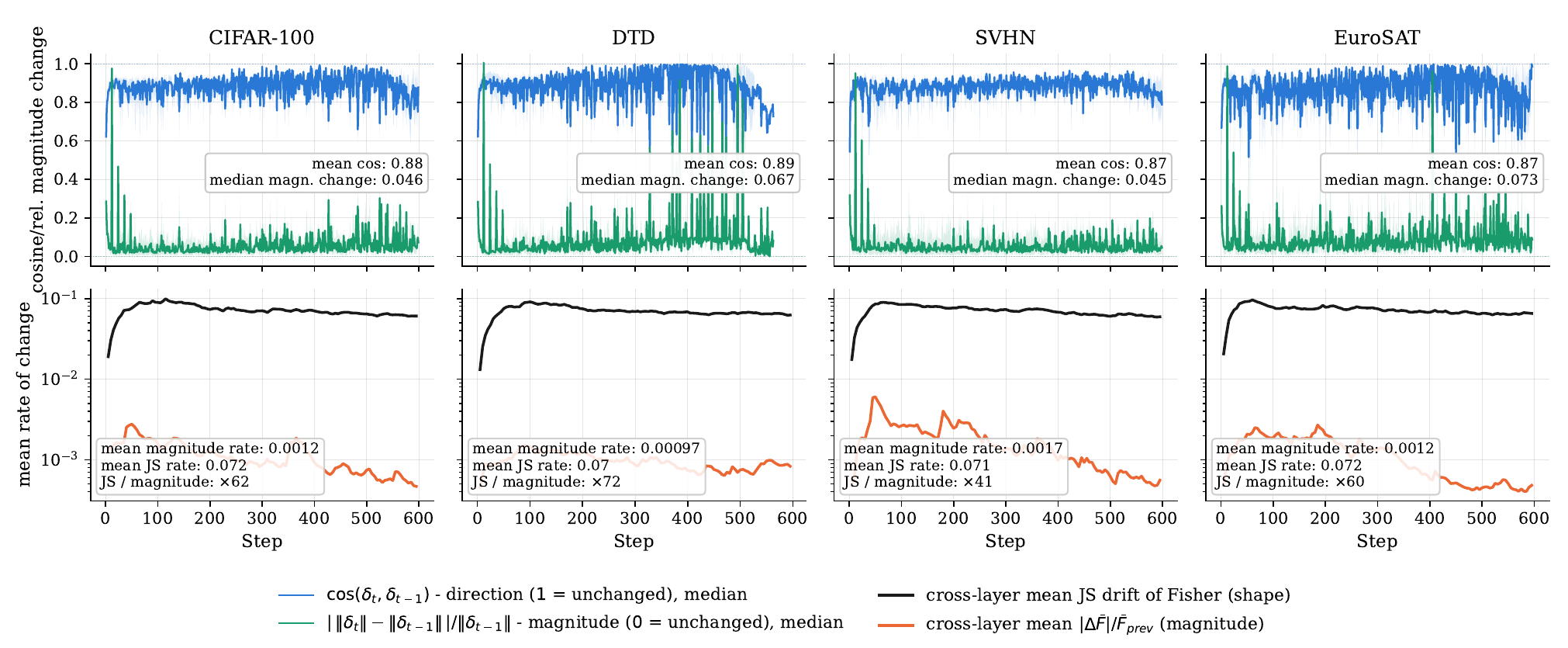}
    \caption{Empirical validation of Assumptions~\ref{asm:slow_fisher} and~\ref{asm:smooth_updates} during full fine-tuning of DINOv2 ViT-S/14 on four VTAB-1k tasks. \emph{Top:} relative change in update magnitude and cosine similarity between consecutive updates (median and 10-90th percentiles across parameter tensors). Update magnitude remains stable (median $0.045$-$0.073$), while showing modest directional variation. \emph{Bottom:} relative Fisher-scale change and structural Fisher drift. Fisher scale varies by only $0.10$-$0.17\%$, while structural drift is $41$-$72\times$ larger. See Appendix~\ref{appendix:assumption_diagnostics} for more details.}
    \label{fig:assumption_check}
\end{figure}

The experimental protocol and metric definitions for the empirical assessment of Assumptions~\ref{asm:slow_fisher} and~\ref{asm:smooth_updates} are reported in Figure~\ref{fig:assumption_check}. We analyze one full fine-tuning trajectory for each of CIFAR-100, DTD, SVHN, and EuroSAT from VTAB-1k. For each task, a pretrained DINOv2 ViT-S/14 is fine-tuned for 50 epochs (600 optimizer steps) using AdamW with batch size 64, a peak learning rate of $5\times10^{-5}$, linear warmup followed by cosine decay without weight decay. Update and Fisher diagnostics are recorded from the same training trajectory.

\paragraph{Update-magnitude stability.}
We empirically assess this behavior in Figure~\ref{fig:assumption_check}. As a tractable surrogate for variation in the Fisher-weighted update magnitude, we measure the relative change in Euclidean update norm:
$
\frac{
\bigl|
\|\boldsymbol{\delta}_{t,\ell}\|_2
-
\|\boldsymbol{\delta}_{t-1,\ell}\|_2
\bigr|
}{
\|\boldsymbol{\delta}_{t-1,\ell}\|_2
}.
$
Across four VTAB-1k tasks, its median ranges from $0.045$ to $0.073$, indicating gradually varying update magnitudes. However, consecutive updates are not directionally identical, with mean cosine similarities of $0.87$-$0.89$. Update-magnitude stability alone therefore does not imply that adaptation has stabilized, motivating a signal that captures changes in task-specific geometry.

For parameter tensor $\ell$ we define the update at iteration $t$ as
$\boldsymbol{\delta}_{t,\ell} = \boldsymbol{\theta}_{t+1,\ell} - \boldsymbol{\theta}_{t,\ell}$
and report the relative change in its magnitude:
\begin{equation}
\rho_{t,\ell}
\;:=\;
\frac{\bigl|\,\|\boldsymbol{\delta}_{t,\ell}\|_2
      - \|\boldsymbol{\delta}_{t-1,\ell}\|_2\,\bigr|}
     {\|\boldsymbol{\delta}_{t-1,\ell}\|_2}
\;=\;
|r_{t,\ell}-1|,
\qquad
r_{t,\ell} \;:=\;
\frac{\|\boldsymbol{\delta}_{t,\ell}\|_2}{\|\boldsymbol{\delta}_{t-1,\ell}\|_2},
\end{equation}
where the ratio $r_{t,\ell}$ is retained as shorthand for use below.
The quantity $\rho_{t,\ell}$ is the Euclidean surrogate for the Fisher-metric
ratio constrained by Assumption~\ref{asm:smooth_updates}. 
Across the four tasks the median of
$\rho_{t,\ell}$ is $0.046$ (CIFAR-100), $0.067$ (DTD), $0.045$ (SVHN) and
$0.073$ (EuroSAT), with 90th percentiles of $0.13$-$0.19$ and no upward trend
over training.

We additionally report the directional diagnostic
$s_{t,\ell} = \boldsymbol{\delta}_{t,\ell}^{\top}\boldsymbol{\delta}_{t-1,\ell}
/(\|\boldsymbol{\delta}_{t,\ell}\|_2\|\boldsymbol{\delta}_{t-1,\ell}\|_2)$,
which their mean is $0.87$-$0.89$, corresponding to a median rotation of
$24^\circ$-$28^\circ$ per optimizer step. The two diagnostics combine through the
exact identity:
\begin{equation}
\frac{\|\boldsymbol{\delta}_{t,\ell}-\boldsymbol{\delta}_{t-1,\ell}\|^2_2}
     {\|\boldsymbol{\delta}_{t-1,\ell}\|^2_2}
\;=\;
\underbrace{\rho_{t,\ell}^2}_{\text{radial}}
\;+\;
\underbrace{2r_{t,\ell}\bigl(1-s_{t,\ell}\bigr)}_{\text{angular}},
\label{eq:update_decomp}
\end{equation}
in which the two terms have medians of $0.045$-$0.073$ (radial) and $0.41$-$0.47$
(angular) respectively. Figure~\ref{fig:assumption_check} (top) reports the
two underlying diagnostics directly, $\rho_{t,\ell}$ and $s_{t,\ell}$, rather
than the terms of Eq.~\ref{eq:update_decomp}. Note that these are opposite in
sense ($s_{t,\ell}\to1$ and $\rho_{t,\ell}\to0$ both denote an unchanged
update), so the vertical separation between the two curves in that panel
carries no meaning. We emphasize that
Assumption~\ref{asm:smooth_updates} does \emph{not} require consecutive updates
to be close as vectors. The total relative discrepancy has a median of
$0.41$-$0.47$ on these trajectories, and is dominated by the angular part
($0.41$-$0.47$) rather than the radial part ($0.045$-$0.073$), a factor of
$6$-$10$. The magnitude term in Eq.~\ref{eq:cost_drift_appendix} depends only on the
radial part, which is why Assumption~\ref{asm:smooth_updates} is stated as a
constraint on update scale and not on update direction.

\paragraph{Local Fisher-scale stability.}
We empirically assess this assumption during DINOv2 ViT-S/14 fine-tuning on four VTAB-1k tasks (Figure~\ref{fig:assumption_check}). We approximate the FIM using AdaFisher's diagonal Kronecker form \citep{gomesAdaFisherAdaptiveSecond2024}:
\vspace{-2mm}
\begin{equation}
\tilde{\mathbf{F}}^{D}_\ell
=
\mathbf{H}_{D_{\ell-1}}
\otimes
\mathbf{S}_{D_\ell},
\label{eq:adafisher_fisher_form_appendix}
\end{equation}
where $\mathbf{H}_{D_{\ell-1}}$ and $\mathbf{S}_{D_\ell}$ are diagonal activation and gradient correlation factors, yielding linear computational and memory complexity. We measure scale variation by the relative change in mean absolute Fisher magnitude and structural change by the Jensen-Shannon (JS) distance between consecutive Fisher-induced distributions. Metric definitions are given below, and the JS construction is detailed in Appendix~\ref{appendix:js-distances}. Across all four tasks, Fisher-scale change is only $0.10$-$0.17\%$, while structural drift persists, supporting locally stable scale without stationary Fisher structure.

To examine the slowly varying Fisher-scale assumption, we define the mean absolute Fisher magnitude of tensor $\ell$ at iteration $t$ as:
\begin{equation}
\bar F_{t,\ell}
=
\frac{1}{N_\ell}
\sum_{i=1}^{N_\ell}
|F^t_\ell(i)|,
\end{equation}
and its relative temporal change as:
\begin{equation}
r_{t,\ell}
=
\frac{
|\bar F_{t,\ell}-\bar F_{t-1,\ell}|
}{
\bar F_{t-1,\ell}+\varepsilon
}.
\end{equation}
The Fisher is estimated using the efficient diagonal Kronecker approximation proposed by AdaFisher \cite{gomesAdaFisherAdaptiveSecond2024}, collecting it every five optimizer steps. We additionally compute the JS distance between consecutive Fisher-value distributions, which is robust to uniform rescaling (Appendix~\ref{appendix:js-distances}), to separate changes in Fisher \emph{structure} from changes in overall \emph{scale}.
Both quantities are smoothed using an exponential moving average with decay $0.8$, and Figure~\ref{fig:assumption_check} reports their cross-layer means.

\section{FisherAdapTune: Algorithm and Implementation}
\label{appendix:algorithm_details}

\subsection{Parameter Grouping}
\label{appendix:detailed_algorithm}

\noindent Algorithm~\ref{alg:fisher_js_select3} expands the main-text Algorithm \ref{alg:fisher_adap_tune_alg}.
FisherAdapTune operates on parameter groups rather than freezing an entire
tensor at once. For a matrix-shaped parameter tensor
$\mathbf{W}_i\in\mathbb{R}^{d_i^{\mathrm{out}}\times d_i^{\mathrm{in}}}$,
we divide its columns into $K$ contiguous, non-overlapping blocks. Let
\[
b_k=\left\lfloor \frac{k\,d_i^{\mathrm{in}}}{K}\right\rfloor,
\qquad k=0,\ldots,K.
\]
The $k$th group is then defined as
\[
\mathbf{W}_{ik}
=
\mathbf{W}_i[:,\,b_{k-1}:b_k],
\qquad k=1,\ldots,K,
\]
with the corresponding Fisher block
\[
\tilde{\mathbf{F}}_{\mathbf{W}_{ik}}^{\,t}
=
\tilde{\mathbf{F}}_{D_i}^{\,t}
[:,\,b_{k-1}:b_k].
\]
Thus, every group contains all output rows but only a subset of the input
columns. The blocks collectively cover the tensor without overlap, and their
numbers of columns differ by at most one when
$d_i^{\mathrm{in}}$ is not divisible by $K$. Convolutional kernels are first
viewed in matrix form by flattening the input-channel and spatial dimensions,
while one-dimensional tensors are partitioned into contiguous segments using
the same rule. Each resulting block is treated as an independent member
$(i,k)$ of the active set $\mathcal{S}$ and may therefore be frozen at a
different training step.

The block count $K$ controls the granularity of parameter selection. A small
$K$ produces coarse groups and lower monitoring overhead, whereas a larger
$K$ permits more localized freezing at the cost of maintaining more Fisher
distributions and drift statistics. Setting $K=1$ reduces the procedure to
tensor-wise selection. In practice, the effective number of blocks for a
dimension of size $d$ is $\min(K,d)$, preventing empty groups.

\subsection{Drift Smoothing and Adaptive Freezing}

The EMA factor $\gamma\in[0,1)$ controls the temporal smoothing of the
AdaFisher activation and gradient statistics. Larger values yield more stable
Fisher estimates while responding more slowly to changes during fine-tuning. The
second EMA factor $\beta\in[0,1)$ independently smooths the consecutive
Jensen--Shannon (JS) drift values, where larger $\beta$ suppresses short-term
fluctuations, and smaller $\beta$ makes the freezing criterion more
responsive. The Fisher interval $m$ specifies how often the Fisher statistics
and group-wise JS drift are updated. The freeze interval $n$ specifies how
often the active groups are evaluated for freezing, with $m<n$ allowing
multiple Fisher observations to be collected between consecutive freezing
decisions.
At each freezing step, the threshold scale $\lambda$ determines the
data-adaptive cutoff:
\[
\tau=\mu+\lambda\sigma,
\]
where $\mu$ and $\sigma$ are computed from the running mean JS drifts of the
currently active groups. Groups with
$\bar d_{\mathrm{JS},ik}<\tau$ are regarded as stabilized since their Fisher
distributions change relatively little, and are removed from
$\mathcal{S}$. Decreasing $\lambda$ lowers the threshold and therefore yields
more conservative freezing, whereas increasing $\lambda$ raises the threshold
and freezes more groups. Finally, $T_{\max}$ is the total training budget and
$\eta$ is the optimizer learning rate. Once a group is removed from
$\mathcal{S}$, its gradient update is masked for all subsequent iterations,
while the remaining active groups continue to be optimized
(Appendix~\ref{appendix:overhead} details when masked tensors are released).

\subsection{Complete Algorithm}

\begingroup
\small
\begin{algorithm}[H]
\caption{FisherAdapTune Algorithm}
\label{alg:fisher_js_select3}
\begin{algorithmic}[1]
\Require Parameters $\theta=\{\theta_i\}_{i=1}^{L}$, dataset $\mathcal{D}$, EMA factors $\gamma,\beta$, threshold scale $\lambda$, slice count $K$, Fisher interval $m$, freeze interval $n$,  total steps $T_{\max}$. 
\State Initialize active parameter groups set $\mathcal{S}$ to all supported groups.
\For{$t = 1$ \textbf{to} $T_{\max}$}
\State Compute loss $\mathcal{L}(\mathcal{D};\theta)$ and gradients $\nabla_\theta \mathcal{L}$.
\If{$t \bmod m = 0$}
\State Update AdaFisher activation and gradient statistics:
\[
\mathbf{H}_{D_{i-1}}^{t} \leftarrow \gamma \mathbf{H}_{D_{i-1}}^{t-1} + (1-\gamma)\widehat{\mathbf{H}}_{D_{i-1}}^{t},
\quad
\mathbf{S}_{D_i}^{t} \leftarrow \gamma \mathbf{S}_{D_i}^{t-1} + (1-\gamma)\widehat{\mathbf{S}}_{D_i}^{t}.
\]
\State Estimate layer Fisher tensors
$\tilde{\mathbf{F}}_{D_i}^{t}=\mathbf{H}_{D_{i-1}}^{t}\otimes\mathbf{S}_{D_i}^{t}$.
\State Partition each $\tilde{\mathbf{F}}_{D_i}^{t}$ into $K$ column-wise parameter groups $\{\tilde{\mathbf{F}}_{w_{ik}}^{t}\}_{k}$.
\State Build PMFs $p_{ik}^{t}=p(\tilde{\mathbf{F}}_{w_{ik}}^{t})$ using Eqs.~\ref{eq:fisher_log}-\ref{eq:fisher_probability}.
\State Compute consecutive Fisher drift:
\[
\tilde{d}_{\mathrm{JS},ik}^{t}
=
d_{\mathrm{JS}}(p_{ik}^{t-1},p_{ik}^{t})
=
\sqrt{\tfrac{1}{2}\mathrm{KL}(p_{ik}^{t-1}\|M_{ik}^{t})
+\tfrac{1}{2}\mathrm{KL}(p_{ik}^{t}\|M_{ik}^{t})},
\quad
M_{ik}^{t}=\tfrac{1}{2}(p_{ik}^{t-1}+p_{ik}^{t}).
\]
\State Smooth and accumulate the drift:
$\hat{d}_{\mathrm{JS},ik}^{t} \leftarrow
\beta\hat{d}_{\mathrm{JS},ik}^{t-1}+(1-\beta)\tilde{d}_{\mathrm{JS},ik}^{t}$,
$\mathcal{J}_{ik}\leftarrow \mathcal{J}_{ik}\cup\{\hat{d}_{\mathrm{JS},ik}^{t}\}$.
\State Compute running mean
$\bar{d}_{\mathrm{JS},ik} \leftarrow |\mathcal{J}_{ik}|^{-1}\sum_{d\in\mathcal{J}_{ik}} d$.
\EndIf
\If{$t \bmod n = 0$}
\State Compute global drift statistics over active parameter groups:
$\mu\leftarrow\mathrm{mean}_{(i,k)\in\mathcal{S}}(\bar{d}_{\mathrm{JS},ik})$,
$\sigma\leftarrow\mathrm{std}_{(i,k)\in\mathcal{S}}(\bar{d}_{\mathrm{JS},ik})$.
\State Set $\tau\leftarrow \mu+\lambda\sigma$ and freeze stabilized parameter groups:
$\mathcal{S}\leftarrow\{(i,k)\in\mathcal{S}:\bar{d}_{\mathrm{JS},ik}\geq\tau\}$.
\EndIf
\State Update only active parameter groups: $\theta_{\mathcal{S}}\leftarrow\theta_{\mathcal{S}}-\eta\nabla_{\theta_{\mathcal{S}}}\mathcal{L}(\mathcal{D};\theta)$.
\EndFor
\end{algorithmic}
\end{algorithm}
\endgroup

\subsection{Training-Time and Memory Overhead}
\label{appendix:overhead}

\begin{figure}[H]
    \centering
    \includegraphics[width=0.8\linewidth]{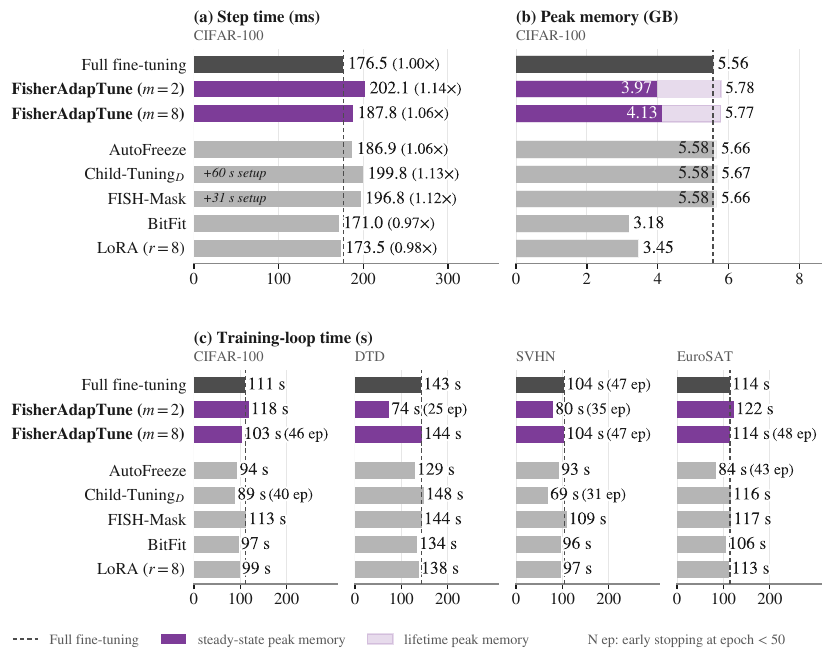}
    \caption{Training-time and memory overhead with DINOv2 ViT-S/14
(batch size 64, one RTX~6000 Ada). \textbf{(a)} CUDA-synchronized
per-step time on CIFAR-100, excluding validation; one-time setup costs
are annotated separately and excluded from step timing.
\textbf{(b)} Steady-state and lifetime peak allocated memory on
CIFAR-100; the lifetime peak includes the initial all-trainable phase.
\textbf{(c)} Taskwise training-loop time, excluding setup, validation,
and evaluation. 
}
    \label{fig:training_overhead}
\end{figure}

Figure~\ref{fig:training_overhead} reports per-step cost and peak memory
on CIFAR-100, as well as taskwise training-loop times. Fisher
statistics and JS distances are computed periodically. Until all $K$
blocks of a tensor are frozen, block-level masking zeros the corresponding
gradients and optimizer moments, while autograd continues to process the
tensor. Once all blocks are frozen, the tensor is set to non-trainable,
releasing its gradient computation, saved activations, and optimizer
state. With DINOv2 ViT-S/14, at $m{=}2$, 142 tensors containing 10.2M of 22.1M parameters are
fully frozen by the end of the timed region. Without tensor-level
release, peak memory is 5.73\,GB.

With $m{=}2$, FisherAdapTune requires $1.14\times$ the per-step time of
full fine-tuning while reducing steady-state peak memory from 5.56\,GB
to 3.97\,GB. Its lifetime peak is 5.78\,GB since training begins with
all parameters active. Increasing the collection interval to $m{=}8$
reduces the per-step ratio to $1.06\times$, comparable to AutoFreeze,
while maintaining a 4.13\,GB steady-state peak. Performance remains
stable across the evaluated collection intervals
(Appendix~\ref{appendix:hyperparameter_sensitivity}). Taskwise
training-loop times are also reported, dependent on both per-step cost
and the early-stopping epoch. The per-step measurements therefore
provide the more controlled comparison of computational overhead.

\section{Experimental Setup}
\label{appendix:experimental_setup}

\subsection{VTAB-1k Classification}
The VTAB-1k evaluation comprises all 19 tasks across its three categories: seven natural-image tasks, four specialized tasks, and eight structured tasks. Our objective is to compare adaptation strategies under a common experimental protocol. We report Top-1 accuracy for each task, with macro results computed as the unweighted mean across all 19 tasks.
We use a DINOv2 ViT-S/14 backbone pretrained on LVD-142M (\texttt{timm} model \texttt{vit\_small\_patch14\_dinov2.lvd142m}) and initialize a task-specific linear classification head for each dataset. Following the low-data VTAB-1k setting, each task uses 800 training samples, with a held-out or official validation split for model selection and early stopping. Evaluation is performed on the full official test split when available, and on the held-out remainder otherwise. All inputs are processed at $224\times224$ resolution and normalized using ImageNet statistics. During training, we apply random resized cropping, horizontal flipping, and color jitter; validation and test images are resized and center-cropped.

Training minimizes cross-entropy loss with label smoothing of $0.1$ using AdamW with a learning rate of $5\times10^{-5}$, weight decay of $10^{-2}$, and batch size 64 for up to 50 epochs. The learning rate is linearly warmed up for the first 5 epochs and subsequently cosine-decayed. Early stopping is applied when the validation loss does not improve for 10 consecutive epochs. Unless otherwise specified in the ablation studies, FisherAdapTune uses $k{=}4$, $m{=}2$, $n{=}6$, $\beta{=}0.8$, and $\lambda{=}-0.65$. Full fine-tuning and FisherAdapTune share the same data pipeline, task-head initialization, and optimization recipe. FisherAdapTune differs only through its progressive restriction of the trainable backbone parameters. Experiments were run on Digital Research Alliance of Canada NVIDIA H100 GPUs and local NVIDIA RTX~6000 Ada Generation GPUs.

\subsection{Crack Segmentation}
The OmniCrack30k dataset contains approximately 30,000 images aggregated from more than 20 crack datasets and covers concrete, asphalt, ceramic, masonry, and steel surfaces. Its held-out test set is used for in-distribution evaluation. Zero-shot generalization is evaluated on Concrete3k, Facade390, and Road420, which represent distinct acquisition conditions and surface domains. We report F1-score and IoU for each dataset and their unweighted mean across the three zero-shot datasets.
All crack-segmentation experiments use AdamW with a cosine learning-rate schedule and batch size 2. We use a learning rate of $10^{-6}$ for SAM2 and $10^{-4}$ for SegFormer, and optimize the sum of BCE-with-logits and Dice losses. Images and masks are resized to $256\times256$ for both training and evaluation. Within each backbone, FisherAdapTune uses the same initialization, data splits, optimizer, loss, and schedule as the baselines. FisherAdapTune differs only by progressively removing parameter groups from the optimizer once their smoothed Fisher JS drift satisfies the adaptive freezing criterion.

\section{Comparative Results}
\label{appendix:comparative_results}

\subsection{In-Distribution and Zero-Shot Segmentation Results}
\label{appendix:segmentation_results}

\begin{figure}[!htbp]
    \centering
    \begin{subfigure}[t]{0.48\linewidth}
        \centering
   \includegraphics[width=\linewidth]{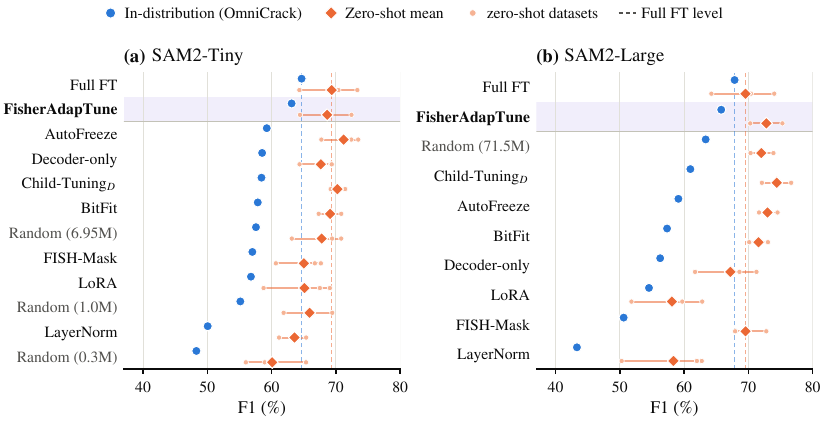}
        \label{fig:sam2_comparatives}
    \end{subfigure}
    \hfill
    \begin{subfigure}[t]{0.48\linewidth}
        \centering
\includegraphics[width=\linewidth]{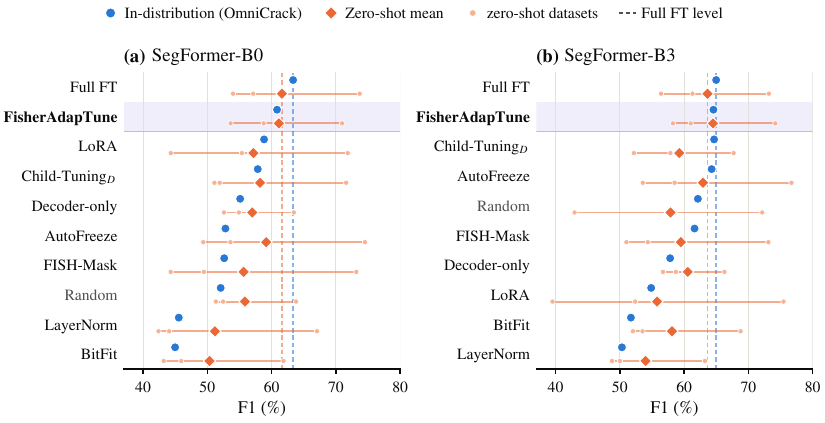}
\label{fig:segformer_comparatives}
    \end{subfigure}
       \caption{Comparison of in-distribution and zero-shot crack-segmentation performance across fine-tuning methods for SAM2 and SegFormer architectures.}
\label{fig:crack_segmentation_results}
\end{figure}

Table~\ref{tab:sfinetuning_comparisons} and Figure \ref{fig:crack_segmentation_results} compare fine-tuning methods across four backbones on crack segmentation. FisherAdapTune provides a strong balance between in-distribution performance, zero-shot generalization, and parameter efficiency. It consistently outperforms matched-budget random selection and achieves the highest average zero-shot performance for SegFormer-MiT-B3 while remaining competitive across the other backbones.
Table~\ref{tab:crack_multiseed} further reports three-seed results for FisherAdapTune and matched-budget random selection. FisherAdapTune consistently improves in-distribution performance and exhibits substantially lower variability. Its zero-shot gains are strongest for SegFormer, while remaining comparable to random selection for SAM2, confirming that Fisher-guided selection is stable across seeds and more effective than sparsity alone.

\begin{table}[H]
\centering
\caption{Performance comparison of fine-tuning methods for crack segmentation in terms of in-distribution performance (OmniCrack test) and zero-shot generalization. Bold entries show the best per column within each model block; \underline{underline} shows the second best. Effect. Params denotes number of Effective Tunable Parameters.}
\label{tab:sfinetuning_comparisons}
\scriptsize
\setlength{\tabcolsep}{2.5pt}
\renewcommand{\arraystretch}{1.15}

\begin{tabularx}{\textwidth}{
>{\raggedright\arraybackslash}p{1.2cm}
>{\raggedright\arraybackslash}m{1.8cm}
>{\centering\arraybackslash}m{1.cm}
cc
cccc
cccc
}
\toprule
\textbf{Model} &
\textbf{Fine-tune Method} &
\makecell{\textbf{Effect.}\\\textbf{Params}}&
\multicolumn{2}{c}{\textbf{OmniCrack}} &
\multicolumn{4}{c}{\textbf{Zero-shot F1 (\%)}} &
\multicolumn{4}{c}{\textbf{Zero-shot IoU (\%)}} \\

\cmidrule(lr){4-5}
\cmidrule(lr){6-9}
\cmidrule(lr){10-13}

& & &
\textbf{F1} & \textbf{IoU} &
\textbf{Conc.} & \textbf{Fac.} & \textbf{Road} & \textbf{Avg.$\pm$Std} &
\textbf{Conc.} & \textbf{Fac.} & \textbf{Road} & \textbf{Avg.$\pm$Std} \\

\midrule

\multirow{13}{*}{SAM2-Tiny}
& Freeze     & 0.0M  & 4.37 & 2.24 & 8.21 & 4.13 & 4.47 & $5.60_{\pm2.27}$ & 4.28 & 2.11 & 2.28 & $2.89_{\pm1.21}$ \\
& Full FT    & 38.9M & \textbf{64.67} & \textbf{47.79} & \textbf{73.34} & 70.39 & 64.32 & $69.35_{\pm4.60}$ & \textbf{57.91} & 54.31 & 47.40 & $53.21_{\pm5.34}$ \\
& LoRA       & 0.28M & 56.79 & 39.65 & 69.07 & 67.51 & 58.76 & $65.11_{\pm5.56}$ & 52.75 & 50.96 & 41.60 & $48.44_{\pm5.99}$ \\
& BitFit     & 0.11M & 57.85 & 40.69 & 70.84 & 69.17 & 67.33 & $69.11_{\pm1.76}$ & 54.85 & 52.87 & 50.74 & $52.82_{\pm2.05}$ \\
& LayerNorm  & 0.03M & 50.06 & 33.39 & 64.15 & 65.38 & 61.16 & $63.56_{\pm1.77}$ & 47.23 & 48.56 & 44.05 & $46.61_{\pm1.89}$ \\
& Decoder-only    & 4.2M  & 58.53 & 41.48 & 69.26 & 69.36 & 64.35 & $67.66_{\pm2.86}$ & 52.98 & 53.09 & 47.44 & $51.17_{\pm3.23}$ \\
& Random     & 0.3M  & 48.30 & 31.84 & 56.00 & 65.36 & 58.93 & $60.09_{\pm3.91}$ & 38.88 & 48.54 & 41.77 & $43.07_{\pm4.05}$ \\
& Random     & 1.0M  & 55.14 & 38.07 & 66.41 & 69.44 & 61.88 & $65.91_{\pm3.11}$ & 49.71 & 53.19 & 44.80 & $49.24_{\pm3.44}$ \\
& Random     & 6.95M & 57.57 & 40.42 & 69.41 & 70.83 & 63.15 & $67.80_{\pm3.34}$ & 53.15 & 54.83 & 46.14 & $51.38_{\pm3.76}$ \\
& AutoFreeze  & 21.40M & 59.25 & 42.10 & 72.41 & \textbf{73.48} & \underline{67.75} & $\mathbf{71.21_{\pm3.05}}$ & 56.76 & \textbf{58.08} & \underline{51.23} & $\mathbf{55.36_{\pm3.63}}$ \\
& Child-Tuning$_D$  & 9.67M  & 58.43 & 41.27 & 70.07 & \underline{71.45} & \textbf{69.20} & $\underline{70.24_{\pm1.13}}$ & 53.93 & \underline{55.58} & \textbf{52.91} & $\underline{54.14_{\pm1.35}}$ \\
& FISH-Mask & 0.19M  & 57.01 & 39.87 & 67.67 & 66.73 & 60.67 & $65.02_{\pm3.80}$ & 51.14 & 50.07 & 43.54 & $48.25_{\pm4.11}$ \\
\rowcolor{blue!8}
& \textbf{Fisher\-AdapTune} & 6.95M & \underline{63.12} & \underline{46.11} & \underline{72.43} & 69.10 & 64.42 & $68.65_{\pm4.02}$ & \underline{56.77} & 52.79 & 47.52 & $52.36_{\pm4.64}$ \\

\midrule

\multirow{11}{*}{SAM2-Large}
& Freeze    & 0.0M   & 3.54 & 1.80 & 6.29 & 2.41 & 3.93 & $4.21_{\pm1.95}$ & 3.25 & 1.22 & 2.00 & $2.16_{\pm1.02}$ \\
& Full FT   & 224M   & \textbf{67.86} & \textbf{51.35} & \textbf{74.03} & 64.25 & 70.39 & $69.56_{\pm5.00}$ & \textbf{58.77} & 47.33 & 54.31 & $53.47_{\pm5.8}$ \\
& LoRA      & 0.221M & 54.54 & 37.49 & 59.72 & 62.82 & 51.81 & $58.11_{\pm5.68}$ & 42.57 & 45.79 & 34.96 & $41.11_{\pm5.56}$ \\
& BitFit    & 0.41M  & 57.35 & 40.20 & 73.05 & 70.13 & 71.55 & $71.58_{\pm1.19}$ & 57.54 & 54.00 & 55.70 & $55.75_{\pm1.44}$ \\
& LayerNorm & 0.12M  & 43.33 & 27.66 & 50.29 & 62.75 & 61.99 & $58.34_{\pm5.70}$ & 33.59 & 45.72 & 44.92 & $41.41_{\pm5.54}$ \\
& Decoder-only   & 4.2M   & 56.28 & 39.16 & 71.26 & 68.61 & 61.71 & $67.19_{\pm4.93}$ & 55.35 & 52.22 & 44.62 & $50.73_{\pm5.52}$ \\
& Random    & 71.49M & 63.36 & 46.37 & \underline{73.90} & 70.35 & 71.79 & $72.01_{\pm1.46}$ & \underline{58.60} & 54.26 & 55.99 & $56.29_{\pm1.79}$ \\
& AutoFreeze & 90.28M & 59.12 & 41.97 & 71.66 & 74.54 & 72.73 & $\underline{72.98_{\pm1.46}}$ & 55.83 & 59.42 & 57.14 & $\underline{57.46_{\pm1.82}}$ \\
& Child-Tuning$_D$ & 46.76M & 60.98 & 43.86 & 72.11 & \textbf{76.66} & \textbf{74.47} & $\mathbf{74.41_{\pm2.28}}$ & 56.38 & \textbf{62.15} & \textbf{59.33} & $\mathbf{59.29_{\pm2.89}}$ \\
& FISH-Mask & 1.12M & 50.60 & 33.87 & 67.98 & 67.94 & \underline{72.79} & $69.57_{\pm2.79}$ & 51.50 & 51.45 & \underline{57.22} & $53.39_{\pm3.32}$ \\
\rowcolor{blue!8}
& \textbf{Fisher\-AdapTune} & 71.49M & \underline{65.78} & \underline{49.01} & 72.81 & \underline{75.30} & 70.29 & $72.80_{\pm2.04}$ & 57.24 & \underline{60.38} & 54.19 & $57.27_{\pm2.53}$ \\

\midrule

\multirow{10}{*}{\makecell[l]{SegFormer\\-MiT-B0}}
& Full FT      & 3.71M
  & \textbf{63.35} & \textbf{46.36}
  & \underline{73.71} & \underline{57.14} & \textbf{54.00} & $\mathbf{61.61_{\pm8.64}}$
  & \underline{58.37} & \underline{40.00} & \textbf{36.98} & $\mathbf{45.11_{\pm9.45}}$ \\
& LoRA         & 0.23M & 58.83 & 41.68 & 71.85 & 55.38 & 44.31 & $57.18_{\pm11.31}$ & 56.06 & 38.29 & 28.46 & $40.94_{\pm11.42}$ \\
& BitFit       & 0.02M & 44.99 & 29.02 & 61.86 & 45.95 & 43.23 & $50.34_{\pm8.21}$  & 44.79 & 29.83 & 27.57 & $34.06_{\pm7.64}$  \\
& LayerNorm    & 0.01M & 45.56 & 29.50 & 67.07 & 44.07 & 42.41 & $51.18_{\pm11.25}$ & 50.46 & 28.26 & 26.91 & $35.21_{\pm10.79}$ \\
& Decoder-only & 0.40M & 55.11 & 38.04 & 63.47 & 54.91 & 52.58 & $56.98_{\pm4.68}$  & 46.48 & 37.85 & 35.66 & $40.00_{\pm4.67}$  \\
& Random       & 0.12M & 52.08 & 35.21 & 63.81 & 51.33 & 52.45 & $55.87_{\pm5.64}$  & 46.85 & 34.53 & 35.55 & $38.98_{\pm5.58}$  \\
& AutoFreeze   & 1.43M & 52.82 & 35.89 & \textbf{74.53} & 49.37 & 53.60 & $59.17_{\pm13.47}$ & \textbf{59.40} & 32.77 & 36.61 & $42.93_{\pm14.39}$ \\
& Child-Tuning$_D$ & 1.39M & 57.85 & 40.70 & 71.60 & 51.93 & 51.11 & $58.21_{\pm11.60}$ & 55.77 & 35.08 & 34.33 & $41.73_{\pm12.17}$ \\
& FISH-Mask & 0.02M & 52.63 & 35.71 & 73.18 & 49.46 & 44.30 & $55.65_{\pm15.40}$ & 57.70 & 32.85 & 28.46 & $39.67_{\pm15.77}$ \\
\rowcolor{blue!8}
& \textbf{Fisher\-AdapTune} & 0.12M
  & \underline{60.85} & \underline{43.73}
  & 70.97 & \textbf{58.77} & \underline{53.63} & $\underline{61.12_{\pm7.27}}$
  & 55.01 & \textbf{41.62} & \underline{36.64} & $\underline{44.42_{\pm7.75}}$ \\

\midrule

\multirow{10}{*}{\makecell[l]{SegFormer\\-MiT-B3}}
& Full FT      & 44.60M
  & \textbf{64.99} & \textbf{48.14}
  & 73.20 & \textbf{61.31} & 56.42 & $\underline{63.64_{\pm7.04}}$
  & 57.72 & \textbf{44.20} & 39.30 & $\underline{47.07_{\pm7.78}}$ \\
& LoRA         & 1.79M & 54.88 & 37.82 & \underline{75.47} & 52.39 & 39.53 & $55.79_{\pm14.87}$ & \underline{60.60} & 35.49 & 24.63 & $40.24_{\pm15.06}$ \\
& BitFit       & 0.14M  & 51.73 & 34.89 & 68.79 & 52.04 & 53.53          & $58.12_{\pm7.56}$   & 52.42 & 35.17 & 36.55          & $41.38_{\pm7.82}$  \\
& LayerNorm    & 0.05M  & 50.33 & 33.63 & 63.24 & 50.01 & 48.79          & $54.01_{\pm6.54}$   & 46.24 & 33.34 & 32.27          & $37.28_{\pm6.34}$  \\
& Decoder-only & 0.53M  & 57.83 & 40.67 & 66.28 & 56.71 & \textbf{58.72} & $60.57_{\pm4.12}$   & 49.57 & 39.57 & \textbf{41.57} & $43.57_{\pm4.32}$  \\
& Random       & 10.73M & 62.15 & 45.09 & 72.16 & 58.56 & 42.96          & $57.89_{\pm11.93}$  & 56.45 & 41.40 & 27.36          & $41.74_{\pm11.88}$ \\
& AutoFreeze   & 22.72M & 64.29 & 47.38 & \textbf{76.71} & 53.58 & \underline{58.53} & $62.94_{\pm12.18}$ & \textbf{62.22} & 36.59 & \underline{41.37} & $46.73_{\pm13.63}$ \\
& Child-Tuning$_D$ & 9.34M & \underline{64.67} & \underline{47.79} & 67.71 & 57.87 & 52.19 & $59.26_{\pm7.85}$ & 51.18 & 40.72 & 35.31 & $42.40_{\pm8.07}$ \\
& FISH-Mask & 0.22M & 61.63 & 44.54 & 73.12 & 54.36 & 51.02 & $59.50_{\pm11.91}$ & 57.63 & 37.32 & 34.25 & $43.07_{\pm12.71}$ \\
\rowcolor{blue!8}
& \textbf{Fisher\-AdapTune} & 10.73M
  & 64.57 & 47.68
  & 74.19 & \underline{61.04} & 58.26 & $\mathbf{64.49_{\pm6.94}}$
  & 58.97 & \underline{43.93} & 41.11 & $\mathbf{48.00_{\pm7.83}}$ \\

\bottomrule
\end{tabularx}

\end{table}

\FloatBarrier

\begin{table}[H]
\centering
\caption{Comparison of FisherAdapTune and matched-budget random selection across four crack-segmentation experiments using three random seeds. In-distribution results are reported on OmniCrack30K, and zero-shot results are averaged across Concrete3k, Facade390, and Road420. Values denote mean $\pm$ standard deviation across seeds.}
\label{tab:crack_multiseed}
\footnotesize
\begin{tabular}{l cc cc}
\toprule
& \multicolumn{2}{c}{In-distribution} & \multicolumn{2}{c}{Zero-shot mean} \\
\cmidrule(lr){2-3} \cmidrule(lr){4-5}
Method & F1 & IoU & F1 & IoU \\
\midrule
\multicolumn{5}{l}{\emph{SegFormer-B0}} \\
FisherAdapTune ($\lambda{=}-0.95$) & $61.21_{\pm0.48}$ & $44.11_{\pm0.50}$ & $61.24_{\pm1.40}$ & $44.55_{\pm1.36}$ \\
Random subset ($0.124$M) & $43.38_{\pm18.64}$ & $28.84_{\pm14.40}$ & $49.35_{\pm12.45}$ & $34.37_{\pm9.51}$ \\
\midrule
\multicolumn{5}{l}{\emph{SegFormer-B3}} \\
FisherAdapTune ($\lambda{=}-0.95$) & $64.41_{\pm0.14}$ & $47.51_{\pm0.15}$ & $64.40_{\pm0.90}$ & $48.03_{\pm0.95}$ \\
Random subset ($10.73$M) & $63.84_{\pm1.71}$ & $46.91_{\pm1.84}$ & $61.40_{\pm3.26}$ & $44.98_{\pm3.04}$ \\
\midrule
\multicolumn{5}{l}{\emph{SAM2-Tiny}} \\
FisherAdapTune ($\lambda{=}-0.85$) & $63.20_{\pm0.16}$ & $46.20_{\pm0.17}$ & $68.88_{\pm0.21}$ & $52.65_{\pm0.25}$ \\
Random subset ($6.95$M) & $60.70_{\pm2.71}$ & $43.61_{\pm2.76}$ & $68.82_{\pm0.93}$ & $52.54_{\pm1.06}$ \\
\midrule
\multicolumn{5}{l}{\emph{SAM2-Large}} \\
FisherAdapTune ($\lambda{=}-0.87$) & $66.02_{\pm0.31}$ & $49.28_{\pm0.34}$ & $72.63_{\pm0.16}$ & $57.06_{\pm0.19}$ \\
Random subset ($71.49$M) & $63.47_{\pm0.58}$ & $46.49_{\pm0.62}$ & $72.83_{\pm0.75}$ & $57.32_{\pm0.94}$ \\
\bottomrule
\end{tabular}
\end{table}

\section{Ablation Studies and Hyperparameter Sensitivity}
\label{appendix:ablations}

\subsection{Comparison of Parameter-Selection Signals}
\label{appendix:signal_ablation}

\begin{figure}[ht]
  \centering
  \includegraphics[width=\linewidth]{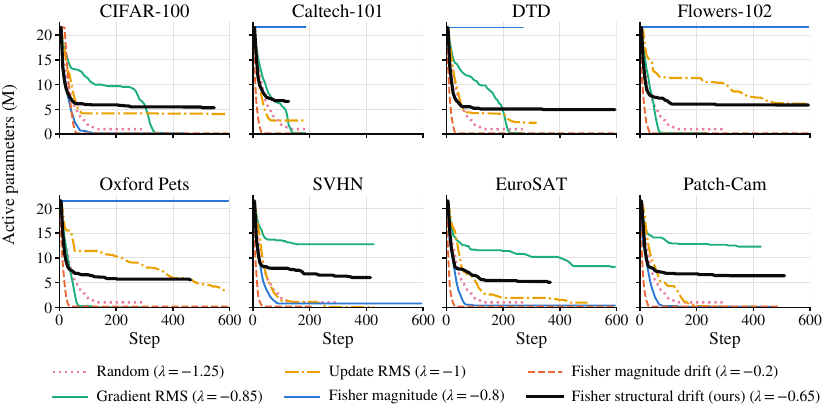}
  \caption{%
    Evolution of the active trainable-parameter budget across freezing events for each selection signal on the eight VTAB-1k tasks. Each alternative is shown at the tuned \(\lambda\) where the parameter budget is closest to that of Fisher structural drift, with all other components of FisherAdapTune kept fixed (Table~\ref{tab:signal_ablation_matched}). 
  }
  \label{fig:signal_ablation}
\end{figure}

\begin{figure}[H]
  \centering
  \includegraphics[width=\linewidth]{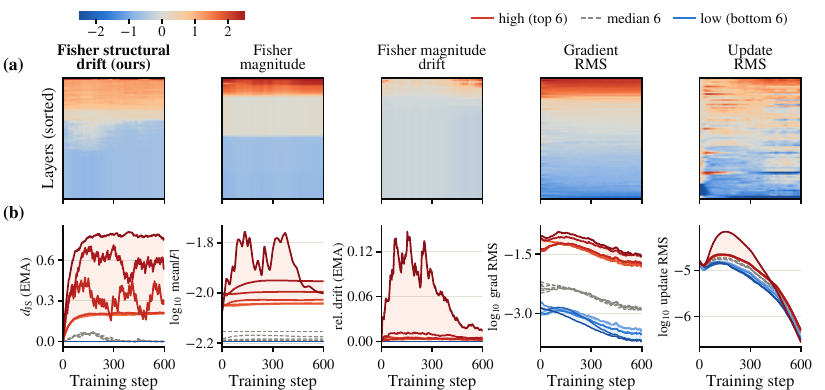}
  \caption{Parameter-selection signals monitored across DINOv2 layers during CIFAR-100 full fine-tuning. (a) Layerwise signal trajectories, with layers sorted within each signal. (b) Trajectories for six high-, six median-, and six low-signal layers. Fisher structural drift distinguishes groups with sustained drift from those that stabilize, while the other signals exhibit different layerwise and temporal patterns.}
\label{fig:signal_heterogeneity_2x5}
\end{figure}

To assess the selection signal, we replace Fisher structural drift with Fisher magnitude, Fisher-magnitude drift, gradient RMS, update RMS, and random selection while keeping the remaining algorithm unchanged. For parameter group $g$ with $d_g$ parameters, gradient RMS is defined as
$\sqrt{d_g^{-1}\sum_{i=1}^{d_g}(\partial\mathcal{L}/\partial\theta_{g,i})^2}$,
update RMS replaces the gradients with the corresponding parameter updates $\Delta\theta_{g,i}$.
Figure~\ref{fig:signal_heterogeneity_2x5} monitors these signals across DINOv2 layers during fine-tuning on CIFAR-100. Fisher structural drift separates groups whose Fisher distributions continue to change from groups whose drift approaches zero. Fisher magnitude also differs across layers, while its ordering is largely persistent. Fisher-magnitude drift is small for most groups apart from transient peaks. Gradient RMS is concentrated in a few layers yet declines broadly during training, while update RMS falls similarly and the group trajectories converge. Thus, variation across layers alone does not establish that a signal can distinguish \emph{continued adaptation} from stabilization over time.
Each alternative is reported at the tuned $\lambda$ where trainable budget comes closest to that of Fisher structural drift. For some signals, no $\lambda$ produces a close match and further adjustment retains all groups as trainable instead of yielding finer progressive freezing. Figure~\ref{fig:signal_ablation} shows the resulting budget trajectories, and Table~\ref{tab:signal_ablation_matched} reports their performance across eight VTAB-1k tasks. Taken together, these results support Fisher structural drift as a useful selection signal beyond the effect of reducing the trainable budget.

\subsection{Hyperparameter Sensitivity on VTAB-1k}
\label{appendix:hyperparameter_sensitivity}

We examine the sensitivity of FisherAdapTune in two complementary settings. First, we conduct a one-at-a-time sensitivity analysis on four representative VTAB-1k classification tasks using DINOv2 ViT-S/14. This analysis considers the parameter-group count $k$, Fisher collection interval $m$, freezing interval $n$, EMA smoothing factor $\beta$, and freezing-threshold scale $\lambda$. Second, we study the effect of $\lambda$ more extensively on the downstream crack-segmentation task using SAM2-Tiny, SAM2-Large, SegFormer-MiT-B0, and SegFormer-MiT-B3. The classification experiments assess general robustness to the algorithmic settings, whereas the segmentation experiments examine how the freezing threshold controls adaptation, parameter efficiency, and zero-shot generalization across model families.

Table~\ref{tab:hparam_sensitivity} reports the one-at-a-time sensitivity analysis on CIFAR-100, DTD, SVHN, and EuroSAT. Each hyperparameter is varied while the remaining values are fixed to the base configuration
($k{=}4$, $m{=}2$, $n{=}6$, $\beta{=}0.8$, and $\lambda{=}-0.65$). 
The results show that $m$, $n$, and $\beta$ have relatively limited effects on the performance. Across the evaluated collection intervals, macro accuracy varies from 77.76\% to 78.58\%, indicating that Fisher statistics do not need to be collected at every opportunity (i.e., iteration) to maintain performance. Increasing $m$ also reduces the frequency and computational cost of Fisher estimation. Similarly, varying $n$ produces a macro-accuracy range of 77.23\%-78.34\%. Although a larger $n$ reduces the frequency of freezing decisions, it delays parameter removal and therefore increases the average trainable budget from 5.38M at $n{=}3$ to 12.53M at $n{=}65$. The small variation across $\beta$ values further indicates that the method is not highly sensitive to the precise degree of temporal smoothing.

In contrast, $k$ and $\lambda$ directly determine the performance-efficiency operating point. Increasing $k$ creates finer-grained parameter groups, allowing FisherAdapTune to isolate and freeze smaller subsets of parameters. This reduces the mean trainable budget from 14.64M at $k{=}1$ to 1.59M at $k{=}16$, but overly fine grouping degrades macro accuracy. The selected value $k{=}4$ retains roughly half the average trainable parameters of $k{=}2$ while achieving comparable accuracy, providing a favorable balance between granularity and performance.
The threshold scale $\lambda$ has an even stronger effect on the parameter budget. Lower values apply a more conservative freezing criterion and retain more parameter groups, whereas larger values freeze groups more aggressively. Values between $-1.0$ and $-0.65$ yield similar macro accuracy, while $\lambda{=}-0.65$ reduces the mean trainable budget to 6.52M. Increasing the threshold to $-0.5$ or $-0.3$ causes premature freezing and a substantial loss in accuracy. Thus, tuning $\lambda$ allows FisherAdapTune to retain groups with continuing Fisher geometry variation while removing groups that have stabilized.

\begin{table}[H]
\centering
\caption{%
  Hyperparameter sensitivity on four VTAB-1k datasets and DINOv2 ViT-S/14. Each block varies one hyperparameter
  from a base configuration
  ($K{=}4$, $m{=}2$, $n{=}6$, $\beta{=}0.8$,
  $\lambda{=}-0.65$). $^{\star}$ marks the value adopted in this work for further experiments; the starred rows are the run of Table~\ref{tab:vtab19_peft_comparison} (weight decay $10^{-2}$, early-stopping patience 10), whereas the remaining rows use weight decay 0 and patience 15. M denotes the trajectory-averaged
  trainable-parameter budget in millions. 
}
\label{tab:hparam_sensitivity}
\setlength{\tabcolsep}{4pt}
\small
\begin{tabular}{l cc cc cc cc c c}
\toprule
& \multicolumn{2}{c}{CIFAR-100} & \multicolumn{2}{c}{DTD}
& \multicolumn{2}{c}{SVHN} & \multicolumn{2}{c}{EuroSAT}
& Macro & Mean \\
\cmidrule(lr){2-3} \cmidrule(lr){4-5} \cmidrule(lr){6-7} \cmidrule(lr){8-9}
Setting & Top-1 & M & Top-1 & M & Top-1 & M & Top-1 & M & Top-1 & M \\
\midrule
\multicolumn{11}{l}{\small\textit{Block count $K$}} \\[-2pt]
$k=1$  & 57.31 & 14.73 & 68.19 & 14.52 & 89.52 & 14.85 & 95.75 & 14.47 & 77.69 & 14.64 \\
$k=2$  & 58.01 & 12.94 & 70.05 & 10.68 & 89.88 & 13.11 & 95.76 & 12.39 & 78.42 & 12.28 \\
$k=4^{\star}$
 & 57.11 &  6.25 & 71.65 &  5.62 & 87.67 &  7.56 & 95.76 &  6.65 & 78.05 &  6.52 \\
$k=8$  & 53.35 &  2.58 & 70.00 &  2.02 & 86.34 &  2.67 & 95.17 &  1.61 & 76.22 &  2.22 \\
$k=16$ & 46.57 &  1.63 & 68.56 &  1.39 & 82.54 &  1.88 & 95.24 &  1.47 & 73.23 &  1.59 \\
\midrule
\multicolumn{11}{l}{\small\textit{Freezing interval $n$}} \\[-2pt]
$n=3$  & 57.83 &  5.88 & 71.91 &  4.77 & 87.63 &  4.85 & 96.00 &  6.04 & 78.34 &  5.38 \\
$n=6^{\star}$
 & 57.11 &  6.25 & 71.65 &  5.62 & 87.67 &  7.56 & 95.76 &  6.65 & 78.05 &  6.52 \\
$n=13$ & 57.22 &  8.16 & 71.86 &  7.10 & 86.29 &  9.48 & 95.80 &  6.68 & 77.79 &  7.85 \\
$n=26$ & 55.36 &  9.32 & 70.05 &  9.80 & 88.02 & 10.40 & 95.51 &  9.90 & 77.23 &  9.85 \\
$n=65$ & 55.78 & 12.11 & 69.10 & 12.17 & 89.47 & 12.50 & 95.29 & 13.33 & 77.41 & 12.53 \\
\midrule
\multicolumn{11}{l}{\small\textit{Collection interval $m$}} \\[-2pt]
$m=1$  & 57.45 &  7.20 & 72.02 &  5.07 & 87.75 &  6.81 & 95.39 &  7.23 & 78.15 &  6.58 \\
$m=2^{\star}$
 & 57.11 &  6.25 & 71.65 &  5.62 & 87.67 &  7.56 & 95.76 &  6.65 & 78.05 &  6.52 \\
$m=4$  & 57.73 &  6.82 & 71.17 &  5.42 & 86.35 &  7.83 & 95.80 &  5.78 & 77.76 &  6.46 \\
$m=8$  & 58.29 &  6.57 & 72.07 &  6.51 & 88.06 &  7.24 & 95.90 &  6.08 & 78.58 &  6.60 \\
\midrule
\multicolumn{11}{l}{\small\textit{JS smoothing $\beta$}} \\[-2pt]
$\beta=0$    & 57.47 & 6.92 & 70.90 & 5.87 & 88.41 & 6.41 & 95.79 & 5.08 & 78.14 & 6.07 \\
$\beta=0.5$  & 57.03 & 6.97 & 72.02 & 5.82 & 88.71 & 6.80 & 95.78 & 5.79 & 78.38 & 6.35 \\
$\beta=0.8^{\star}$
 & 57.11 &  6.25 & 71.65 &  5.62 & 87.67 &  7.56 & 95.76 &  6.65 & 78.05 &  6.52 \\
$\beta=0.95$ & 57.08 & 6.89 & 71.17 & 5.67 & 88.79 & 7.30 & 95.80 & 6.32 & 78.21 & 6.55 \\
\midrule
\multicolumn{11}{l}{\small\textit{Freezing threshold $\lambda$}} \\[-2pt]
$\lambda=-1.0$ & 58.19 & 12.69 & 70.16 & 12.53 & 89.49 & 13.04 & 94.55 & 14.03 & 78.10 & 13.07 \\
$\lambda=-0.8$ & 58.72 &  9.79 & 70.64 &  9.50 & 87.90 & 10.60 & 95.98 & 10.06 & 78.31 &  9.99 \\
$\lambda=-0.65^{\star}$
 & 57.11 &  6.25 & 71.65 &  5.62 & 87.67 &  7.56 & 95.76 &  6.65 & 78.05 &  6.52 \\
$\lambda=-0.5$ & 46.18 &  2.85 & 57.82 &  0.94 & 52.11 &  2.57 & 90.72 &  1.56 & 61.71 &  1.98 \\
$\lambda=-0.3$ &  8.56 &  1.04 & 41.65 &  0.76 & 38.43 &  0.84 & 88.37 &  0.75 & 44.25 &  0.85 \\
\bottomrule
\end{tabular}
\end{table}

\subsection{Freezing-Threshold Sensitivity on Crack Segmentation}
\label{appendix:threshold_ablations}
We further evaluate the freezing threshold, $\lambda$, on crack segmentation as it is the main hyperparameter controlling which parameter groups remain active. These experiments use OmniCrack30k for downstream adaptation and evaluate zero-shot transfer on Concrete3k, Facade390, and Road420. The Fisher collection and freezing intervals are fixed at $m{=}300$ and $n{=}3000$ training iterations, respectively. Tables~\ref{tab:sam2_lambda_ablation} and~\ref{tab:segformer_lambda_ablation} report the results for the SAM2 and SegFormer model families.
We report both final and effective number of trainable parameters. The final count represents the parameter groups that remain active at the end of training, whereas the effective number of trainable parameters  captures the average trainable-parameter count over the complete optimization. The latter accounts for parameters that contribute updates before being frozen and therefore provides a more representative measure of training-time parameter efficiency.

Table~\ref{tab:sam2_lambda_ablation} shows that more conservative thresholds generally improve in-distribution performance at the cost of larger trainable budgets. For SAM2-Tiny, moving from $\lambda=0$ to $-1$ increases OmniCrack F1 from 61.96\% to 64.51\% while raises the average budget from 3.82M to 18.07M. The selected $\lambda=-0.85$ maintains 63.12\% F1 with only 0.27M parameters active at the end of training. For SAM2-Large, $\lambda=-0.87$ achieves 65.78\% F1, 49.01\% IoU, and strong zero-shot performance with a final budget of 3.0M. Comparable or slightly better results at $\lambda=-0.9$ and $-0.95$ require substantially larger final budgets of 56.8M and 74.5M, respectively.
Table~\ref{tab:segformer_lambda_ablation} shows similar trends for SegFormer. For MiT-B0, the selected $\lambda=-0.95$ achieves 60.85\% F1 with an average budget of 0.124M and only 128 parameters remaining active, whereas further in-distribution gains require much larger budgets. For MiT-B3, $\lambda=-0.95$ provides the best overall balance, achieving 64.57\% F1 and 47.68\% IoU on OmniCrack and the highest mean zero-shot F1 and IoU of 64.49\% and 48.00\%. Although $\lambda=-1.0$ slightly improves in-distribution performance, it increases the average budget and reduces zero-shot performance, indicating that retaining more parameters does not necessarily improve generalization.

\begin{table}[H]
\centering
\caption{Ablation on the freezing threshold $\lambda$ for \textbf{FisherAdapTune} applied to crack segmentation using SAM2-Tiny and SAM2-Large. Larger $|\lambda|$ retains more trainable parameter groups. Bold entries show the best per column within each model block; \underline{underline} shows second best. Shaded rows indicate the selected $\lambda$. Effect. Params denotes number of Effective Tunable Parameters.}
\label{tab:sam2_lambda_ablation}
\scriptsize
\setlength{\tabcolsep}{3.5pt}
\renewcommand{\arraystretch}{1.15}

\begin{tabularx}{\textwidth}{
>{\centering\arraybackslash}m{1.3cm}
>{\centering\arraybackslash}m{0.7cm}
>{\centering\arraybackslash}m{0.8cm}
>{\centering\arraybackslash}m{0.8cm}
cc
cccc
cccc
}
\toprule
\textbf{Model} &
$\boldsymbol{\lambda}$ &
\makecell{\textbf{Final}\\\textbf{Tunable} \\\textbf{Params}}&
\makecell{\textbf{Effect.} \\\textbf{Params}}&
\multicolumn{2}{c}{\textbf{OmniCrack}} &
\multicolumn{4}{c}{\textbf{Zero-shot F1 (\%)}} &
\multicolumn{4}{c}{\textbf{Zero-shot IoU (\%)}} \\

\cmidrule(lr){5-6}
\cmidrule(lr){7-10}
\cmidrule(lr){11-14}

& & & &
\textbf{F1} & \textbf{IoU} &
\textbf{Conc.} & \textbf{Fac.} & \textbf{Road} & \textbf{Avg.$\pm$Std} &
\textbf{Conc.} & \textbf{Fac.} & \textbf{Road} & \textbf{Avg.$\pm$Std} \\

\midrule

\multirow{8}{*}{SAM2-Tiny}
& 0.0   & 4.7K  & 3.82M  & 61.96 & 44.89 & 72.77 & \textbf{69.65} & \textbf{66.49} & $\mathbf{69.64_{\pm3.14}}$ & 57.20 & \textbf{53.43} & \textbf{49.80} & $\mathbf{53.48_{\pm3.70}}$ \\
& -0.5  & 4.7K  & 3.08M  & 61.42 & 44.32 & 71.87 & \underline{69.22} & 65.85 & $68.98_{\pm3.02}$ & 56.09 & \underline{52.93} & 49.09 & $52.70_{\pm3.51}$ \\
& -0.8  & 0.20M & 5.40M  & 62.40 & 45.35 & 72.51 & 68.58 & \underline{66.29} & $\underline{69.13_{\pm3.15}}$ & 56.88 & 52.19 & \underline{49.58} & $\underline{52.88_{\pm3.70}}$ \\
& \cellcolor{blue!8}-0.85 & \cellcolor{blue!8}0.27M & \cellcolor{blue!8}6.95M  & \cellcolor{blue!8}63.12 & \cellcolor{blue!8}46.11 & \cellcolor{blue!8}72.43 & \cellcolor{blue!8}69.10 & \cellcolor{blue!8}64.42 & \cellcolor{blue!8}$68.65_{\pm4.02}$ & \cellcolor{blue!8}56.77 & \cellcolor{blue!8}52.79 & \cellcolor{blue!8}47.52 & \cellcolor{blue!8}$52.36_{\pm4.64}$ \\
& -0.87 & 9.9M  & 12.13M & 64.20 & 47.27 & \underline{72.98} & 68.84 & 64.98 & $68.93_{\pm4.00}$ & \underline{57.45} & 52.48 & 48.14 & $52.69_{\pm4.67}$ \\
& -0.9  & 11.8M & 13.69M & 64.28 & 47.36 & \textbf{72.99} & 68.74 & 64.99 & $68.91_{\pm4.00}$ & \textbf{57.47} & 52.37 & 48.14 & $52.66_{\pm4.67}$ \\
& -0.95 & 13.1M & 15.02M & \underline{64.41} & \underline{47.50} & 72.92 & 68.99 & 64.97 & $68.96_{\pm3.98}$ & 57.39 & 52.67 & 48.11 & $52.72_{\pm4.64}$ \\
& -1.0    & 16.6M & 18.07M & \textbf{64.51} & \textbf{47.61} & 72.94 & 69.05 & 64.97 & $68.98_{\pm3.99}$ & 57.40 & 52.73 & 48.11 & $52.75_{\pm4.65}$ \\

\midrule

\multirow{6}{*}{SAM2-Large}
& 0.0   & 7.1K  & 17.41M  & 57.43 & 40.29 & 65.17 & 73.55 & 66.44 & $68.39_{\pm3.69}$ & 48.33 & 58.17 & 49.74 & $52.08_{\pm4.34}$ \\
& -0.5  & 7.1K  & 25.62M  & 59.60 & 42.45 & 68.94 & 73.93 & 67.63 & $70.17_{\pm2.72}$ & 52.60 & 58.64 & 51.09 & $54.11_{\pm3.26}$ \\
& -0.7  & 1.3M  & 34.36M  & 62.77 & 45.74 & 72.63 & 74.10 & 68.58 & $71.77_{\pm2.33}$ & 57.02 & 58.85 & 52.18 & $56.02_{\pm2.81}$ \\
& -0.85 & 2.0M  & 59.40M  & 64.57 & 47.67 & 71.99 & 74.69 & 68.04 & $71.57_{\pm2.73}$ & 56.25 & 59.61 & 51.56 & $55.80_{\pm3.30}$ \\
& -0.86 & 2.1M  & 55.36M  & 64.33 & 47.41 & 71.81 & \textbf{75.79} & 68.54 & $72.05_{\pm2.96}$ & 56.02 & \textbf{61.01} & 52.14 & $56.39_{\pm3.63}$ \\
& \cellcolor{blue!8}-0.87 & \cellcolor{blue!8}3.0M  & \cellcolor{blue!8}71.49M  & \cellcolor{blue!8}\underline{65.78} & \cellcolor{blue!8}\underline{49.01} & \cellcolor{blue!8}72.81 & \cellcolor{blue!8}\underline{75.30} & \cellcolor{blue!8}70.29 & \cellcolor{blue!8}$\underline{72.80_{\pm2.04}}$ & \cellcolor{blue!8}57.24 & \cellcolor{blue!8}60.38 & \cellcolor{blue!8}54.19 & \cellcolor{blue!8}$\underline{57.27_{\pm2.53}}$ \\
& -0.88 & 2.5M  & 63.52M  & 65.38 & 48.57 & 72.17 & 74.90 & 69.33 & $72.13_{\pm2.28}$ & 56.46 & 59.87 & 53.05 & $56.46_{\pm2.78}$ \\
& -0.9  & 56.8M & 82.54M  & 64.72 & 47.84 & \textbf{73.32} & 74.72 & \textbf{71.27} & $\mathbf{73.10_{\pm1.42}}$ & \textbf{57.88} & 59.64 & \textbf{55.36} & $\mathbf{57.63_{\pm1.76}}$ \\
& -0.95 & 74.5M & 97.12M  & \textbf{65.81} & \textbf{49.04} & \underline{73.13} & 73.89 & 69.24 & $72.08_{\pm2.03}$ & \underline{57.64} & 58.59 & 52.95 & $56.39_{\pm2.46}$ \\
& -1.0  & 81.9M & 101.14M & 65.56 & 48.77 & 72.10 & \underline{75.37} & \underline{70.60} & $72.69_{\pm1.99}$ & 56.38 & \underline{60.48} & \underline{54.56} & $\underline{57.14_{\pm2.48}}$ \\

\bottomrule
\end{tabularx}
\end{table}
\FloatBarrier

\begin{table}[H]
\centering
\caption{Ablation on the freezing threshold $\lambda$ for \textbf{FisherAdapTune} applied to crack segmentation using SegFormer-MiT-B0 and SegFormer-MiT-B3. Larger $|\lambda|$ retains more trainable parameter groups. Bold shows best per column within each model block; \underline{underline} shows second best. Shaded rows indicate the selected $\lambda$. Effect. Params denotes number of Effective Tunable Parameters.}
\label{tab:segformer_lambda_ablation}
\scriptsize
\setlength{\tabcolsep}{2pt}
\renewcommand{\arraystretch}{1.15}

\begin{tabularx}{\textwidth}{
>{\raggedright\arraybackslash}p{1.4cm}
>{\centering\arraybackslash}m{0.7cm}
>{\centering\arraybackslash}m{0.9cm}
>{\centering\arraybackslash}m{0.7cm}
cc
cccc
cccc
}
\toprule
\textbf{Model} &
$\boldsymbol{\lambda}$ &
\makecell{\textbf{Final}\\\textbf{Tunable}\\\textbf{Params}} &
\makecell{\textbf{Effect.}\\\textbf{Params}} &
\multicolumn{2}{c}{\textbf{OmniCrack}} &
\multicolumn{4}{c}{\textbf{Zero-shot F1 (\%)}} &
\multicolumn{4}{c}{\textbf{Zero-shot IoU (\%)}} \\

\cmidrule(lr){5-6}
\cmidrule(lr){7-10}
\cmidrule(lr){11-14}

& & & &
\textbf{F1} & \textbf{IoU} &
\textbf{Conc.} & \textbf{Fac.} & \textbf{Road} & \textbf{Avg.$\pm$Std} &
\textbf{Conc.} & \textbf{Fac.} & \textbf{Road} & \textbf{Avg.$\pm$Std} \\

\midrule

\multirow{11}{*}{SegFormer-B0}
& $-0.50$ & 0.032K & 0.047M & 58.44 & 41.28 & 70.18 & 58.70 & 52.30 & $60.39_{\pm7.39}$ & 54.06 & 41.54 & 35.41 & $43.66_{\pm7.76}$ \\
& $-0.65$ & 0.032K & 0.074M & 59.53 & 42.37 & 71.40 & \underline{59.17} & 52.03 & $60.86_{\pm7.99}$ & 55.52 & \underline{42.02} & 35.17 & $44.23_{\pm8.45}$ \\
& $-0.75$ & 0.032K & 0.085M & 59.65 & 42.50 & 71.36 & \textbf{59.92} & \textbf{56.06} & $\mathbf{62.44_{\pm6.49}}$ & 55.47 & \textbf{42.77} & \textbf{38.95} & $\mathbf{45.73_{\pm7.06}}$ \\
& $-0.80$ & 0.064K & 0.089M & 60.16 & 43.02 & 70.71 & 58.99 & 54.99 & $\underline{61.56_{\pm6.67}}$ & 54.70 & 41.83 & \underline{37.92} & $44.81_{\pm7.16}$ \\
& $-0.85$ & 0.064K & 0.093M & 60.23 & 43.09 & 69.66 & 58.54 & 54.52 & $60.90_{\pm6.40}$ & 53.44 & 41.38 & 37.48 & $44.09_{\pm6.79}$ \\
& $-0.90$ & 0.096K & 0.100M & 60.09 & 42.95 & 68.77 & 58.50 & 52.33 & $59.86_{\pm6.78}$ & 52.41 & 41.35 & 35.43 & $43.06_{\pm7.03}$ \\
\rowcolor{blue!8}
& $-0.95$ & 0.128K & 0.124M & 60.85 & 43.73 & 70.97 & 58.77 & 53.63 & $61.12_{\pm7.27}$ & 55.01 & 41.62 & 36.64 & $44.42_{\pm7.75}$ \\
& $-1.00$ & 0.160K & 0.144M & 60.79 & 43.67 & 70.62 & 59.02 & 53.29 & $60.97_{\pm7.20}$ & 54.59 & 41.87 & 36.32 & $44.26_{\pm7.64}$ \\
& $-1.25$ & 0.048M & 0.360M & 62.28 & 45.22 & 70.52 & 58.89 & 52.59 & $60.67_{\pm7.43}$ & 54.46 & 41.73 & 35.67 & $43.96_{\pm7.83}$ \\
& $-1.5$  & 1.31M  & 1.85M  & 62.96 & 45.95 & \textbf{73.03} & 56.58 & \underline{54.98} & $61.53_{\pm8.16}$ & \textbf{57.52} & 39.45 & 37.91 & $44.96_{\pm8.90}$ \\
& $-2$    & 3.37M  & 3.54M  & \underline{63.23} & \underline{46.23} & \underline{73.02} & 58.01 & 53.64 & $\underline{61.56_{\pm8.30}}$ & \underline{57.50} & 40.85 & 36.65 & $\underline{45.00_{\pm9.00}}$ \\
& $-3$    & 3.71M  & 3.71M  & \textbf{63.25} & \textbf{46.25} & 72.76 & 56.96 & 53.56 & $61.09_{\pm8.36}$ & 57.18 & 39.82 & 36.57 & $44.52_{\pm9.05}$ \\

\midrule

\multirow{8}{*}{SegFormer-B3}
& $-0.50$ & 3K  & 5.02M  & 61.84 & 44.76 & 73.21 & \underline{60.80} & 52.66 & $62.22_{\pm8.44}$ & 57.74 & \underline{43.68} & 35.74 & $45.72_{\pm9.09}$ \\
& $-0.65$ & 4K  & 6.19M  & 62.83 & 45.81 & 72.90 & 60.73 & 51.72 & $61.78_{\pm8.67}$ & 57.36 & 43.61 & 34.88 & $45.28_{\pm9.25}$ \\
& $-0.75$ & 6K  & 7.10M  & 62.94 & 45.92 & 73.12 & 60.19 & 53.45 & $62.25_{\pm8.16}$ & 57.63 & 43.05 & 36.47 & $45.71_{\pm8.84}$ \\
& $-0.80$ & 7K  & 7.80M  & 63.76 & 46.80 & 73.64 & 59.21 & 53.38 & $62.07_{\pm8.51}$ & 58.29 & 42.06 & 36.40 & $45.58_{\pm9.27}$ \\
& $-0.85$ & 7K  & 8.48M  & 63.54 & 46.57 & 73.04 & 59.89 & 55.38 & $62.77_{\pm7.49}$ & 57.53 & 42.74 & 38.30 & $46.19_{\pm8.22}$ \\
& $-0.90$ & 7K  & 9.38M  & 63.90 & 46.95 & 73.35 & 59.39 & \underline{57.76} & $\underline{63.50_{\pm6.99}}$ & 57.91 & 42.24 & \underline{40.60} & $46.91_{\pm7.80}$ \\
\rowcolor{blue!8}
& $-0.95$ & 11K & 10.73M & \underline{64.57} & \underline{47.68} & \underline{74.19} & \textbf{61.04} & \textbf{58.26} & $\mathbf{64.49_{\pm6.94}}$ & \underline{58.97} & \textbf{43.93} & \textbf{41.11} & $\mathbf{48.00_{\pm7.83}}$ \\
& $-1.00$ & 14K & 12.81M & \textbf{64.97} & \textbf{48.11} & \textbf{75.85} & 57.83 & 56.52 & $63.40_{\pm8.81}$ & \textbf{61.10} & 40.68 & 39.40 & $\underline{47.06_{\pm9.94}}$ \\

\bottomrule
\end{tabularx}
\vspace{2pt}
\end{table}
\FloatBarrier

Figure~\ref{fig:lambda_ablations} summarizes the threshold trends across both model families. More conservative thresholds retain more parameters and generally improve in-distribution performance, whereas zero-shot performance often peaks at an intermediate budget. Thus, the optimal threshold balances downstream adaptation, zero-shot generalization, and parameter efficiency.

\subsection{Active-Parameter Trajectories Across Thresholds}
\label{appendix:active_param_trajectories}

\begin{figure}[H]
    \centering
    \includegraphics[width=0.76\linewidth,trim={0 0.1in 0 0.25in},clip]{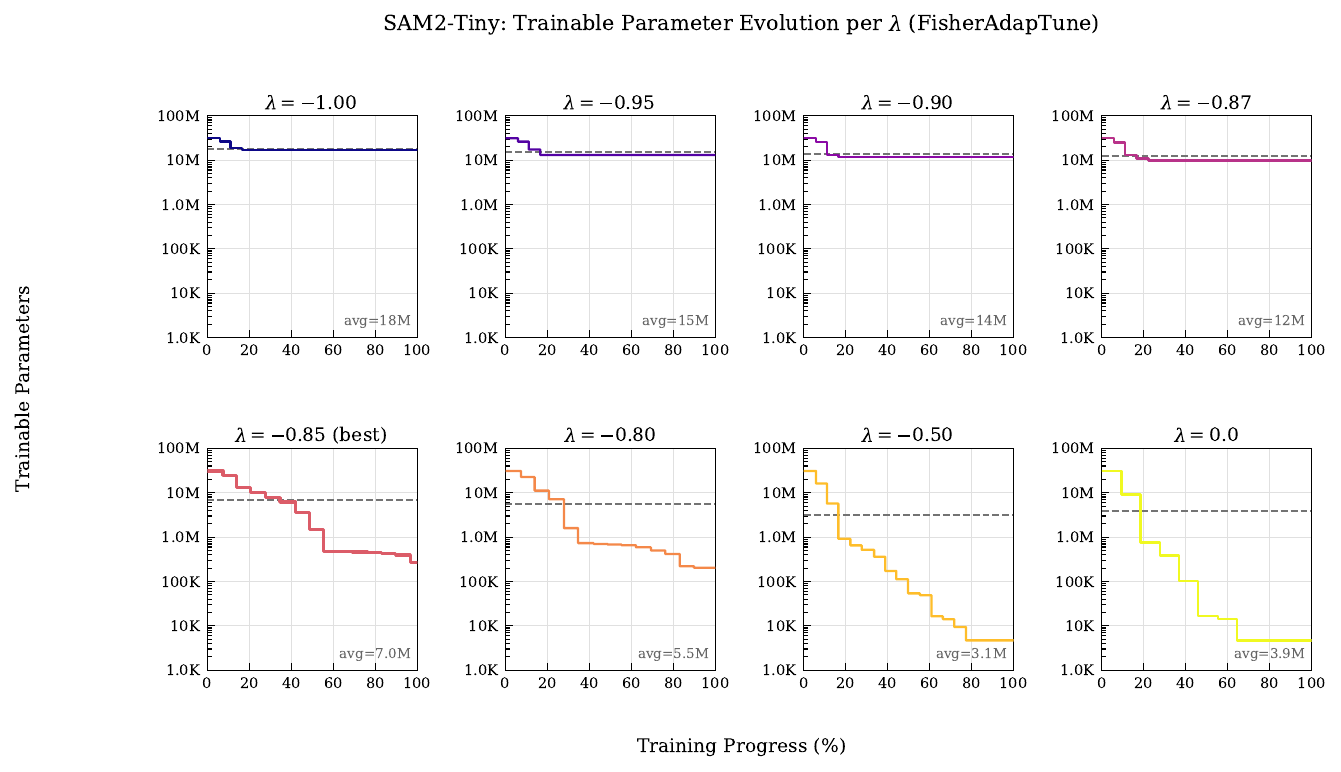}
    \caption{Trainable-parameter evolution for SAM2-Tiny across freezing thresholds $\lambda$. "avg." denotes the effective number of trainable parameters.}
    \label{fig:sam2_tiny_lambda_ablation_grid}
\end{figure}

\begin{figure}[H]
    \centering
    \includegraphics[width=0.76\linewidth,trim={0 0.1in 0 0.25in},clip]{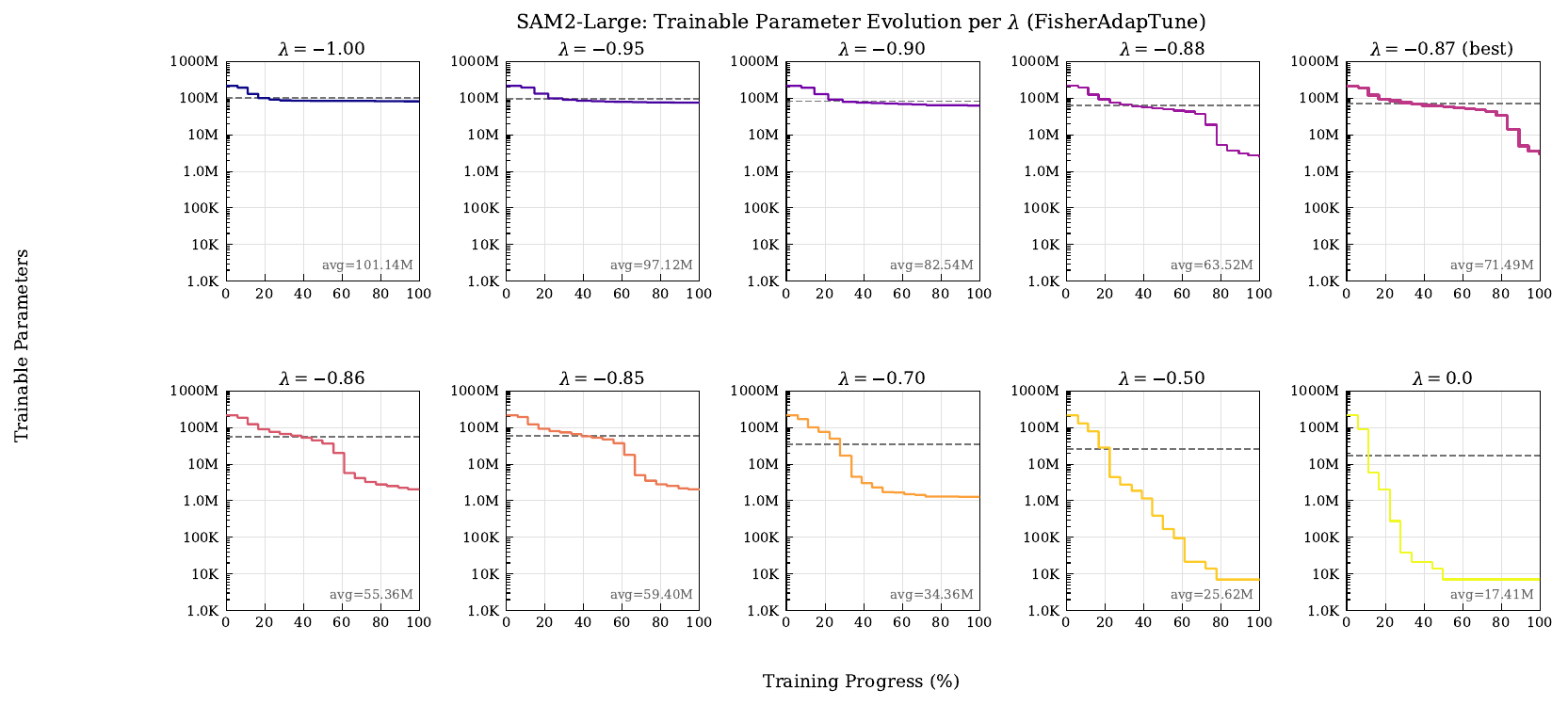}
    \caption{Trainable-parameter evolution for SAM2-Large across freezing thresholds $\lambda$. "avg." denotes the effective number of trainable parameters.}
    \label{fig:sam2_large_lambda_ablation_grid}
\end{figure}

\begin{figure}[H]
    \centering
    \includegraphics[width=0.76\linewidth,trim={0 0.1in 0 0.55in},clip]{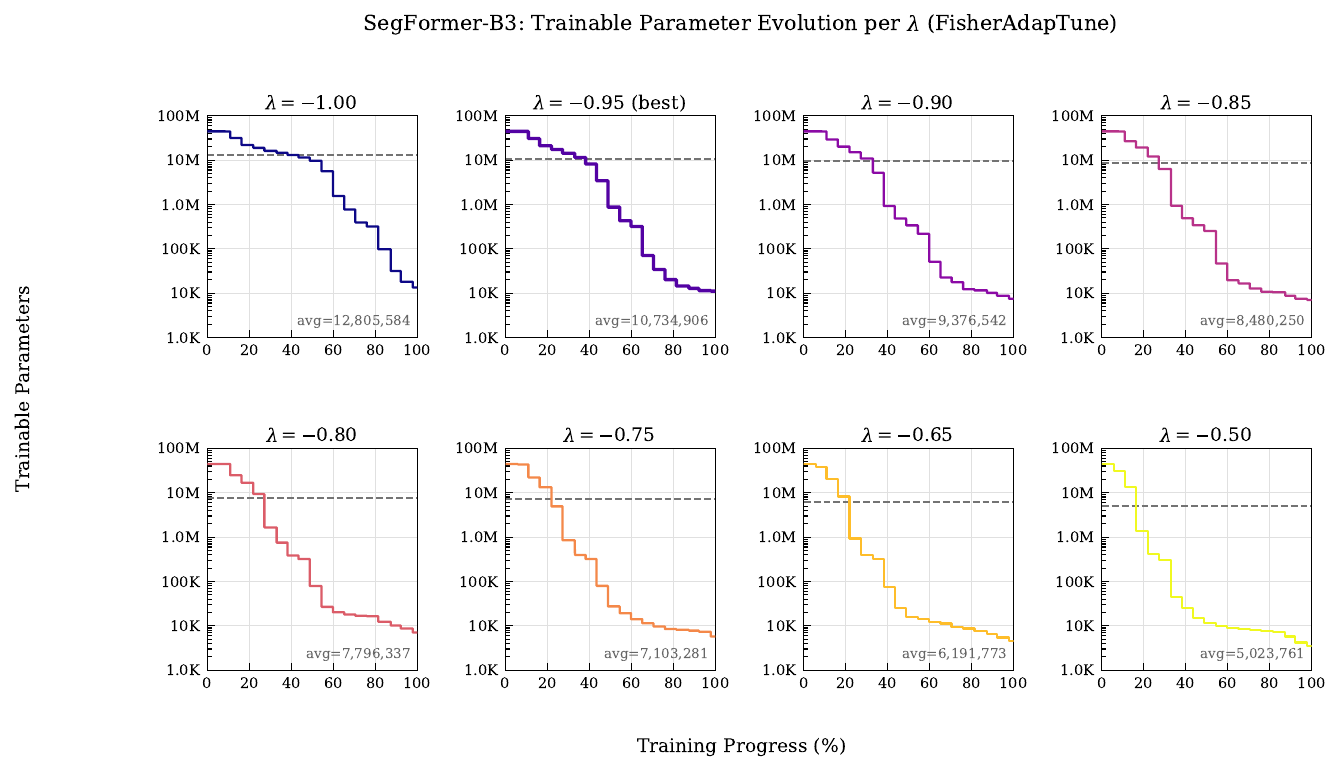}
    \caption{Trainable-parameter evolution for SegFormer-MiT-B3 across freezing thresholds $\lambda$. "avg." denotes the effective number of trainable parameters.}
    \label{fig:segformer_b3_lambda_ablation_grid}
\end{figure}

\begin{figure}[H]
    \centering
    \includegraphics[width=0.68\linewidth,trim={0 0.1in 0 0.45in},clip]{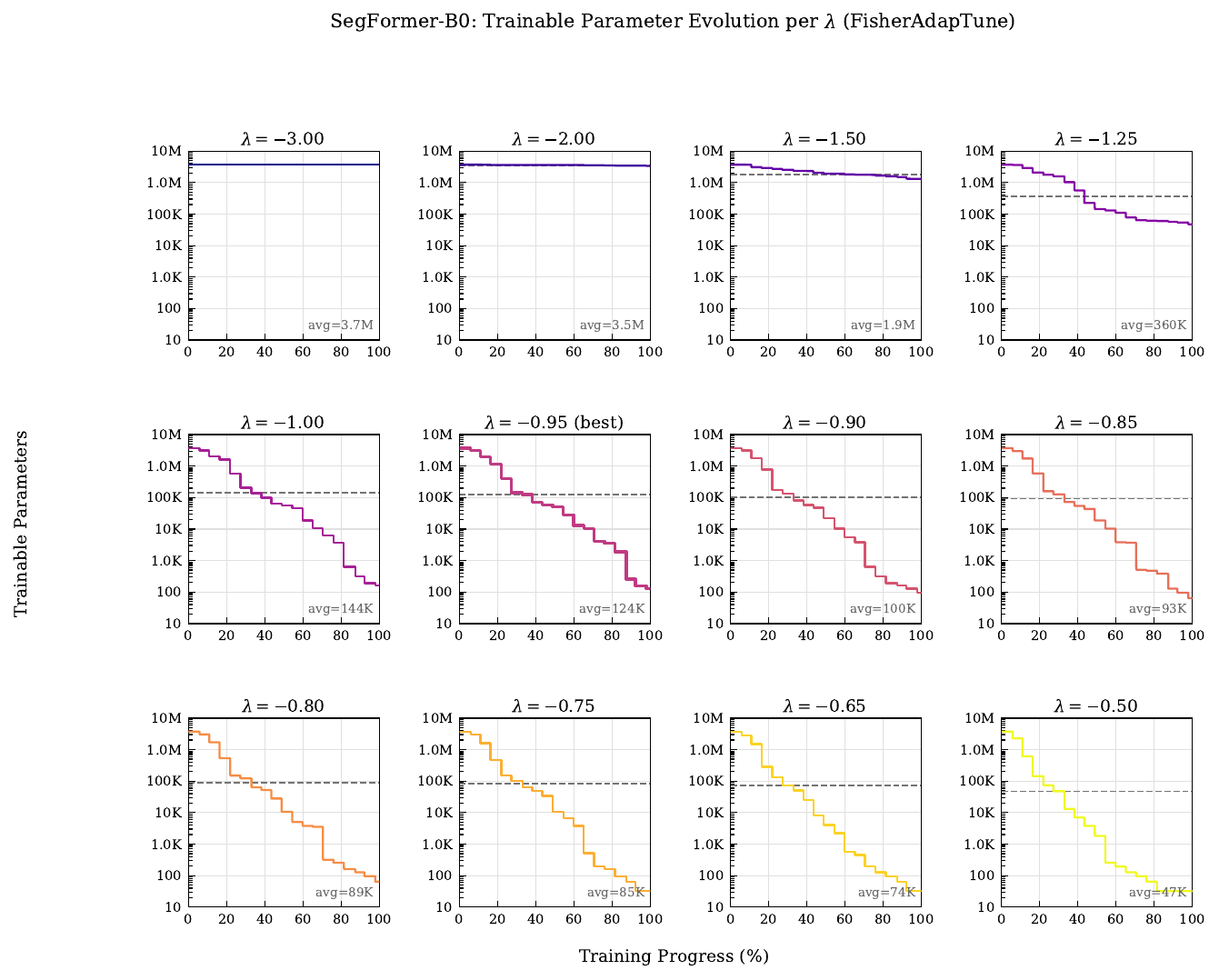}
    \caption{Trainable-parameter evolution for SegFormer-MiT-B0 across freezing thresholds $\lambda$. "avg." denotes the effective number of trainable parameters.}
    \label{fig:segformer_b0_lambda_ablation_grid}
\end{figure}

Figures~\ref{fig:sam2_tiny_lambda_ablation_grid}-\ref{fig:segformer_b0_lambda_ablation_grid} show how the active trainable-parameter budget evolves throughout optimization. These trajectories complement the aggregate counts in Tables~\ref{tab:sam2_lambda_ablation} and~\ref{tab:segformer_lambda_ablation} by showing when parameter groups are frozen. Thresholds with smaller absolute values satisfy the freezing criterion earlier, producing a rapid decline in the active budget. More negative thresholds apply a more conservative criterion, keeping a larger portion of the model trainable for longer.
The trajectories also reveal architecture-dependent transitions in the final active set. In SAM2, small changes near the selected thresholds can cause large differences in the number of retained parameters, particularly for the larger backbone. SegFormer exhibits the same general relationship between $\lambda$ and budget, although its finer parameter grouping permits very small final active sets. Overall, the plots confirm that $\lambda$ controls not only the final number of trainable parameters but also the duration for which different parameter groups contribute to adaptation.

\FloatBarrier

\section{Fisher Drift and Parameter-Selection Dynamics}
\label{appendix:fisher_drift_dynamics}

\subsection{Heterogeneity Across Layers and Parameter Groups}
\label{appendix:Fisher_Drift_Observations}

\begin{figure}[H]
        \includegraphics[width=0.95\linewidth]{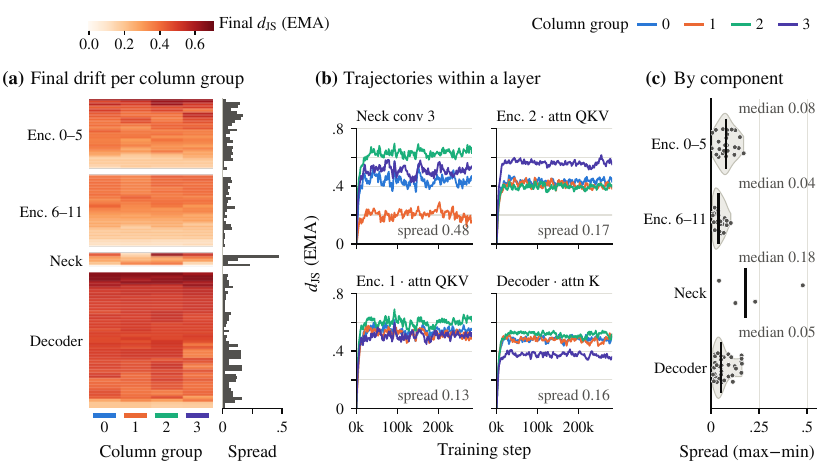}
    \caption{Parameter group-wise Jensen-Shannon (JS) distance between consecutive Fisher distributions in SAM2-Tiny during full fine-tuning on crack segmentation; $EMA[d_{\mathrm{JS}_{i}}^{t}(p(\tilde{\mathbf{F}}_{i}^{t-1}),p(\tilde{\mathbf{F}}_{i}^{t}))]$.}
    \label{fig:sam2_chunks_js_drift}
\end{figure}

\begin{figure}[H]
    \centering
        \includegraphics[width=0.9\linewidth]{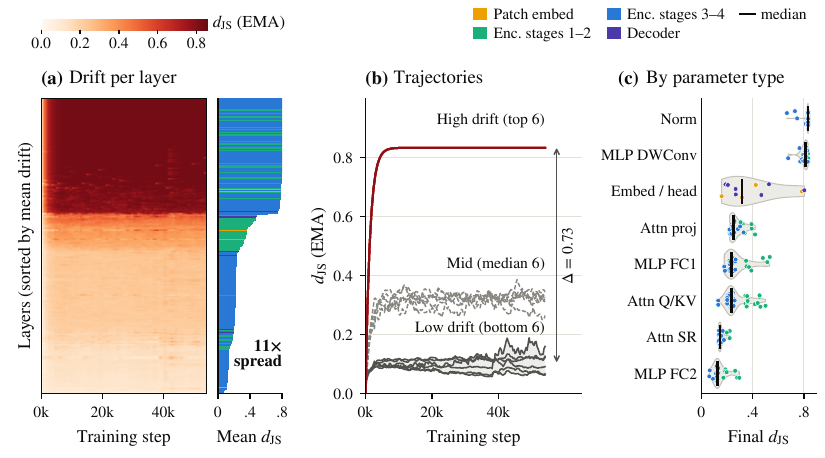}
        \caption{Layer-wise Jensen-Shannon (JS) distance between consecutive Fisher distributions in SegFormer during full fine-tuning on crack segmentation; $EMA[d_{\mathrm{JS}_{i}}^{t}(p(\tilde{\mathbf{F}}_{i}^{t-1}),p(\tilde{\mathbf{F}}_{i}^{t}))]$.}
    \label{fig:additional_segformer_js_drift_observation}
\end{figure}

To examine whether Fisher geometry evolves uniformly during adaptation, we tracked the exponential moving average of the Jensen-Shannon distance between consecutive Fisher-value distributions throughout full fine-tuning. Figures~\ref{fig:sam2_chunks_js_drift} and~\ref{fig:additional_segformer_js_drift_observation} reveal evident heterogeneity at multiple levels. In Figure~\ref{fig:sam2_chunks_js_drift}, Fisher drift varies not only across layers but also among column-wise parameter groups within the same layer, with some groups exhibiting different convergence levels. SegFormer-B3 similarly shows a broad layer-wise spread in drift, while normalization, attention, and MLP/fully connected parameters form distinct drift distributions. These observations indicate that parameter groups stabilize at different rates and that a single architecture-wide freezing schedule cannot capture their heterogeneous adaptation dynamics, motivating FisherAdapTune's progressive, group-specific selection strategy.

\FloatBarrier

\clearpage
\subsection{Architectural Patterns in the Selected Parameters}
\label{appendix:param_selection}

\begin{figure}[H]
    \centering
    \begin{minipage}[t]{0.32\linewidth}
        \vspace{0pt}
        \centering
        \includegraphics[width=\linewidth]{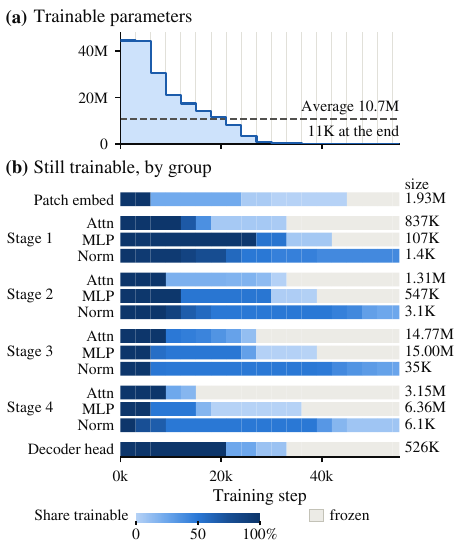}
        \caption{Trainable parameter budget and group lifetimes for SegFormer-MiT-B3.}
        \label{fig:param_selection_dynamics_segformer}
    \end{minipage}\hfill
    \begin{minipage}[t]{0.65\linewidth}
        \vspace{0pt}
        \centering
        \includegraphics[width=\linewidth]{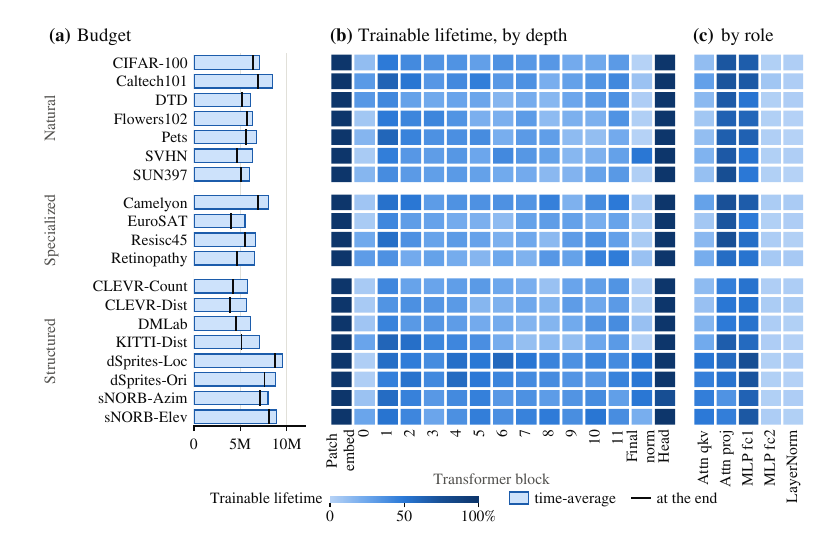}
        \caption{Trainable parameter budgets and group lifetimes for DINOv2 across 19 VTAB-1k tasks.}
        \label{fig:param_selection_dynamics_dino}
    \end{minipage}
\end{figure}

The selected groups reveal different adaptation trajectories across
architectures. For SAM2-Tiny, large encoder attention and MLP groups
freeze relatively early, whereas patch embedding and several
mask-decoder groups remain trainable longer
(Figure~\ref{fig:param_selection_dynamics_sam2}). Several
normalization groups also have long trainable lifetimes despite their
small parameter counts. Thus, the reduction in trainable budget is
driven largely by freezing high-capacity encoder groups. It does not
imply that all components in a stage stop adapting together.
Across all VTAB-1k tasks, DINOv2 exhibits a recurring pattern:
patch embedding and the classification head have the longest trainable lifetimes, while transformer-block groups freeze at different rates
(Figure~\ref{fig:param_selection_dynamics_dino}). This is consistent
with continued adjustment at the input and task-specific output,
although trainable lifetime alone does not establish the function of
those updates. SegFormer-MiT-B3 shows a different allocation where
intermediate encoder groups, particularly Stage 3 attention and MLP,
remain active longer, alongside selected normalization groups
(Figure~\ref{fig:param_selection_dynamics_segformer}). Overall,
the patterns show that FisherAdapTune does not recover a fixed freezing order or a specific block-type training strategy.
Instead, the location and duration of adaptation vary with the
architecture and downstream task.

\end{document}